\documentclass[10pt,twocolumn,letterpaper]{article}

\pdfoutput=1

\usepackage{cvpr}
\usepackage{times}
\usepackage{epsfig}
\usepackage{graphicx}
\usepackage{amsmath}
\usepackage{amssymb}

\usepackage[caption=false]{subfig}
\usepackage{multirow} %
\usepackage[numbers,sort]{natbib} %
\usepackage{mathtools,xparse}
\usepackage{rotating}
\usepackage{cancel} %
\usepackage{color}

\def\vec#1{\mathchoice%
	{\mbox{\boldmath $\displaystyle\bf#1$}}
	{\mbox{\boldmath $\textstyle\bf#1$}}
	{\mbox{\boldmath $\scriptstyle\bf#1$}}
	{\mbox{\boldmath $\scriptscriptstyle\bf#1$}}}
\def\vc#1{\protect\vec #1}

\DeclarePairedDelimiter{\norm}{\lVert}{\rVert}

\DeclareMathOperator*{\argmin}{\arg\!\min}

\renewcommand{\paragraph}[1]{\smallskip\noindent{\bf{#1}}}

\newcommand{\isArXiv}[2]{#1}    %
\usepackage[pagebackref=true,breaklinks=true,letterpaper=true,colorlinks,bookmarks=false]{hyperref}

\hypersetup{pdfauthor={Relja Arandjelovic},pdftitle={NetVLAD}}

    \cvprfinalcopy %

\def\cvprPaperID{115}

\isArXiv{}{
    \ifcvprfinal\pagestyle{empty}\fi
}

\begin{document}

\title{NetVLAD: CNN architecture for weakly supervised place recognition}

\author{
Relja Arandjelovi\'c \\
INRIA
\thanks{
    WILLOW project,
    Departement d'Informatique de l'\'Ecole Normale Sup\'erieure,
    ENS/INRIA/CNRS UMR 8548.}
\and
Petr Gronat \\
INRIA$^*$
\and
Akihiko Torii \\
Tokyo Tech
\thanks{
    Department of Mechanical and Control Engineering,
    Graduate School of Science and Engineering,
    Tokyo Institute of Technology}
\and
Tomas Pajdla \\
CTU in Prague
\thanks{
    Center for Machine Perception,
    Department of Cybernetics,
    Faculty of Electrical Engineering,
    Czech Technical University in Prague}\\
\and
Josef Sivic \\
INRIA$^*$
}

\maketitle

\def\figStoaCaption{
{\bf Comparison of our methods versus off-the-shelf networks and state-of-the-art.}
The base CNN architecture is denoted in brackets: (A)lexNet and (V)GG-16.
Trained representations
({\color{red}red} and {\color{magenta}magenta} for
{\color{red}AlexNet} and {\color{magenta}VGG-16})
 outperform by a large margin off-the-shelf ones
({\color{blue}blue},
{\color{cyan}cyan},
{\color{green}green} for
{\color{blue}AlexNet},
{\color{cyan}Places205},
{\color{green}VGG-16}),
$f_{VLAD}$ (-o-) works better than $f_{max}$ (-x-),
and our $f_{VLAD}$+whitening ({\color{magenta}-$\ast$-}) representation
based on VGG-16
sets the state-of-the-art on all datasets.
}

\def\figFocusCaption{
{\bf What has been learnt?}
Each column corresponds to one image (top row) and the emphasis various networks
(under $f_{max}$) give to different patches.
Each pixel in the heatmap corresponds to the change in representation
when a large gray occluding square ($100 \times 100$) is placed over the image in the same position;
all heatmaps have the same colour scale.
Note that the original image and the heatmaps are not in perfect alignment
as nearby patches
overlap 50\% and patches touching an image edge are discarded to
prevent border effects. All images are from Pitts250k-val that the network
hasn't seen at training.
}

\def\figDim{
\begin{figure}[!t]
\vspace{-0.3cm}
\begin{center}
    \subfloat[Pitts250k-test]{
        \includegraphics[width=0.45\linewidth]{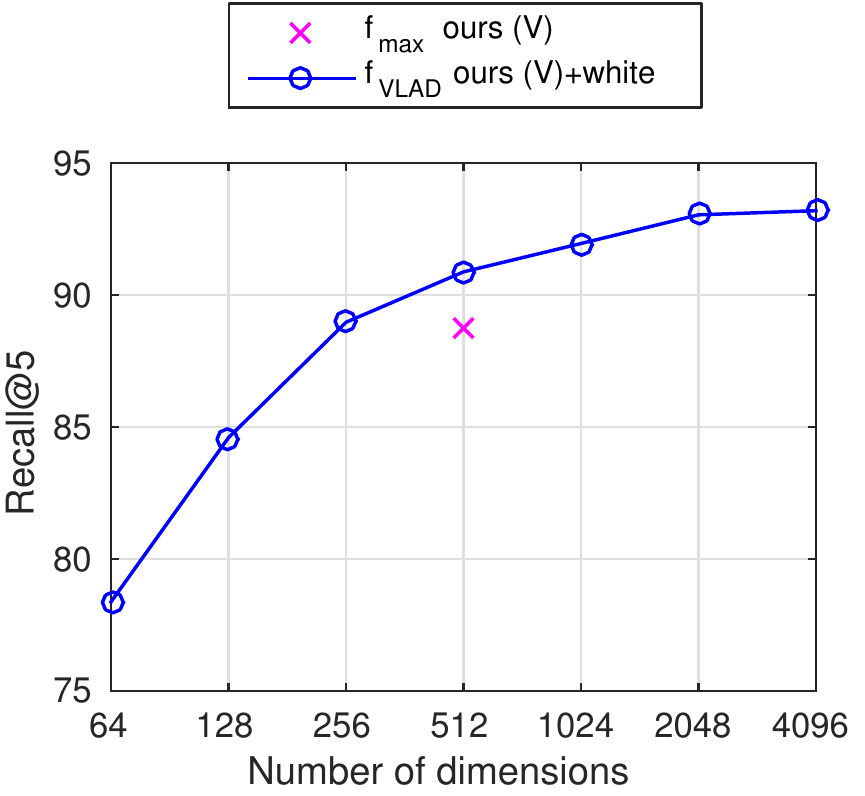}
    }
    \subfloat[Tokyo 24/7]{
        \includegraphics[width=0.45\linewidth]{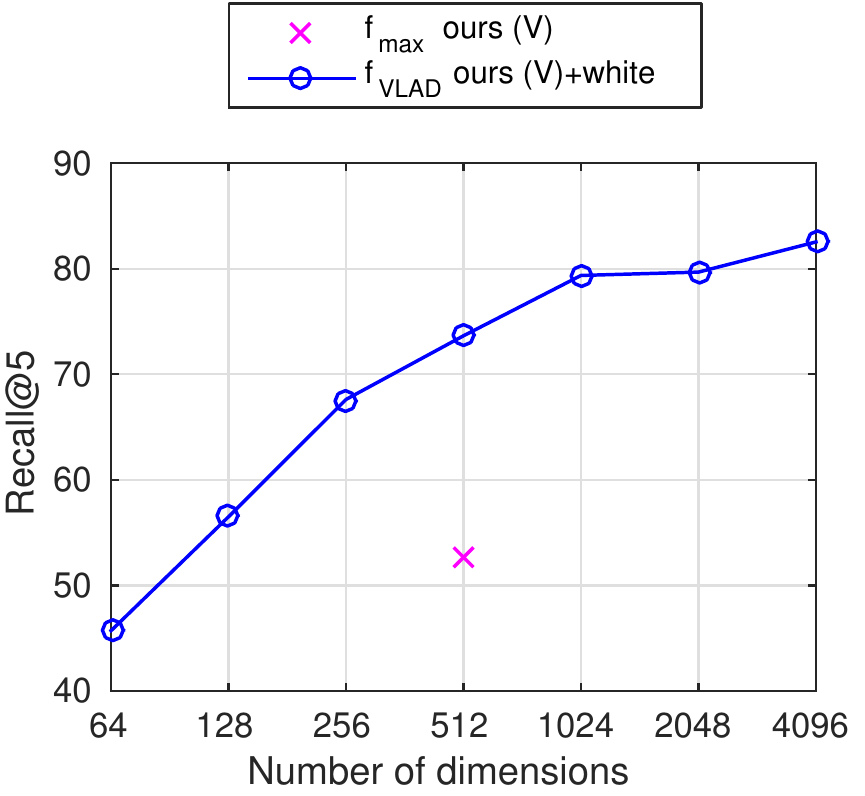}
    }
\caption{ {\bf Place recognition accuracy versus dimensionality.}
Note the log scale of the x-axis.
128-D NetVLAD performs comparably to the $4\times$ larger 512-D $f_{max}$ on Tokyo 24/7.
For the same dimension (512-D) NetVLAD convincingly outperforms $f_{max}$.
\vspace{-0.5cm}
}
\label{fig:dim}
\end{center}
\end{figure}
}

\def\tabResRet{
\begin{table*}[!t]
\begin{center}
\small
\begin{tabular}{l@{~~~}|@{~~~}c@{~~~}c@{~~~}c@{~~~}c@{~~~}c@{~~~}c}
    Method
    & Oxford 5k (full)
    & Oxford 5k (crop)
    & Paris 6k (full)
    & Paris 6k (crop)
    & Holidays (orig)
    & Holidays (rot)
    \\
    \hline
    J\'egou and Zisserman \cite{Jegou14}
        & --
        & 47.2
        & --
        & --
        & 65.7
        & 65.7
    \\
    Gordo \etal \cite{Gordo12}
        & --
        & --
        & --
        & --
        & 78.3
        & --
    \\
    Razavian \etal \cite{Razavian15}
        & 53.3$^\dagger$
        & --
        & 67.0$^\dagger$
        & --
        & 74.2
        & --
    \\
    Babenko and Lempitsky \cite{Babenko15}
        & 58.9
        & 53.1
        & --
        & --
        & --
        & 80.2
    \\
    a. Ours: NetVLAD off-the-shelf
        & 53.4
        & 55.5
        & 64.3
        & 67.7
        & {\bf 82.1}
        & {\bf 86.0}
    \\
    b. Ours: NetVLAD trained
        & {\bf 62.5}
        & {\bf 63.5}
        & {\bf 72.0}
        & {\bf 73.5}
        & 79.9
        & 84.3
    \\
\end{tabular}
\end{center}
 \vspace{-0.2cm}
\caption{ {\bf Comparison with state-of-the-art compact image representations (256-D) on image and object retrieval.}
We compare (b.) our best trained network,
(a.) the corresponding off-the-shelf network (whitening learnt on Pittsburgh),
and the state-of-the-art for compact image representations
on standard image and object retrieval benchmarks.
``orig'' and ``rot'' for Holidays denote whether the original
or the manually rotated dataset \cite{Babenko14,Babenko15} is used.
The ``crop'' and ``full'' for Oxford/Paris correspond to the
testing procedures when the query ROI is respected (the image is cropped as in \cite{Babenko15}),
or ignored (the full image is used as the query), respectively.
$\dagger$ \cite{Razavian15} use square patches whose side is equal to $1.5\times$
the maximal dimension of the query ROI (the detail is available in version 2 of
the arXiv paper \cite{Razavian14a}), so the setting is somewhere in between
``crop'' and ``full'', arguably closer to ``full'' as ROIs become very large.
}
\label{tab:retrieval}
\end{table*}
}

\def\tabResRetSmall{
\begin{table}[!t]
\begin{center}
\small
\hspace*{-0.4cm}%
\begin{tabular}{l@{~~~}|@{~~}c@{~}c@{~~}|@{~~}c@{~}c@{~~}|@{~~}c@{~}c@{~}}
    Method
    & \multicolumn{2}{c}{Oxford 5k}
    & \multicolumn{2}{c}{Paris 6k}
    & \multicolumn{2}{c}{Holidays}
    \\
    & full & crop & full & crop & orig & rot
    \\
    \hline
    J\'egou and Zisserman \cite{Jegou14}
        & --
        & 47.2
        & --
        & --
        & 65.7
        & 65.7
    \\
    Gordo \etal \cite{Gordo12}
        & --
        & --
        & --
        & --
        & 78.3
        & --
    \\
    Razavian \etal \cite{Razavian15}
        & 53.3$^\dagger$
        & --
        & 67.0$^\dagger$
        & --
        & 74.2
        & --
    \\
    Babenko and Lempitsky \cite{Babenko15}
        & 58.9
        & 53.1
        & --
        & --
        & --
        & 80.2
    \\
    a. Ours: NetVLAD off-shelf
        & 53.4
        & 55.5
        & 64.3
        & 67.7
        & {\bf 82.1}
        & {\bf 86.0}
    \\
    b. Ours: NetVLAD trained
        & {\bf 62.5}
        & {\bf 63.5}
        & {\bf 72.0}
        & {\bf 73.5}
        & 79.9
        & 84.3
    \\
\end{tabular}
\end{center}
 \vspace{-0.2cm}
\caption{ {\bf Comparison with state-of-the-art compact image representations (256-D) on image and object retrieval.}
We compare (b.) our best trained network,
(a.) the corresponding off-the-shelf network (whitening learnt on Pittsburgh),
and the state-of-the-art for compact image representations
on standard image and object retrieval benchmarks.
``orig'' and ``rot'' for Holidays denote whether the original
or the manually rotated dataset \cite{Babenko14,Babenko15} is used.
The ``crop'' and ``full'' for Oxford/Paris correspond to the
testing procedures when the query ROI is respected (the image is cropped as in \cite{Babenko15}),
or ignored (the full image is used as the query), respectively.
$\dagger$ \cite{Razavian15} use square patches whose side is equal to $1.5\times$
the maximal dimension of the query ROI (the detail is available in version 2 of
the arXiv paper \cite{Razavian14a}), so the setting is somewhere in between
``crop'' and ``full'', arguably closer to ``full'' as ROIs become very large.
}
\label{tab:retrievalSmall}
\end{table}
}

\def\tabResTM{
\begin{table}
\begin{center}
\begin{tabular}{l|cc}
    Training data & recall@1 & recall@10 \\
    \hline
    Pretrained on ImageNet \cite{Krizhevsky12}  & 33.5 & 68.5 \\
    Pretrained on Places205 \cite{Zhou14}       & 24.8 & 54.4 \\
    Trained without Time Machine        & 38.7 & 68.1 \\
    Trained with Time Machine           & {\bf 68.5} & {\bf 90.8}
\end{tabular}
\end{center}
\vspace{-0.2cm}
\caption{ {\bf Time Machine importance.}
Recall of $f_{max}$ on Pitts30k-val
(AlexNet)
with vs without using Time Machine data
for training.
Training using Time Machine
is essential for generalization.
\vspace{-0.4cm}
}
\label{tab:timemachine}
\end{table}
}

\def\figTeaser{
\def\wTeas{0.45\linewidth}
\begin{figure}[t!]
\vspace{-0.3cm}
\begin{center}
    \subfloat[Mobile phone query]{
        \includegraphics[width=\wTeas]{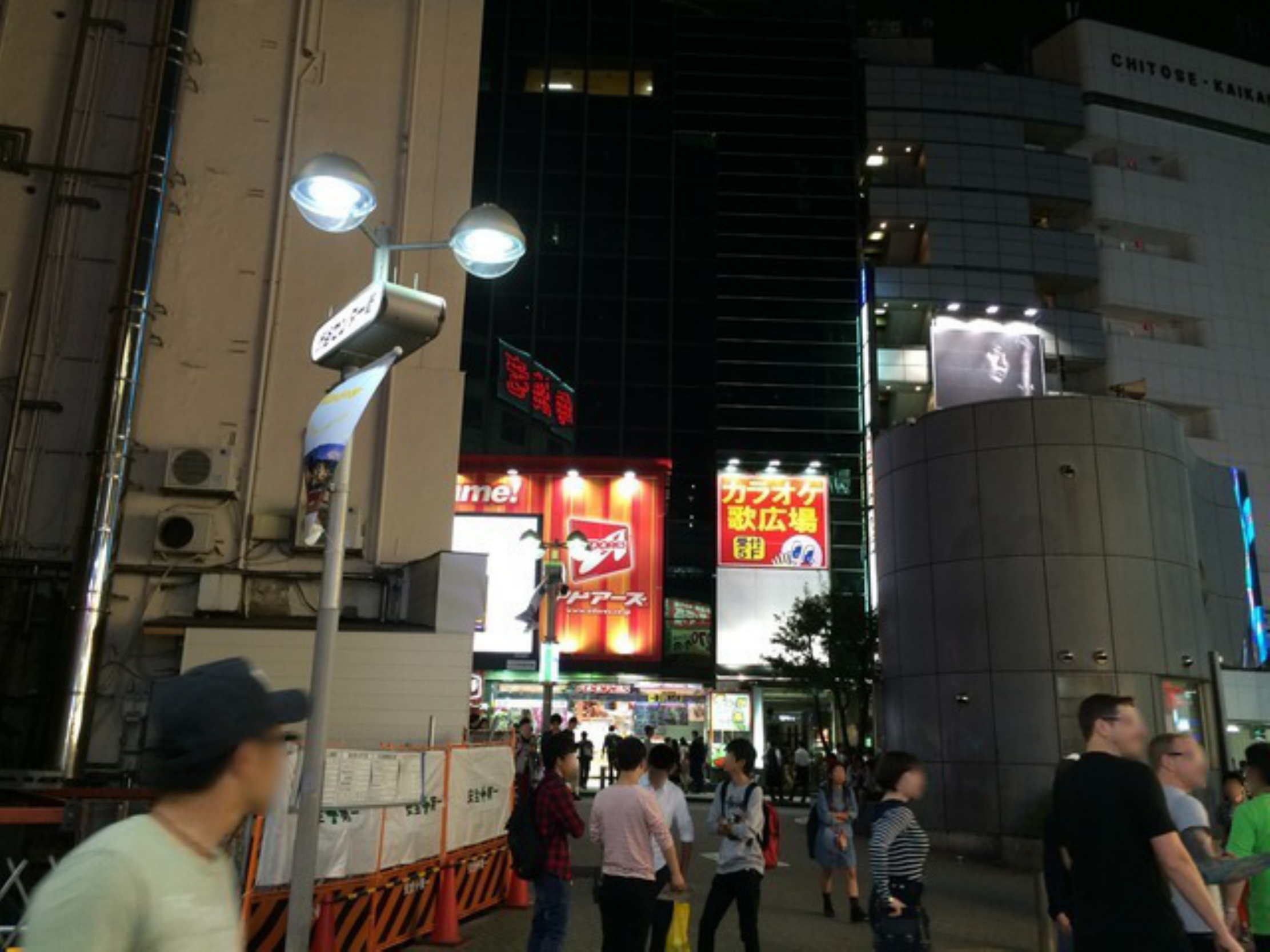}
    }
    \subfloat[Retrieved image of same place]{
        \includegraphics[width=\wTeas]{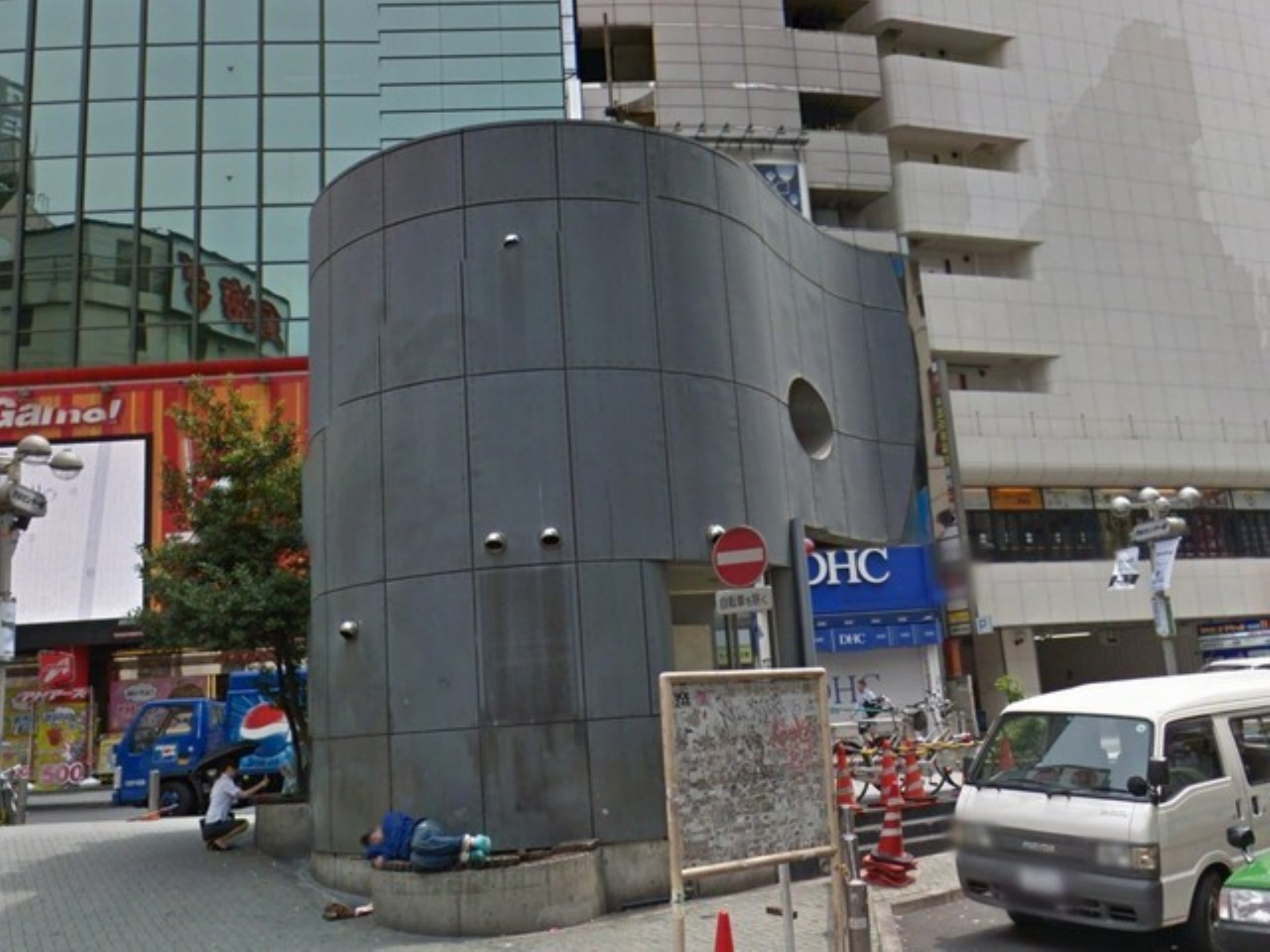}
    }
\end{center}
\vspace{-0.2cm}
    \caption{
Our trained NetVLAD descriptor correctly recognizes the location (b)
of the query photograph (a) despite the large amount of clutter (people, cars), changes in viewpoint and completely different illumination (night vs daytime).
{\bf Please see
\isArXiv{appendix \ref{sup:res}}{the appendix \cite{Arandjelovic15}} for more examples.}
}
\label{fig:teaser}
\vspace{-0.5cm}
\end{figure}
}

\def\figSupVLAD{
\begin{figure}[t!]
\vspace{-0.6cm}
\begin{center}
\includegraphics[width=0.5\linewidth]{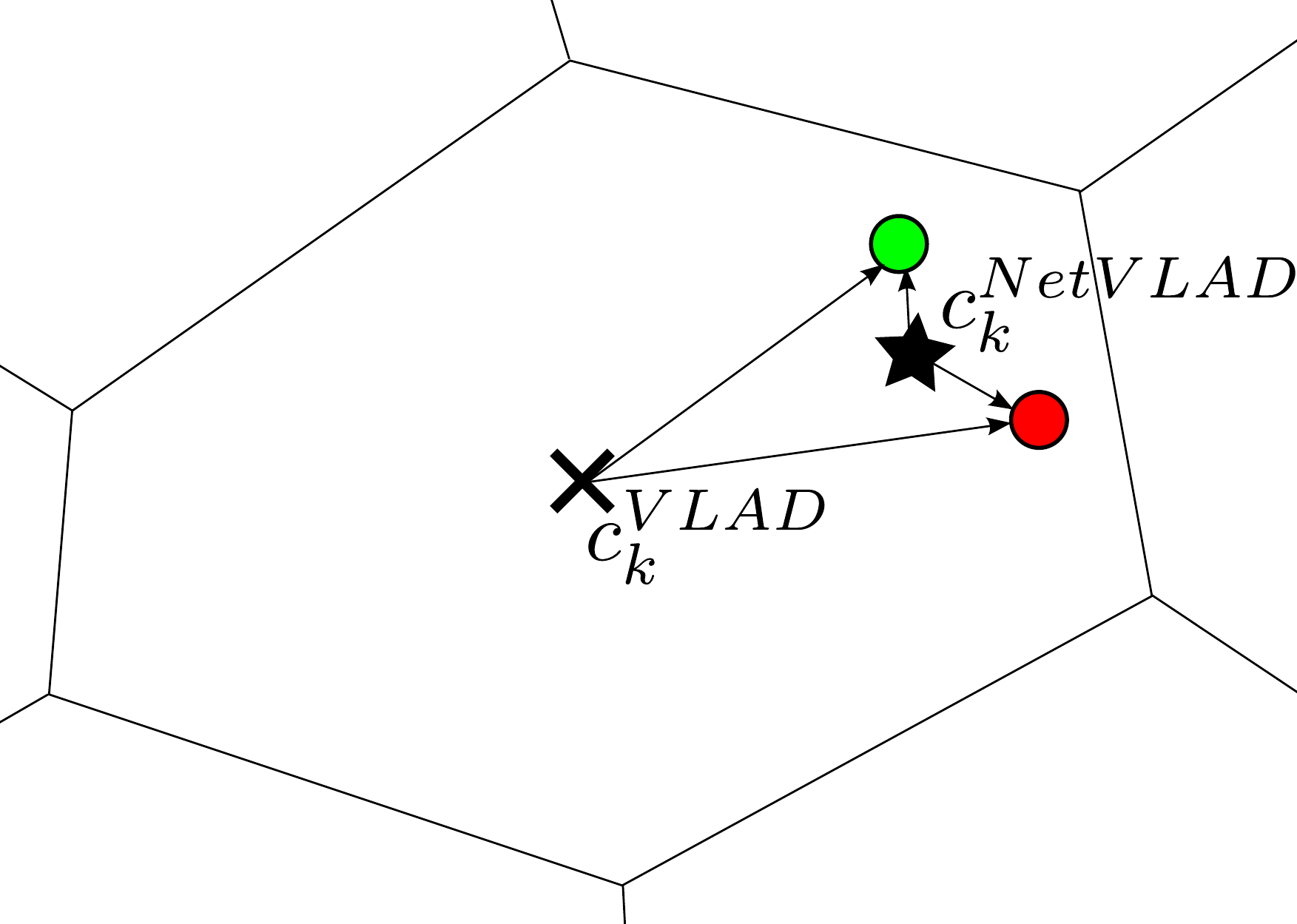}
\end{center}
\vspace{-0.2cm}
    \caption{
{\bf Benefits of supervised VLAD.}
Red and green circles are local descriptors from two different
images, assigned to the same cluster (Voronoi cell).
Under the VLAD encoding,
their contribution to the similarity score between the two images
is the scalar product (as final VLAD vectors are L2-normalized)
between the corresponding residuals, where a residual vector is computed
as the difference between the descriptor and the cluster's anchor point.
The anchor point $\vc c_k$ can be interpreted as the
origin of a new coordinate system local to the the specific cluster $k$.
In standard VLAD, the anchor is chosen as the cluster centre ($\times$)
in order to evenly distribute the residuals across the database.
However, in a supervised setting where the two descriptors are known to
belong to images which should not match, it is possible to learn a better
anchor ($\star$) which causes the scalar product between the new residuals
to be small.
\vspace{-0.2cm}
}
\label{fig:supVLAD}
\end{figure}
}

\def\figVLAD{
\begin{figure*}[t!]
\vspace{-0.2cm}
\begin{center}
\includegraphics[width=0.95\linewidth]{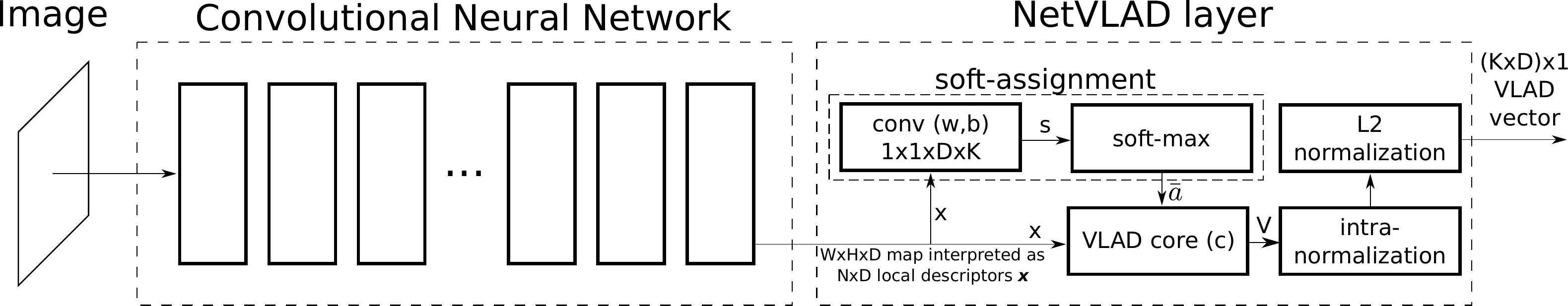}
\end{center}
\vspace{-0.2cm}
    \caption{
{\bf CNN architecture with the NetVLAD layer.}
The layer can be implemented using standard CNN layers
(convolutions, softmax, L2-normalization)
and one easy-to-implement aggregation layer
to perform aggregation in equation \eqref{eq:vladlayer}
(``VLAD core''),
joined up in a directed acyclic graph.
Parameters are shown in brackets.
\vspace{-0.4cm}
}
\label{fig:VLAD}
\end{figure*}
}

\def\figTM{
\def\wTM{0.31\linewidth}
\begin{figure}[t!]
\begin{center}
\hspace*{-0.4cm}
    \begin{tabular}{ccc}
\includegraphics[width=\wTM]{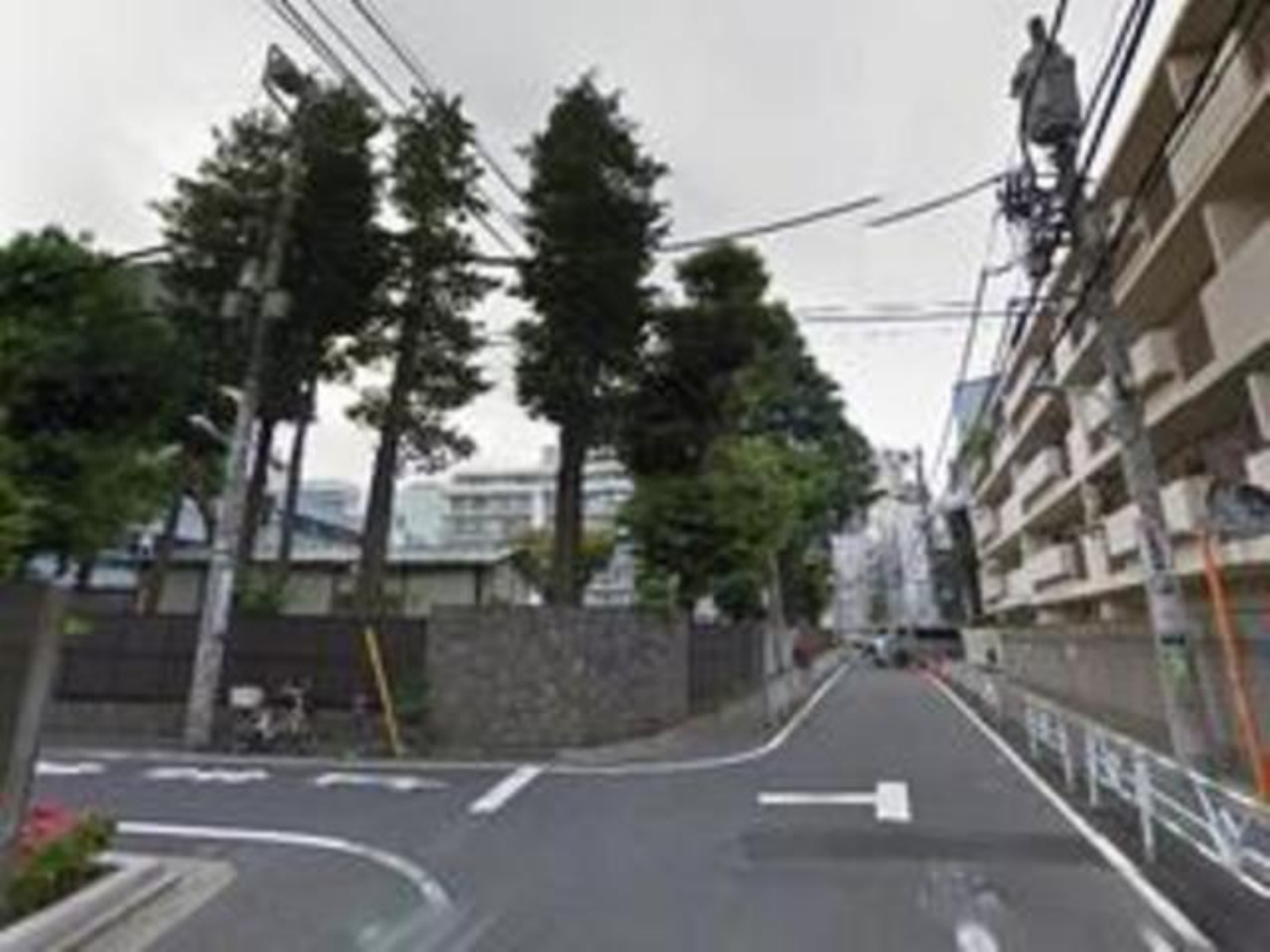}
& \includegraphics[width=\wTM]{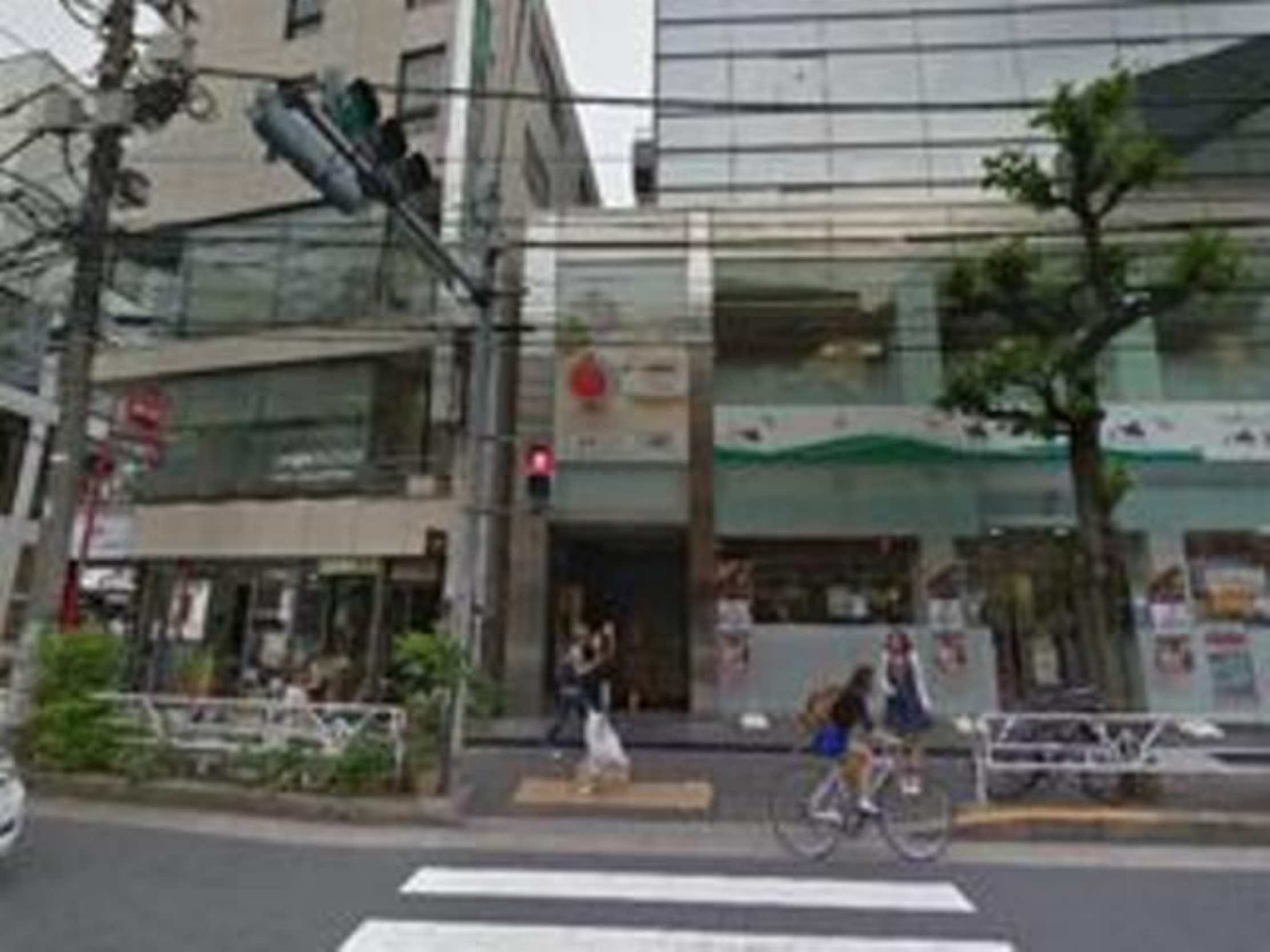}
& \includegraphics[width=\wTM]{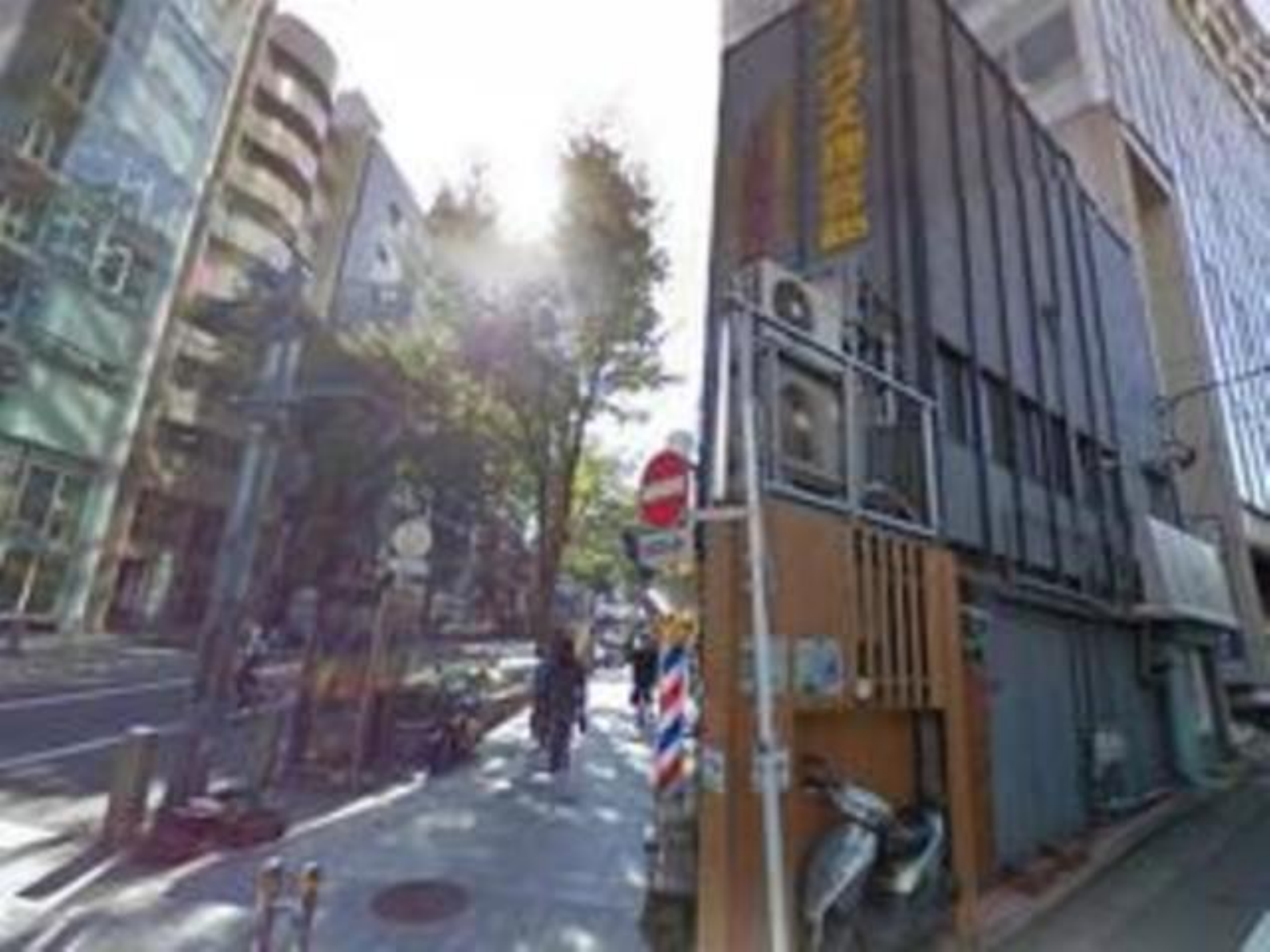}
    \\
\includegraphics[width=\wTM]{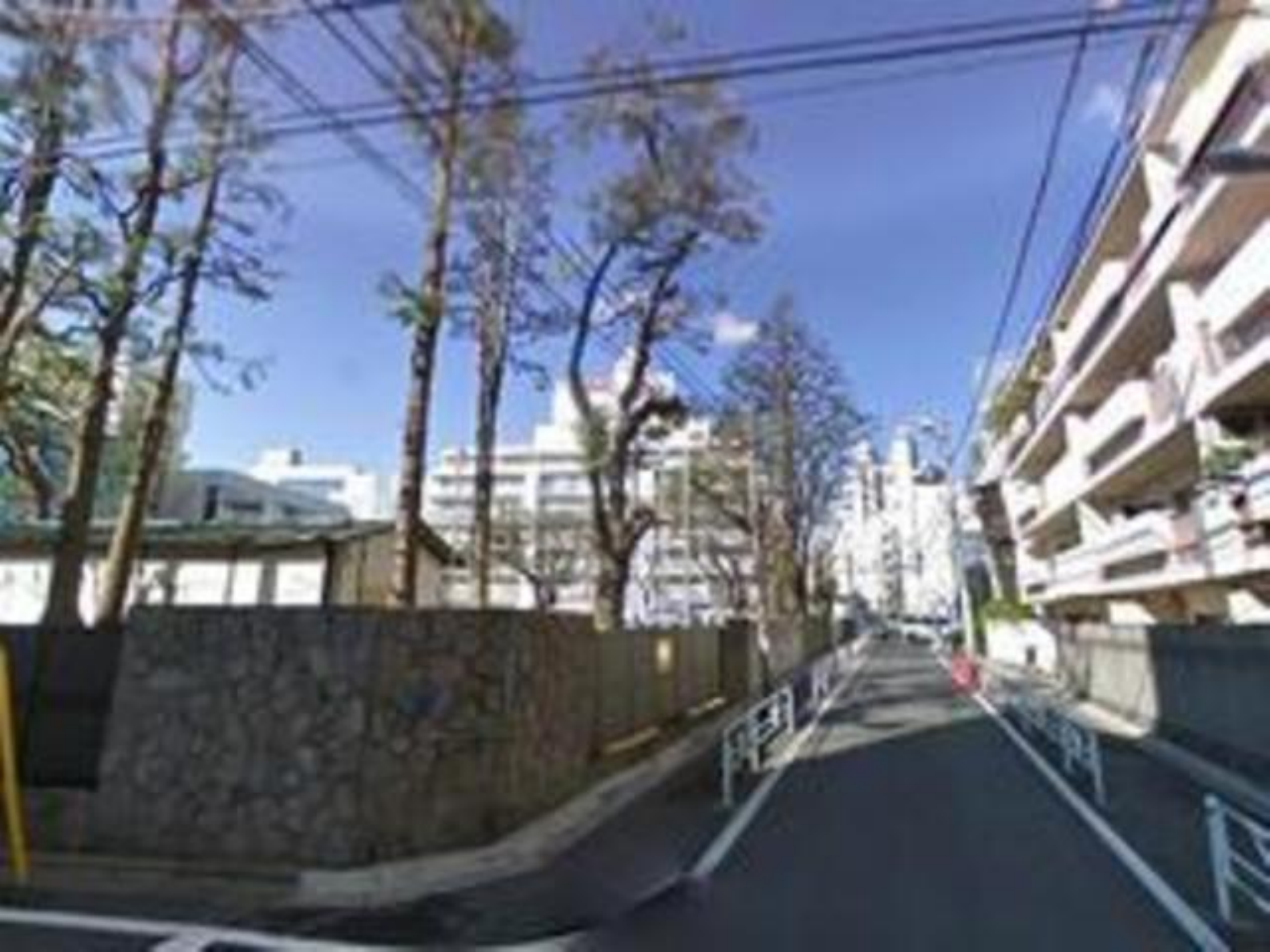}
& \includegraphics[width=\wTM]{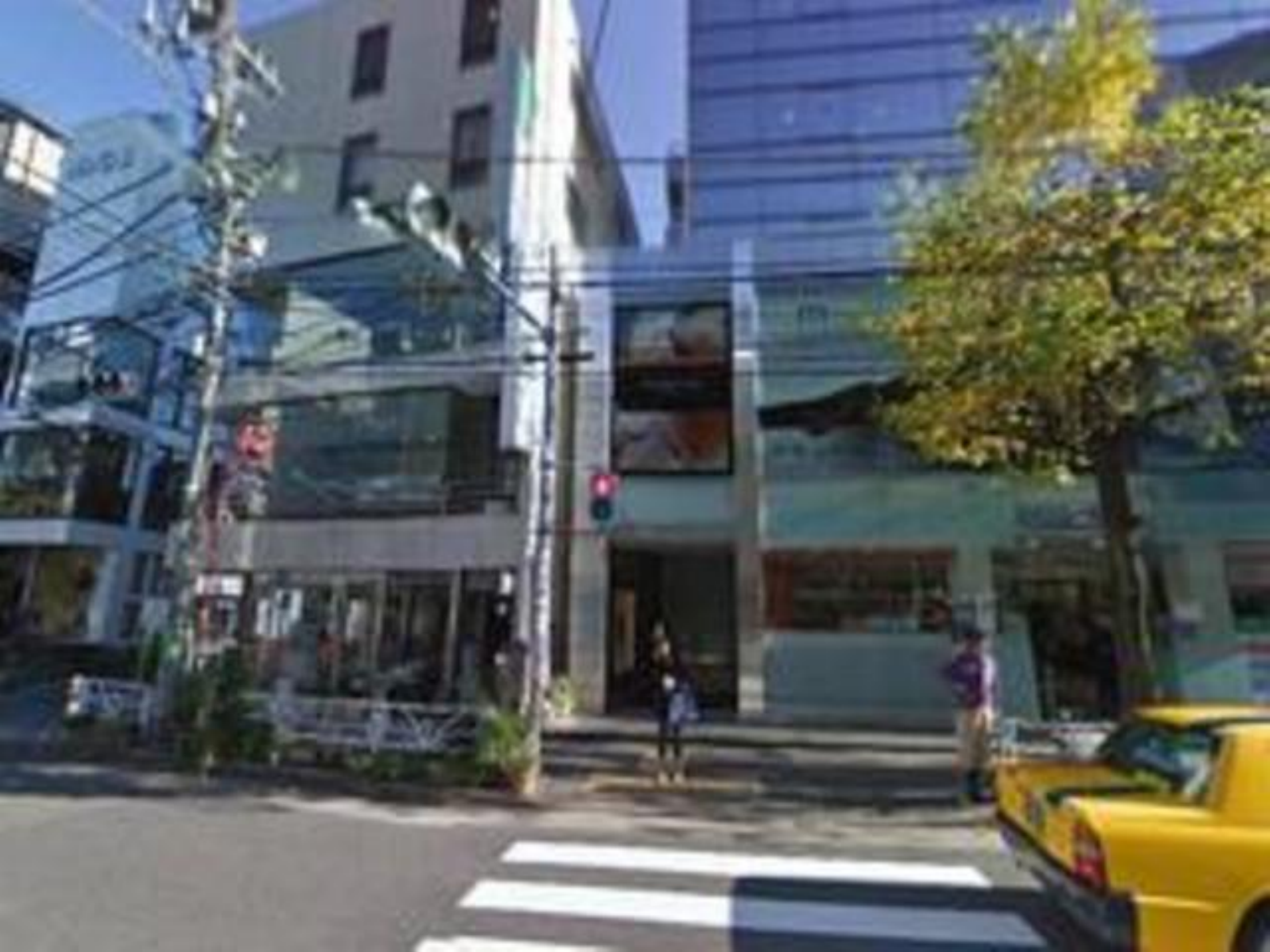}
& \includegraphics[width=\wTM]{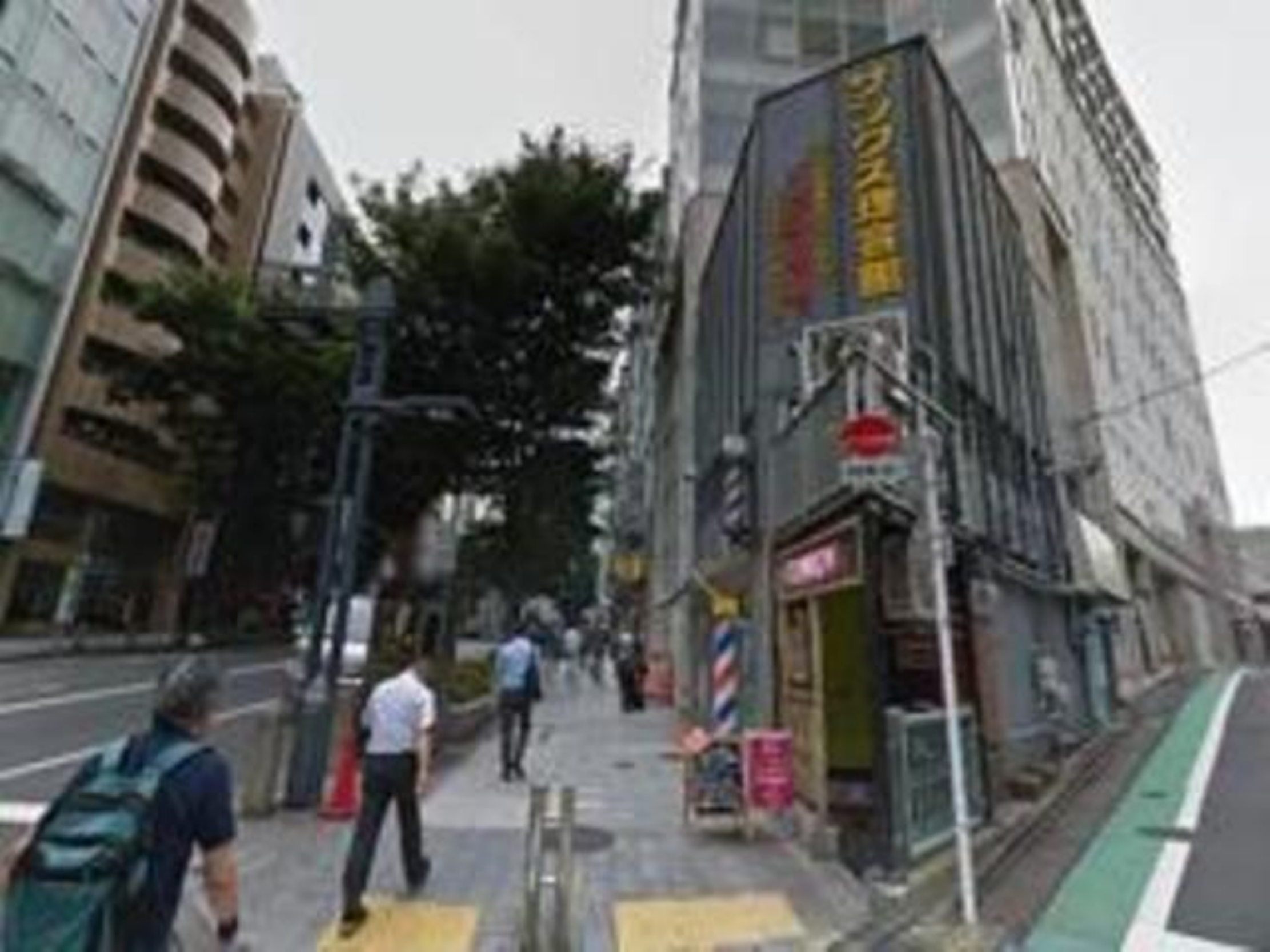}
    \\
\includegraphics[width=\wTM]{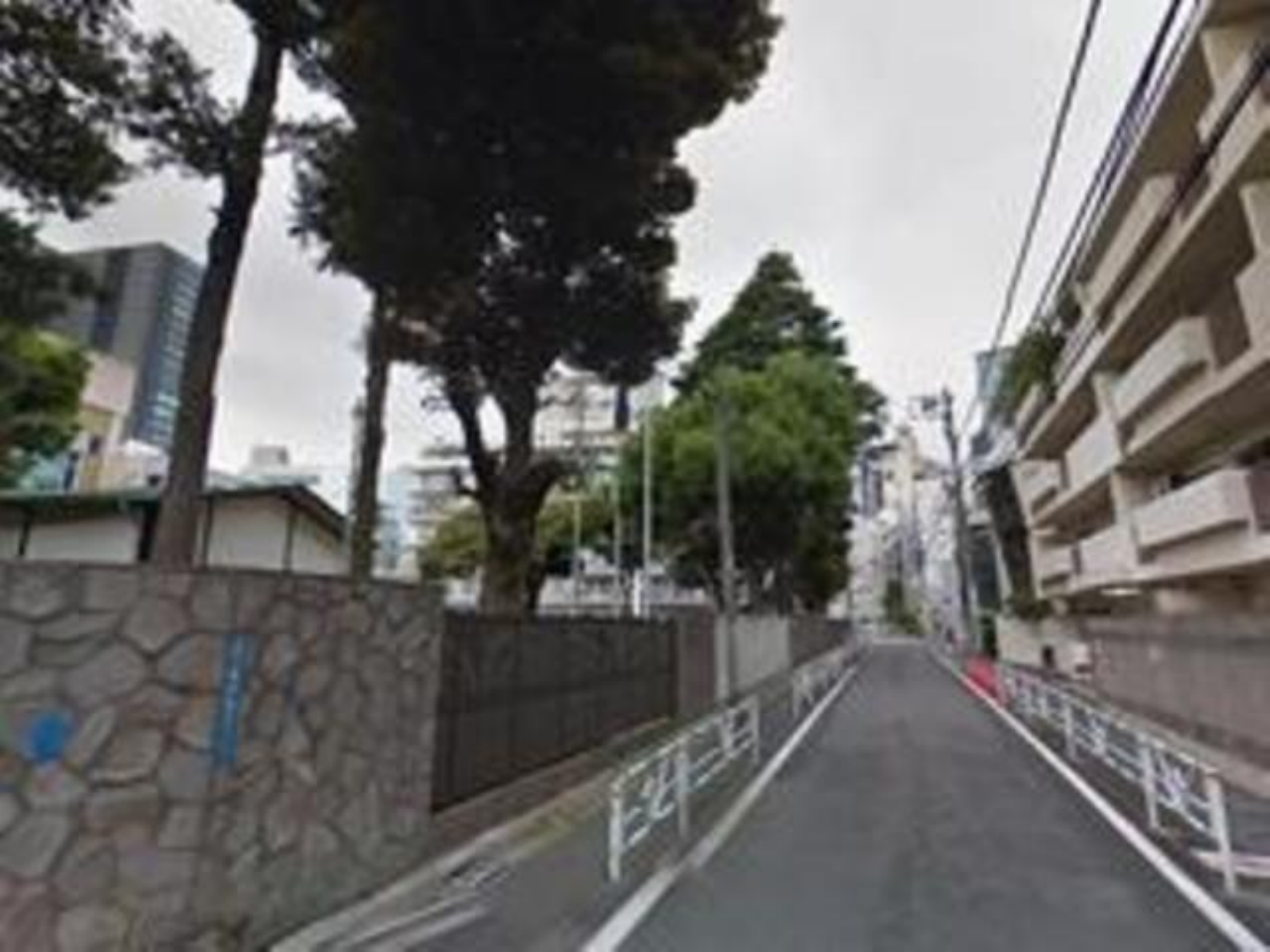}
& \includegraphics[width=\wTM]{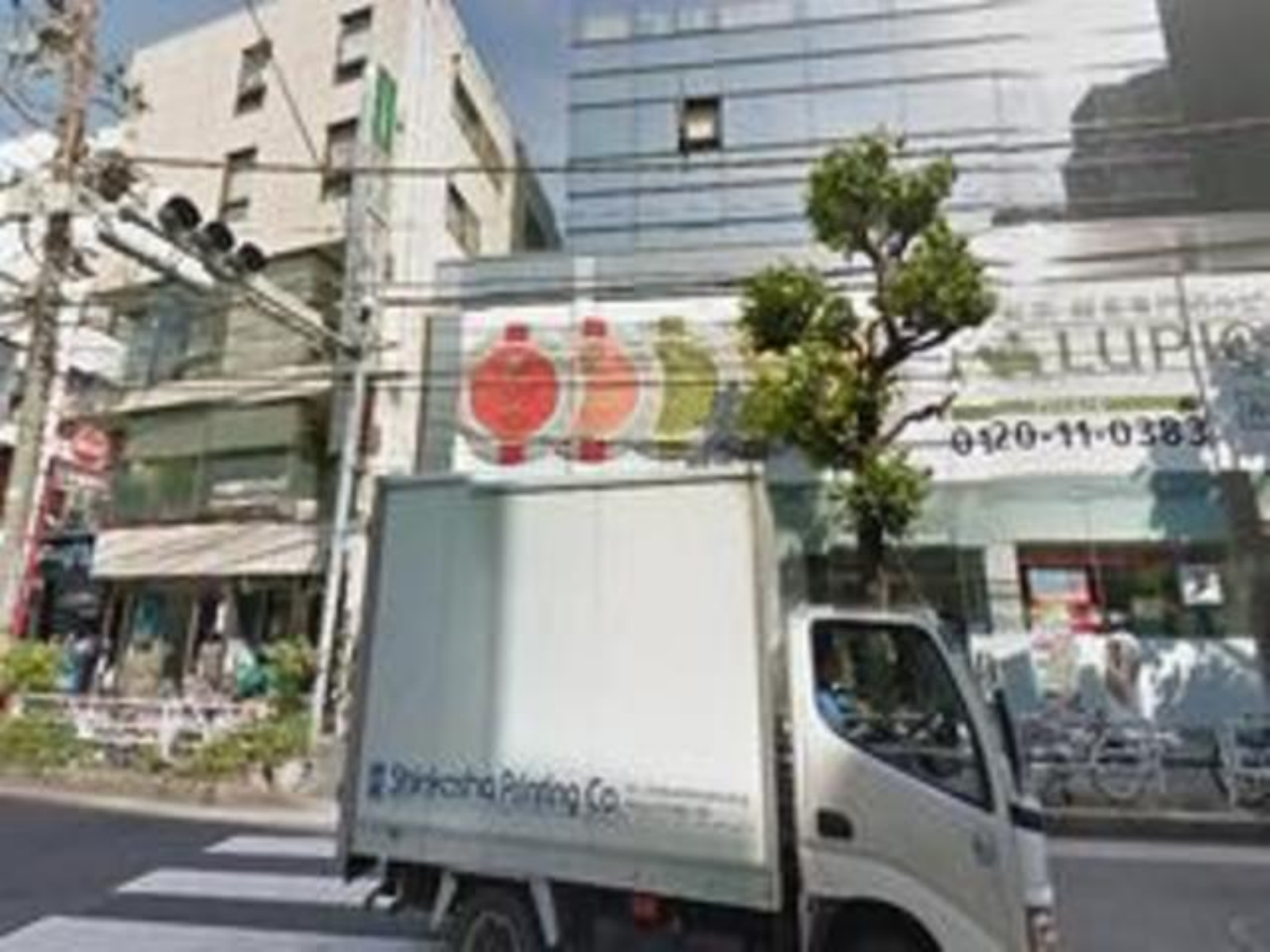}
& \includegraphics[width=\wTM]{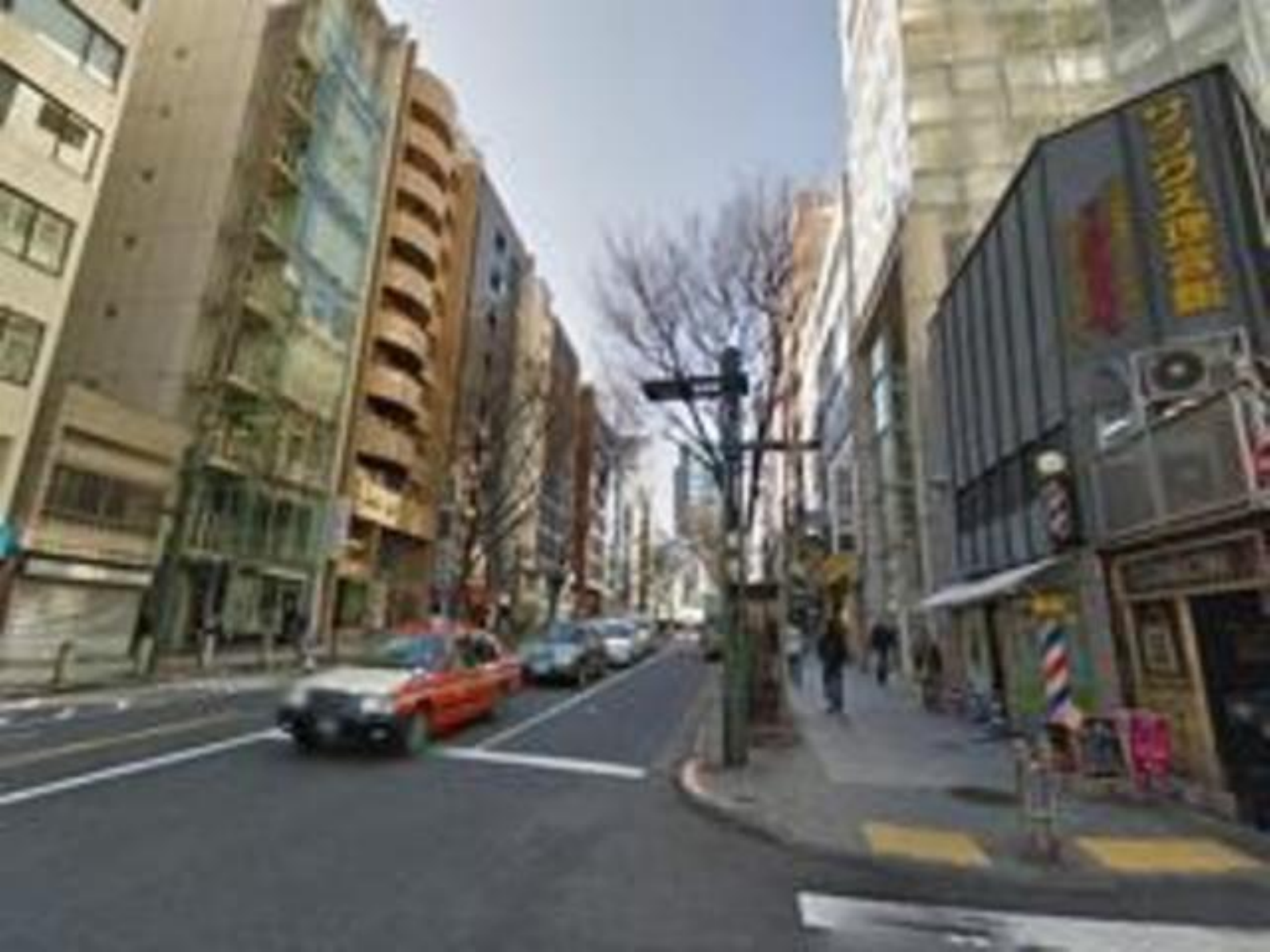}
\\
  {\small(a)}
& {\small(b)}
& {\small(c)}
    \end{tabular}
\end{center}
\vspace{-0.3cm}
    \caption{
{\bf Google Street View Time Machine examples.}
Each column shows perspective images generated from panoramas from nearby locations,
taken at different times.
A well designed method can use this source of imagery to learn
to be invariant to changes in viewpoint and lighting (a-c),
and to moderate occlusions (b).
It can also learn
to suppress confusing visual information such as clouds (a),
vehicles and people (b-c),
and to chose to either ignore vegetation or to learn a season-invariant
vegetation representation (a-c).
More examples are given in \isArXiv{appendix \ref{sup:TM}}{\cite{Arandjelovic15}}.
}
\label{fig:timemachine}
\vspace{-0.2cm}
\end{figure}
}

\def\figStoa{
\def\wStoa{0.22\linewidth}
\def\wStoaLeg{0.13\linewidth}
\begin{figure*}[t!]
\begin{center}
\vspace{-0.7cm}
\hspace*{-0.5cm}
\begin{tabular}{c@{}c@{}c@{}c@{}c}
    \parbox{2.7cm}{
    \vspace*{-4.5cm}
    \includegraphics[width=\linewidth, viewport=115 42 262 240, clip=true]{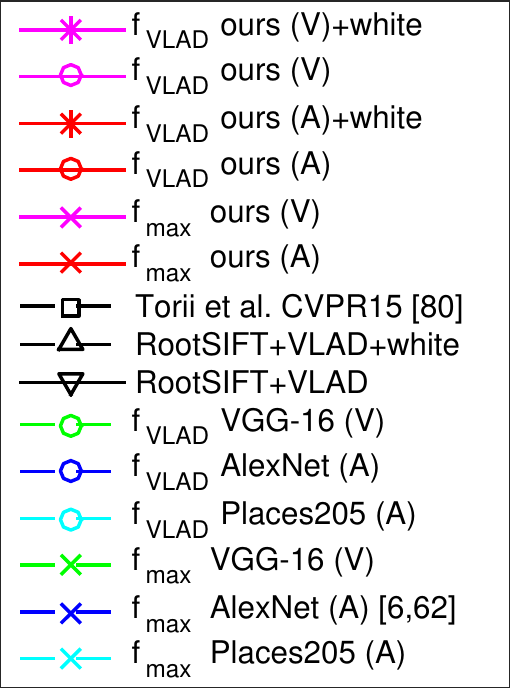}
    }
    &
    \subfloat[Pitts250k-test]{
        \includegraphics[width=\wStoa]{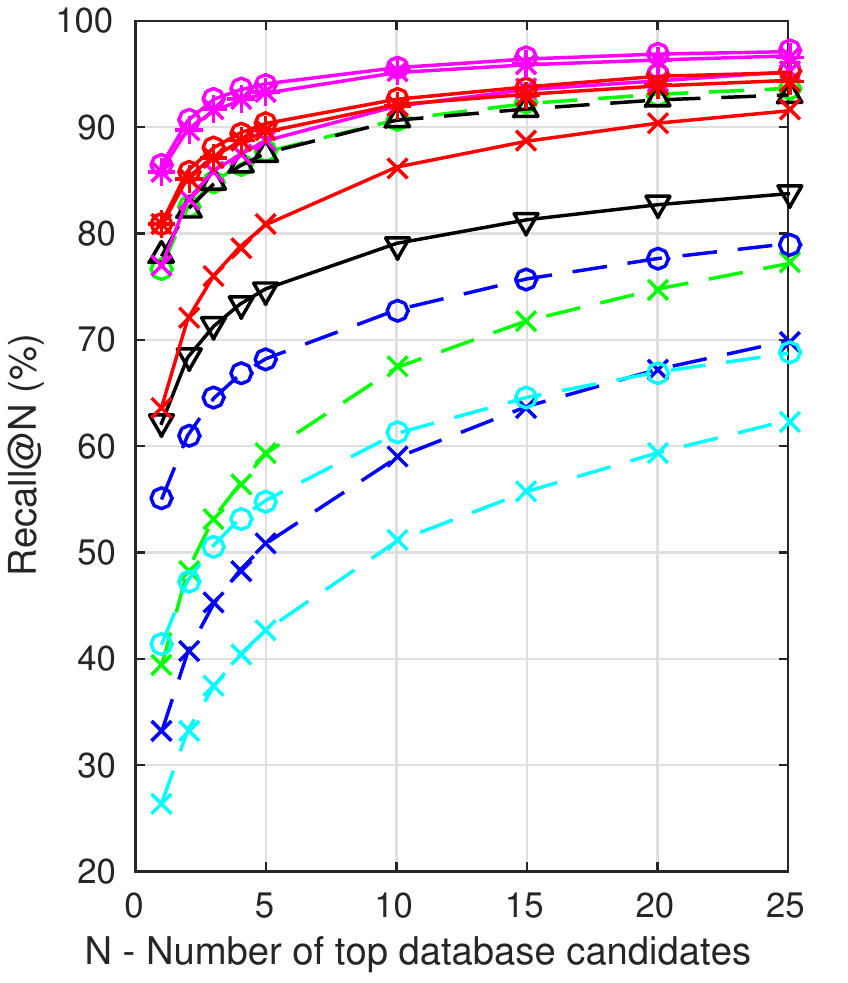}
    }
    &
    \subfloat[TokyoTM-val]{
        \includegraphics[width=\wStoa]{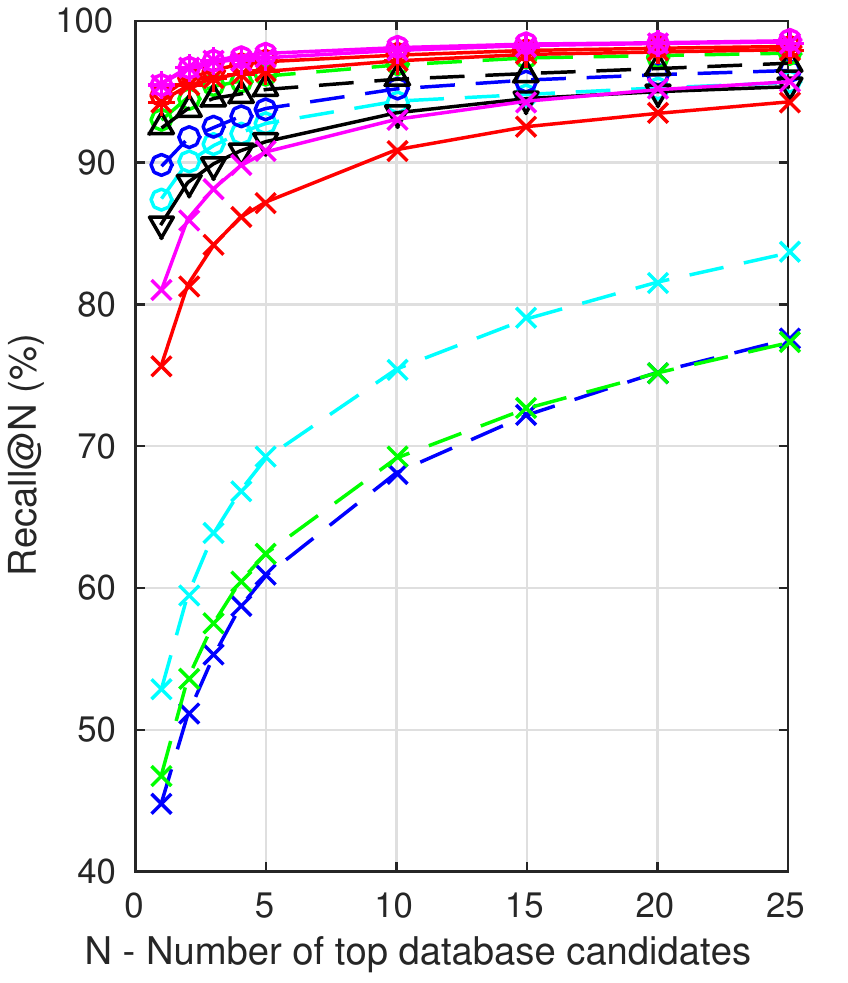}
    }
    &
    \subfloat[Tokyo 24/7 all queries]{
        \includegraphics[width=\wStoa]{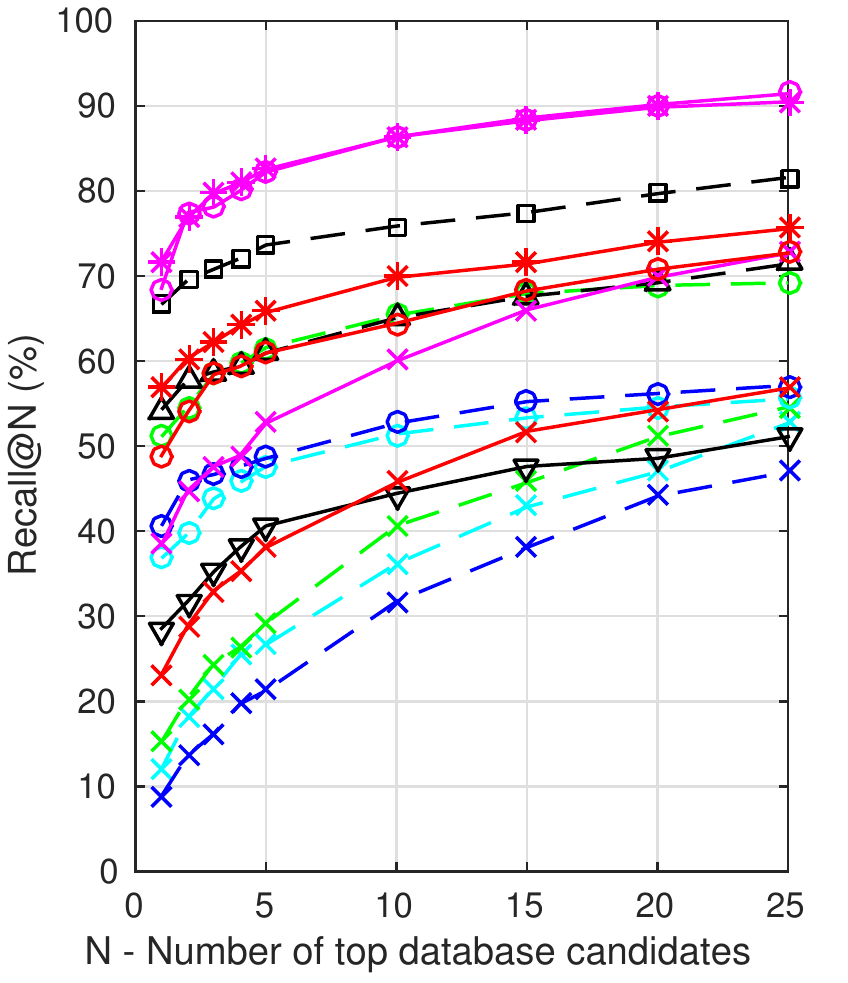}
    }
    &
    \subfloat[Tokyo 24/7 sunset/night]{
        \includegraphics[width=\wStoa]{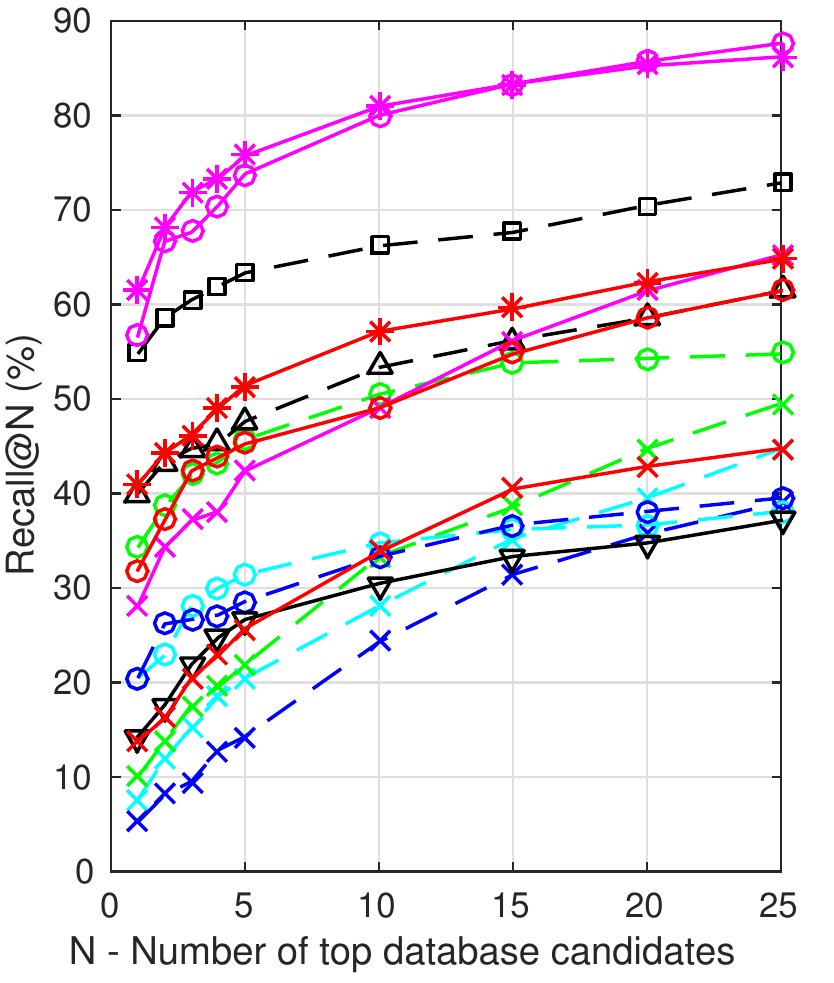}
    }
\end{tabular}
\end{center}
\vspace{-0.2cm}
    \caption{
\figStoaCaption
\cite{Torii15} only evaluated on Tokyo 24/7 as the method relies on depth data not available
in other datasets.
Additional results are shown in \isArXiv{appendix \ref{sup:res}}{the appendix \cite{Arandjelovic15}}.
\vspace{-0.3cm}
}
\label{fig:stoa}
\end{figure*}
}

\def\tabResLayers{
\begin{table}
\begin{center}
\small
\begin{tabular}{l||c@{~~~}c@{~~~}c|c@{~~~}c@{~~~}c}
    \multirow{2}{2cm}{Lowest trained layer}
    & \multicolumn{3}{c|}{$f_{max}$} & \multicolumn{3}{c}{$f_{VLAD}$} \\
    & r@1 & r@5 & r@10 & r@1 & r@5 & r@10 \\
    \hline
    none (off-the-shelf)         & 33.5 & 57.3 & 68.4 &         54.5 & 69.8 & 76.1  \\
    NetVLAD                      & ---  & ---  & ---  &         80.5 & 91.8 & 95.2 \\
    conv5                        & 63.8 & 83.8 & 89.0 &         84.1 & 94.6 & 95.5 \\
    conv4                        & 62.1 & 83.6 & 89.2 &         85.1 & 94.4 & 96.1 \\
    conv3                        & {\bf 69.8} & 86.7 & 90.3 &   {\bf 85.5} & 94.6 & 96.5 \\
    conv2                        & 69.1 & {\bf 87.6} & {\bf 91.5} &         84.5 & 94.6 & {\bf 96.6} \\
    conv1 (full)                 & 68.5 & 86.2 & 90.8 &         84.2 & {\bf 94.7} & 96.1
\end{tabular}
\end{center}
\vspace{-0.2cm}
\caption{ {\bf Partial training.}
Effects of performing backpropagation only down to a certain layer of AlexNet,
\eg `conv4' means that weights of layers from conv4 and above are learnt,
while weights of layers below conv4 are fixed to their pretrained state;
r@N signifies recall@N. Results are shown on the Pitts30k-val dataset.
\vspace{-0.3cm}
}
\label{tab:layers}
\end{table}
}

\def\vertText[#1]{\multirow{1}{*}[1.1cm]{\rotatebox[origin=c]{90}{\parbox{1.5cm}{\footnotesize \centering #1}}}}

\def\figFocus{
\def\wF{0.2\linewidth}
\begin{figure}[t!]
\begin{center}
\hspace*{-0.5cm}
\begin{tabular}{c@{~~~~}cccc}
    \vertText[Input \\ image] &
    \includegraphics[width=\wF]{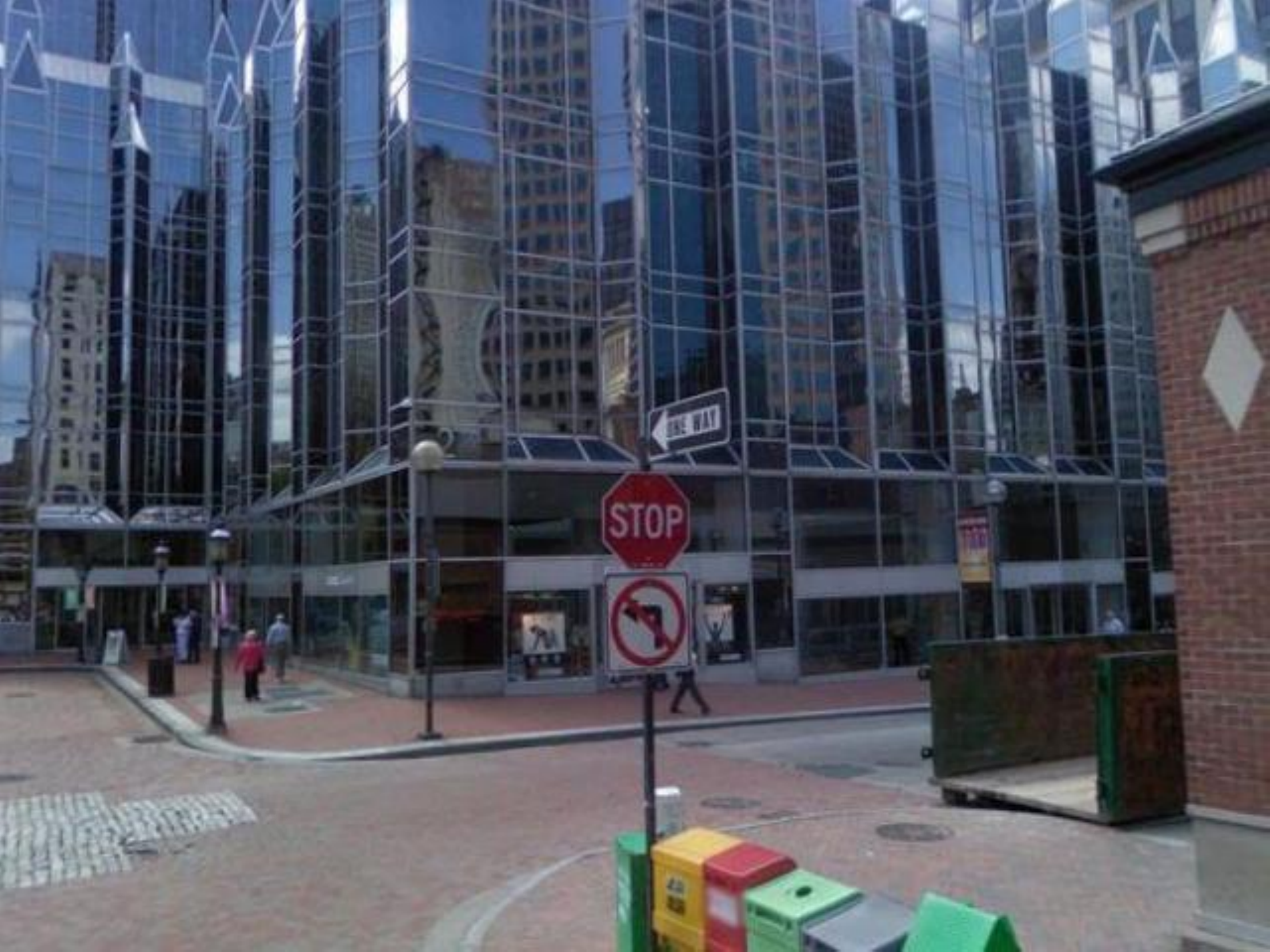} &
    \includegraphics[width=\wF]{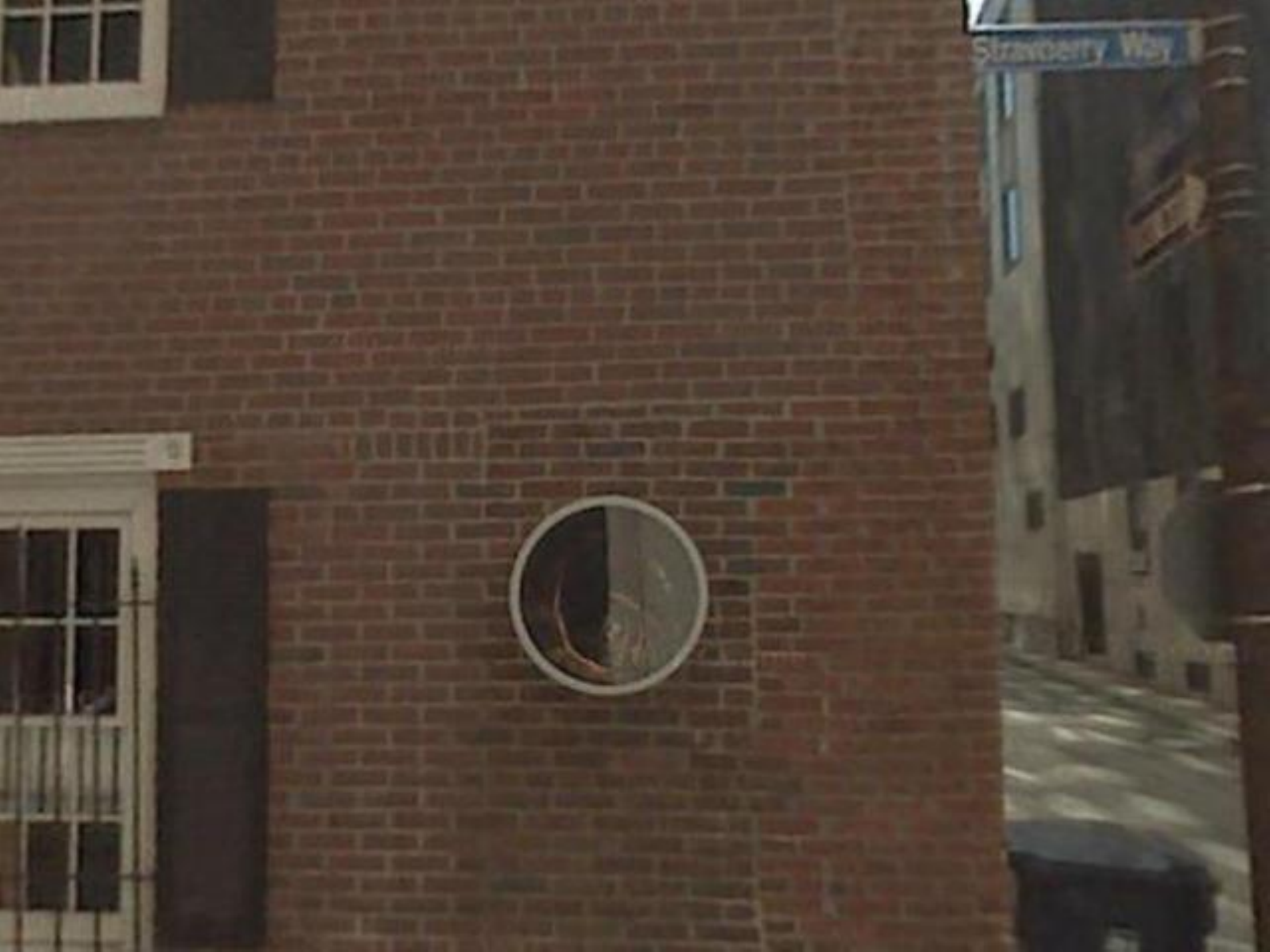} &
    \includegraphics[width=\wF]{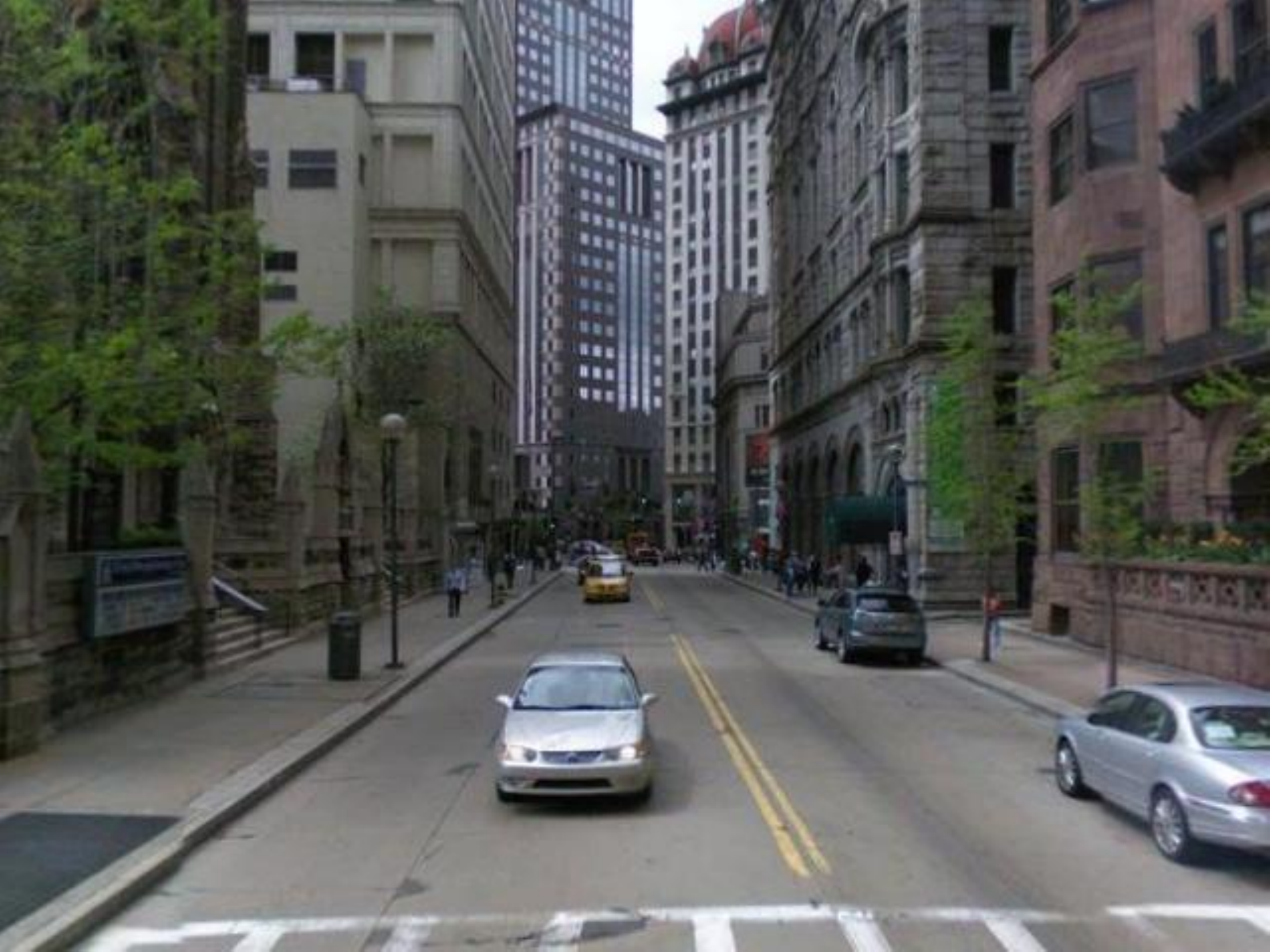} &
    \includegraphics[width=\wF]{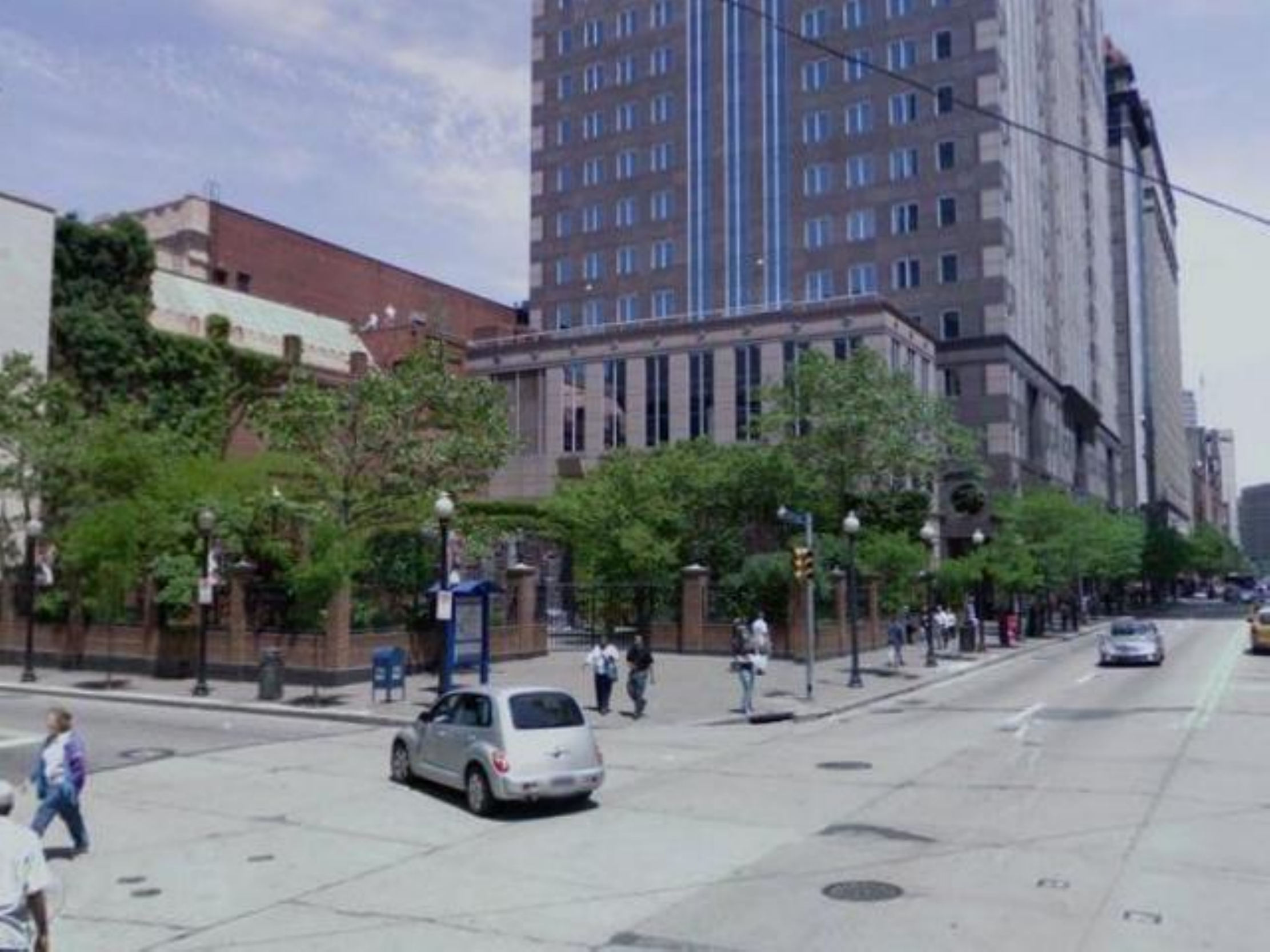}
    \\
    \vertText[AlexNet \\ ours] &
    \includegraphics[width=\wF]{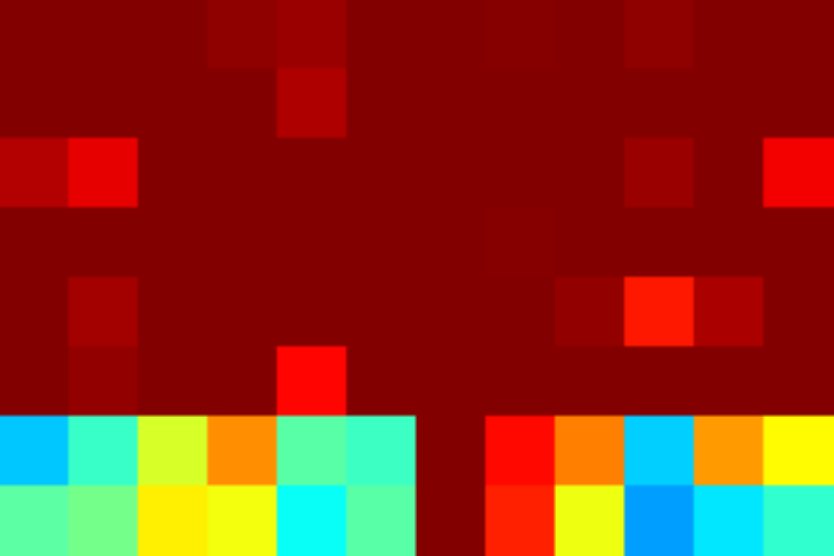} &
    \includegraphics[width=\wF]{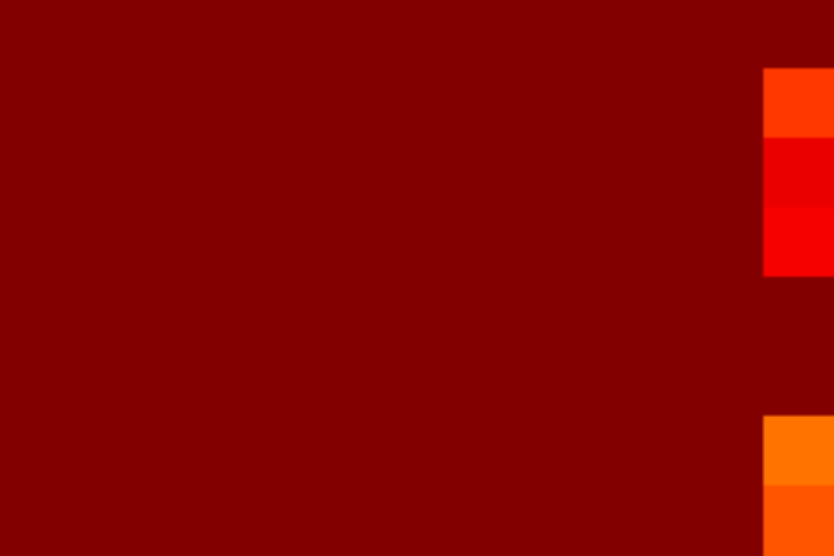} &
    \includegraphics[width=\wF]{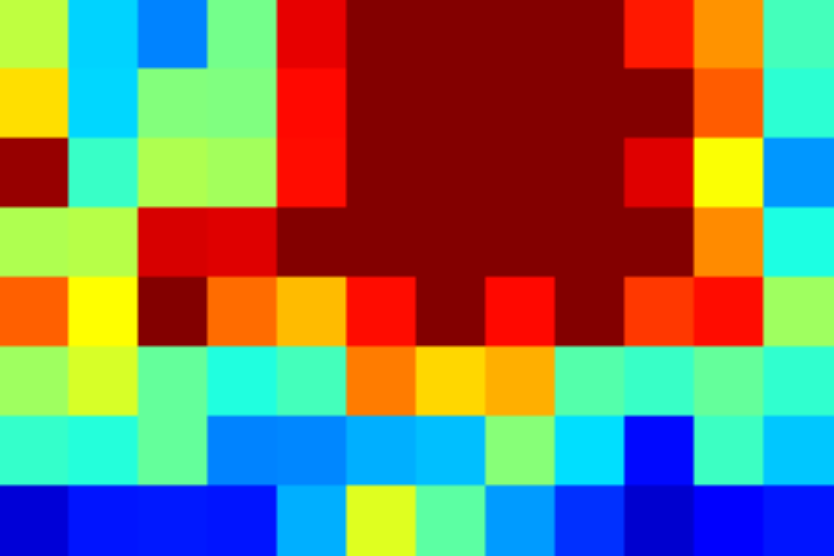} &
    \includegraphics[width=\wF]{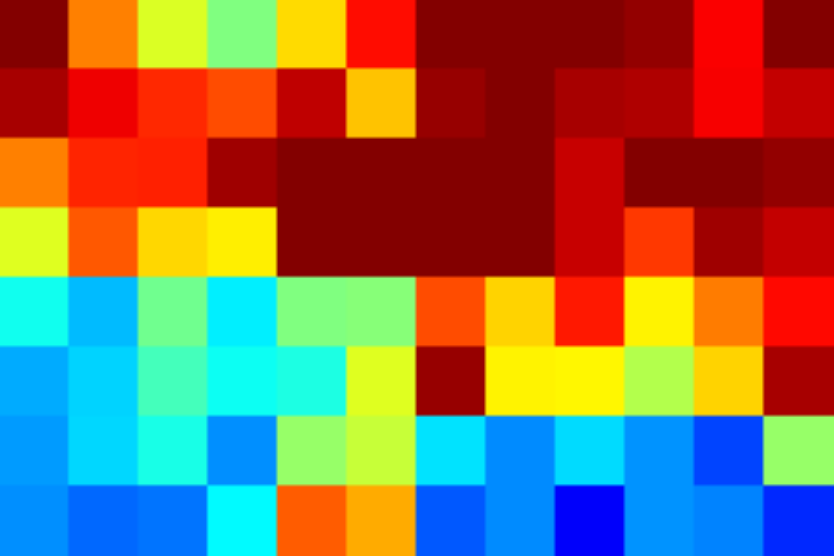}
    \\
    \vertText[AlexNet \\ off-shelf] &
    \includegraphics[width=\wF]{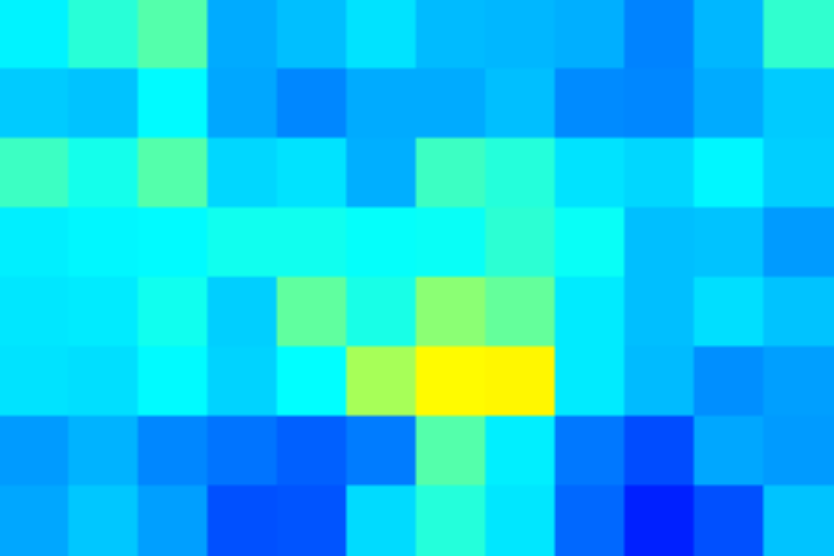} &
    \includegraphics[width=\wF]{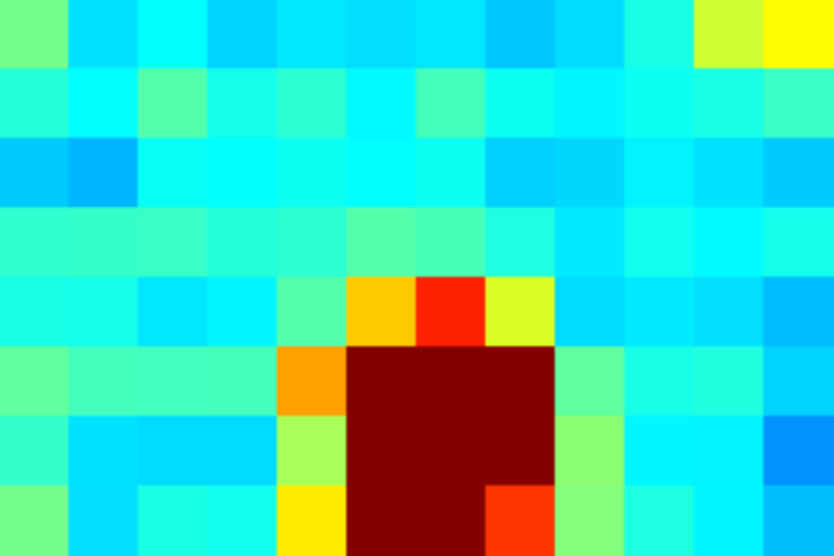} &
    \includegraphics[width=\wF]{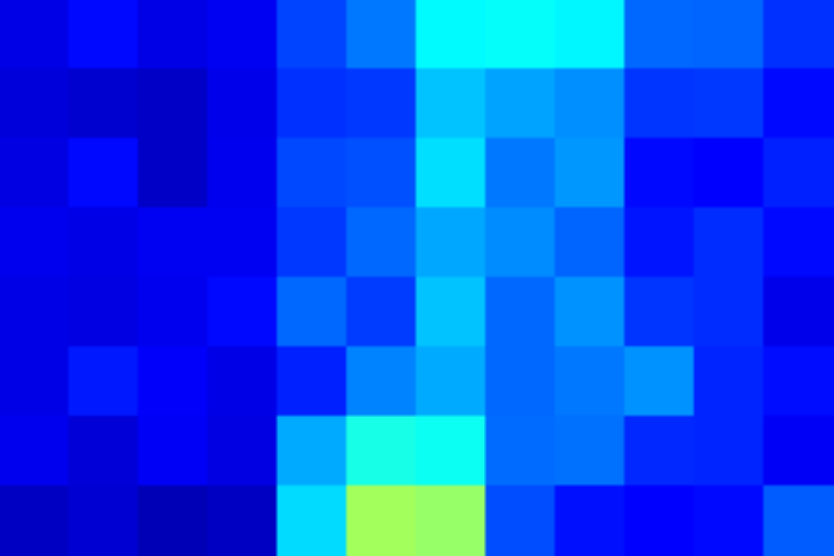} &
    \includegraphics[width=\wF]{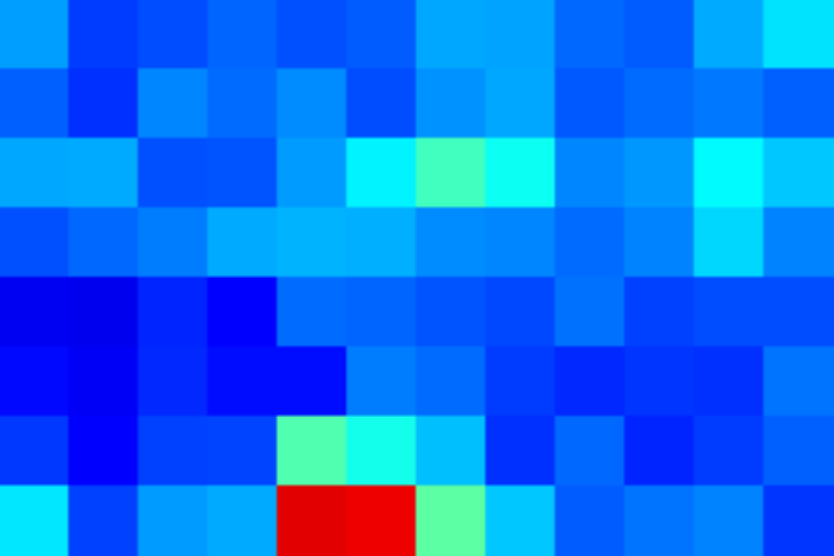}
    \\
    \vertText[Places205 \\ off-shelf] &
    \includegraphics[width=\wF]{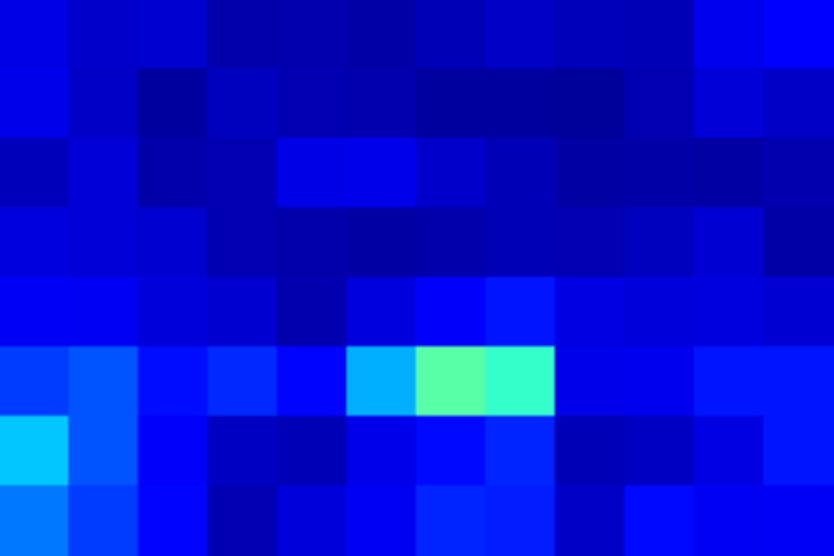} &
    \includegraphics[width=\wF]{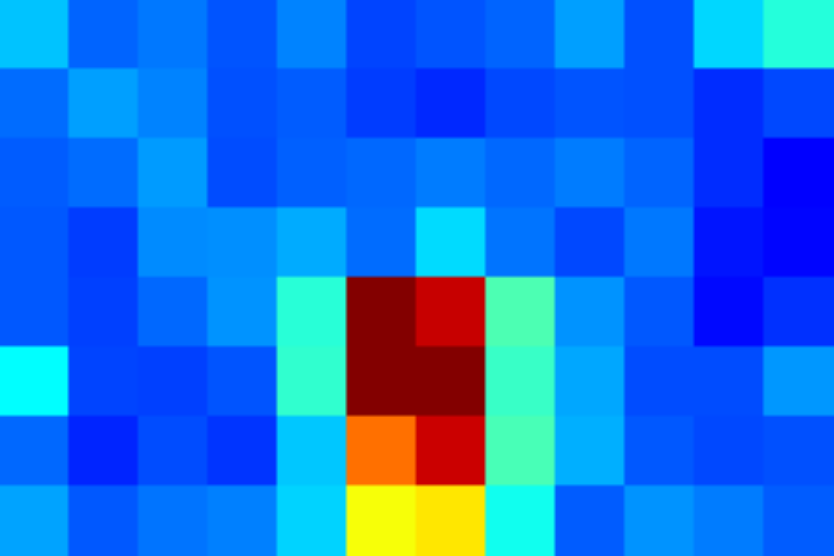} &
    \includegraphics[width=\wF]{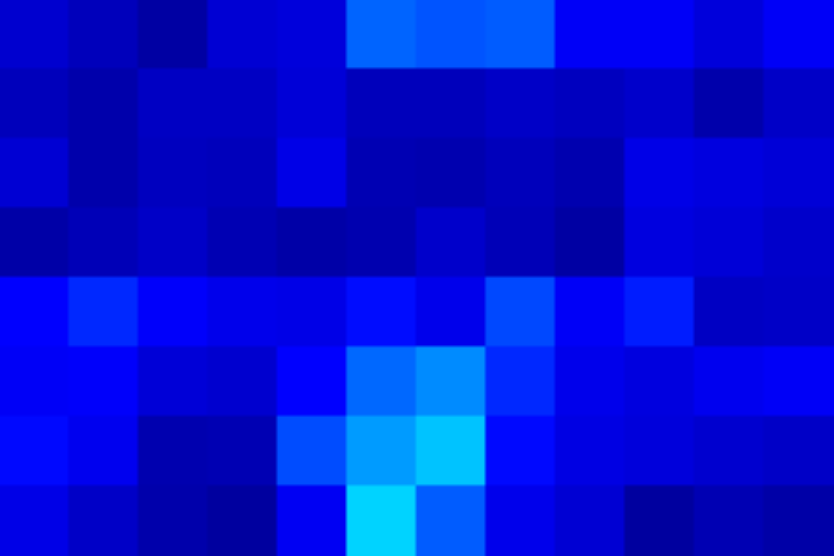} &
    \includegraphics[width=\wF]{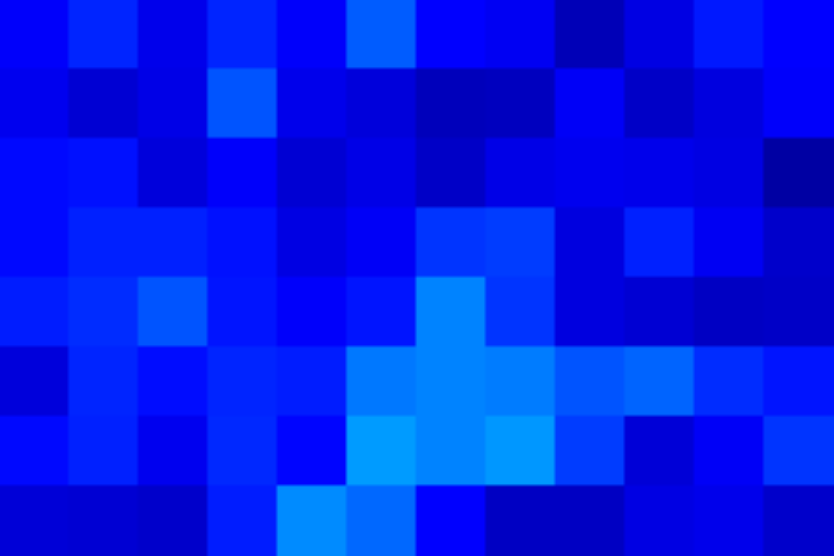}
\end{tabular}
\end{center}
\vspace{-0.3cm}
    \caption{
\figFocusCaption
Further examples are given in \isArXiv{appendix \ref{sup:res}}{the appendix \cite{Arandjelovic15}}.
\vspace{-0.4cm}
}
\label{fig:focus}
\end{figure}
}

\begin{abstract}

We tackle the problem of large scale visual place recognition, where the
task is to quickly and accurately recognize the location of a given query photograph. 
We present the following three principal contributions. First, 
we develop a convolutional neural network (CNN) architecture that is    
 trainable in an end-to-end manner directly for the place recognition task.
The main component of this architecture, NetVLAD, is a new generalized VLAD layer, inspired by the ``Vector of Locally Aggregated Descriptors" image representation commonly used in image retrieval.  The layer is readily pluggable into any CNN architecture and amenable to training via backpropagation.
Second, we develop a training procedure, based on a new weakly supervised ranking loss, to learn parameters of the architecture in an end-to-end manner from
images depicting the same places over time downloaded from Google Street View Time Machine.   
Finally, we show that the proposed architecture significantly outperforms
 non-learnt image representations and off-the-shelf CNN descriptors 
 on two challenging place recognition benchmarks,   
and improves over current state-of-the-art compact image representations on standard image retrieval benchmarks.

\end{abstract}

\section{Introduction}

Visual place recognition has received a significant amount of attention in the past years
both in computer vision~\cite{Schindler07,Knopp10,Chen11b,Sattler11,Sattler12,Gronat13,Cao13,Torii13,arandjelovic14a,Torii15,Sattler15} and robotics communities~\cite{Cummins08,Cummins09,McManus14,Maddern12,Sunderhauf15} motivated by, \eg, applications in autonomous driving~\cite{McManus14}, augmented reality~\cite{Middelberg14} or geo-localizing archival imagery~\cite{Aubry14}.

The place recognition problem, however, still remains extremely challenging. How can we recognize the same street-corner in the entire city or on the scale of the entire country despite the fact it can be captured in different illuminations or change its appearance over time? %
The fundamental scientific question is what is the appropriate representation of a place that is rich enough to distinguish similarly looking places yet compact to represent entire cities or countries.

The place recognition problem has been traditionally cast as an instance retrieval task, %
 where the query image location is estimated using the locations of the most visually similar images obtained by
querying a large geotagged database~\cite{Schindler07,Chen11b,Knopp10,Torii13,Torii15,arandjelovic14a}.
Each database image is represented using local invariant features~\cite{Tuytelaars08} such as SIFT~\cite{Lowe04} that are aggregated into a single vector representation for the entire image such as bag-of-visual-words~\cite{Sivic03,Philbin07}, VLAD~\cite{arandjelovic13,Jegou10} or Fisher vector~\cite{Perronnin10,Jegou12a}. The resulting representation is then usually compressed and efficiently indexed~\cite{Sivic03,Jegou11}. %
The image database can be further augmented by 3D structure that enables recovery of accurate camera pose~\cite{Li12a,Sattler15,Sattler11}.

\figTeaser

In the last few years convolutional neural networks (CNNs)~\cite{Lecun89,Lecun98} have emerged as powerful image representations for various category-level recognition tasks such as object classification~\cite{Krizhevsky12,Oquab14,Simonyan15,Szegedy15}, scene recognition~\cite{Zhou14} or object detection~\cite{Girshick14}. The basic principles of CNNs are known from 80's~\cite{Lecun89,Lecun98} and the recent successes are a combination of advances in GPU-based computation power together with large labelled image datasets~\cite{Krizhevsky12}.
While it has been shown that the trained representations are, to some extent, transferable between recognition tasks~\cite{Donahue13,Girshick14,Oquab14,Sermanet13,Zeiler14}, a direct application of CNN representations trained for object classification~\cite{Krizhevsky12} as black-box descriptor extractors has so far yielded limited improvements in performance on instance-level recognition tasks
\cite{Azizpour14,Babenko15,Gong14,Razavian14,Razavian15}. In this work we investigate whether this gap in performance can be bridged by CNN representations developed and trained directly for place recognition. This requires addressing the following three main challenges.
First, what is a good CNN architecture for place recognition?
Second, how to gather sufficient amount of annotated data for the training?
Third, how can we train the developed architecture in an end-to-end manner tailored for the place recognition task?
To address these challenges we bring the following three innovations. %

First, building on the lessons learnt from the current well performing hand-engineered object retrieval and place recognition pipelines~\cite{Arandjelovic12,arandjelovic13,Jegou12,Torii15}  we develop a convolutional neural network architecture for place recognition that aggregates mid-level (conv5) convolutional features extracted from the entire image
into a compact single vector representation amenable to efficient indexing. To achieve this, we design a new trainable generalized VLAD layer, NetVLAD, inspired by the Vector of Locally Aggregated Descriptors (VLAD) representation~\cite{Jegou10} that has shown excellent performance in image retrieval and place recognition. The layer is readily pluggable into any CNN architecture and amenable to training via backpropagation.
The resulting aggregated representation is then compressed using
Principal Component Analysis (PCA) to obtain the final compact descriptor of the image.

Second, to train the architecture for place recognition, we gather a large dataset of multiple panoramic images depicting the same place from different viewpoints over time from the Google Street View Time Machine. Such data is available for vast areas of the world, but provides only weak form of supervision: we know the two panoramas are captured at approximately similar positions based on their (noisy) GPS but we don't know which parts of the panoramas depict the same parts of the scene.

Third, we develop a learning procedure for place recognition that learns parameters of the architecture in an end-to-end manner tailored for the place recognition task from the weakly labelled Time Machine imagery. The resulting representation is robust to 
changes in viewpoint and lighting conditions, while simultaneously learns to focus on the relevant parts of the image such as the building fa\c{c}ades and the skyline, while ignoring confusing elements such as cars and people that may occur at many different places.

We show that the proposed architecture significantly outperforms non-learnt image representations and off-the-shelf CNN descriptors on two challenging place recognition benchmarks, and improves over current state-of-the-art compact image representations on standard image retrieval benchmarks.

\subsection{Related work}
While there have been many improvements in designing better
image retrieval~\cite{Arandjelovic12,arandjelovic13,Chum07b,Chum11,Delhumeau13,Jegou08,Jegou09a,Jegou10,Jegou12,Jegou14,Mikulik10,Perronnin07,Perronnin10,Philbin07,Philbin08,Simonyan12,Tolias13a,Tolias14,Turcot09} and place recognition~\cite{arandjelovic14a,Chen11b,Cummins08,Cummins09,Gronat13,Cao13,Knopp10,McManus14,Maddern12,Sattler11,Sattler12,Sattler15,Sunderhauf15,Torii13,Torii15} systems, not many works have performed
learning for these tasks.
All relevant learning-based approaches fall into one or both of the following
two categories:
(i) learning for an auxiliary task (\eg some form of distinctiveness of local features~\cite{arandjelovic14a,Cummins08,Jegou07,Knopp10,Qin11,Qin13,Zepeda15}), and (ii) learning on top of shallow hand-engineered descriptors that cannot be fine-tuned for the target task~\cite{Arandjelovic12,Gronat13,Cao13,Knopp10,Qin14a}. Both of these are in spirit opposite to the core idea behind
deep learning that has provided a major boost in performance in various
recognition tasks: end-to-end learning. We will indeed show in
section \ref{sec:res} that training representations directly for the end-task,
place recognition, is crucial for obtaining good performance.

Numerous works concentrate on learning better local descriptors or
metrics to compare them
\cite{Winder09,Philbin10,Makadia10,Mikulik10,Simonyan12,Qin14,SimoSerra14,Paulin15},
but even though some of them show results on image retrieval,
the descriptors are learnt on the task of matching local image patches,
and not directly with image retrieval in mind.
Some of them also make use of hand-engineered features to bootstrap the learning,
\ie to provide noisy training data \cite{Philbin10,Makadia10,Mikulik10,Simonyan12,Paulin15}.

Several works have investigated using CNN-based features for
image retrieval. These include treating activations from certain layers
directly as descriptors by concatenating them \cite{Babenko14,Razavian14},
or by pooling \cite{Azizpour14,Gong14,Babenko15}.
However, none of these works actually train the CNNs for the task at hand,
but use CNNs as black-box descriptor extractors.
One exception is the work of Babenko \etal \cite{Babenko14} in which
the network is fine-tuned on an auxiliary task of classifying 700 landmarks. However,
again the network is not trained directly on the target retrieval task.

Finally, recently \cite{Kendall15} and \cite{Lin15} performed end-to-end
learning for different but related tasks of ground-to-aerial matching \cite{Lin15} and
camera pose estimation \cite{Kendall15}.

\section{Method overview}
\label{sec:overview}

Building on the success of current place recognition systems (\eg~\cite{Schindler07,Knopp10,Chen11b,Sattler11,Sattler12,Torii13,arandjelovic14a,Torii15,Sattler15}), 
we cast place recognition as image retrieval. 
The query image with unknown location is used to visually search a large
geotagged image database, and the locations of top ranked images are used
as suggestions for the location of the query. This is generally done by designing
a function $f$ which acts as the ``image representation extractor'',
such that given an image $I_i$ it produces a fixed size vector $f(I_i)$.
The function is used to extract the representations for the entire
database $\{I_i\}$, which can be done offline,
and to extract the query image representation $f(q)$, done online.
At test time, the visual search is performed by finding the nearest database
image to the query,
either exactly or through fast approximate nearest neighbour search,
by sorting images based on the Euclidean distance $d(q,I_i)$ between
$f(q)$ and $f(I_i)$.

While previous works have mainly used hand-engineered image representations
(\eg $f(I)$ corresponds to extracting SIFT descriptors \cite{Lowe04},
followed by pooling into a bag-of-words vector \cite{Sivic03}
or a VLAD vector \cite{Jegou10}),
here we propose to learn the representation $f(I)$ in an end-to-end manner,
directly optimized for the task of place recognition.
The representation is parametrized with a set of parameters $\theta$
and we emphasize this fact by referring to it as $f_\theta(I)$.
It follows that the Euclidean distance
$d_\theta(I_i,I_j)=\norm{f_\theta(I_i)-f_\theta(I_j)}$
also depends on the same parameters.
An alternative setup would be to learn the distance function itself, but here
we choose to fix the distance function to be Euclidean distance, and to pose
our problem as the search for the explicit feature map $f_\theta$ which works well
under the Euclidean distance.

In section \ref{sec:rep} we describe the proposed representation $f_\theta$
based on a new deep convolutional neural network architecture inspired by the
 compact aggregated image descriptors for instance retrieval. 
In section~\ref{sec:learning} we describe a method to learn the parameters $\theta$ of
the network in an end-to-end manner using weakly supervised training data 
from the Google Street View Time Machine.

\section{Deep architecture for place recognition}
\label{sec:rep}

This section describes the proposed CNN architecture $f_\theta$,
guided by the best practices from the image retrieval community.
Most image retrieval pipelines are based on (i) extracting local descriptors,
which are then (ii) pooled in an orderless manner. The motivation behind this
choice is that the procedure provides
significant robustness to translation and partial occlusion.
Robustness to lighting and viewpoint changes is provided by the descriptors
themselves, and scale invariance is ensured through extracting descriptors
at multiple scales.

In order to learn the representation end-to-end, we design
a CNN architecture that mimics this standard retrieval pipeline in an unified
and principled manner with differentiable modules.
For step (i), we crop the CNN
 at the last convolutional layer and view it
as a dense descriptor extractor.
This has been observed to work well for instance retrieval
\cite{Azizpour14,Babenko15,Razavian15} and texture recognition~\cite{Cimpoi15}.
Namely, the output of the last convolutional layer is a
$H \times W \times D$ map which can be considered as a set of D-dimensional
descriptors extracted at $H \times W$ spatial locations.
For step (ii) we design a new pooling layer inspired by the Vector of Locally Aggregated Descriptors (VLAD)~\cite{Jegou10}
that pools extracted descriptors into a fixed image representation and its parameters are learnable via back-propagation.
 We call this new pooling layer ``NetVLAD" layer and describe it in the next section.

\subsection{NetVLAD: A Generalized VLAD layer ($f_{VLAD}$)}
\label{sec:VLAD}

Vector of Locally Aggregated Descriptors (VLAD)~\cite{Jegou10}
is a popular descriptor pooling method for both instance level retrieval~\cite{Jegou10} and image classification~\cite{Gong14}.
It captures information about the statistics
of local descriptors aggregated over the image. Whereas bag-of-visual-words~\cite{Csurka04,Sivic03} aggregation keeps counts
of visual words, VLAD stores the sum of residuals (difference vector between
the descriptor and its corresponding cluster centre) for each visual word.

Formally, given $N$ D-dimensional local image descriptors $\{\vc x_i\}$ as input,
and $K$ cluster centres (``visual words'') $\{\vc c_k\}$ as VLAD parameters,
the output VLAD image representation $V$ is $K \times D$-dimensional.
For convenience we will write $V$ as a $K \times D$ matrix, but this matrix
is converted into a vector and, after normalization, used as
the image representation. The $(j,k)$ element of $V$ is computed
as follows:
\begin{equation}
V(j,k) = \sum_{i=1}^N a_k(\vc x_i) \left( x_i(j) - c_k(j) \right),
\label{eq:vlad}
\end{equation}
where $x_i(j)$ and $c_k(j)$ are the $j$-th dimensions of the $i$-th descriptor and $k$-th cluster centre, respectively.
$a_k(\vc x_i)$ denotes the membership of
the descriptor $\vc x_i$ to $k$-th visual word, \ie it is $1$ if cluster $\vc c_k$
is the closest cluster to descriptor $\vc x_i$ and $0$ otherwise.
Intuitively, each D-dimensional column $k$ of $V$ records the sum of
residuals $(\vc x_i - \vc c_k)$ of descriptors which are assigned to cluster $\vc c_k$.
The matrix $V$ is then L2-normalized column-wise
(intra-normalization~\cite{arandjelovic13}),
converted into a vector,
and finally L2-normalized in its entirety \cite{Jegou10}.

In order to profit from years of wisdom produced in image retrieval,
we propose to mimic VLAD in a CNN framework
 and design a trainable generalized VLAD layer, \emph{NetVLAD}. The result is a powerful image representation  
  trainable end-to-end on the target task (in our case place recognition).
To construct a layer amenable to training via backpropagation,
it is required that the layer's operation is differentiable with
respect to all its parameters and the input. 
Hence, the key challenge is to make the VLAD pooling differentiable, which we describe next.

\figVLAD

The source of discontinuities in VLAD is the hard assignment $a_k(\vc x_i)$ of descriptors $\vc x_i$ to clusters centres $\vc c_k$.
To make this operation differentiable, we replace it with soft assignment of descriptors to multiple clusters 
\begin{equation}
\bar a_k(\vc x_i) =  \frac{e^{-\alpha\norm{\vc x_i-\vc c_k}^2}}{\sum_{k'}{e^{-\alpha\norm{\vc x_i-\vc c_{k'}}^2}}},
\label{eq:softa}
\end{equation}
which assigns the weight of descriptor $\vc x_i$ to cluster $\vc c_k$ proportional to their proximity, but relative to proximities to other cluster centres. $\bar a_k(\vc x_i)$ ranges between 0 and 1, with the highest weight assigned to the closest cluster centre. $\alpha$ is a parameter (positive constant) that controls the decay of the response with the magnitude of the distance. 
Note that for $\alpha \to +\infty$ this setup replicates the original VLAD exactly
as $\bar a_k(\vc x_i)$ for the closest cluster would be $1$ and $0$ otherwise.

By expanding the squares in~\eqref{eq:softa}, it is easy to see that the term  
$e^{-\alpha\norm{\vc x_i}^2}$ cancels between the numerator and the denominator
resulting in a soft-assignment of the following form
\begin{equation}
\bar a_k(\vc x_i) = \frac{e^{\vc w_k^T \vc x_i + b_k}}{\sum_{k'}{e^{\vc w_{k'}^T \vc x_i + b_{k'}}}},
\label{eq:softa2}
\end{equation}
where vector $\vc w_k=2 \alpha \vc c_k$ and scalar $b_k=-\alpha \norm{\vc c_k}^2$.
The final form of the NetVLAD layer is obtained by 
plugging the soft-assignment~\eqref{eq:softa2} into the VLAD descriptor~\eqref{eq:vlad} resulting in
\begin{equation}
V(j,k) = \sum_{i=1}^N
    \frac{e^{\vc w_k^T \vc x_i + b_k}}{\sum_{k'}{e^{\vc w_{k'}^T \vc x_i + b_{k'}}}}
    \left( x_i(j) - c_k(j) \right),
\label{eq:vladlayer}
\end{equation}
where
$\{\vc w_k\}$, $\{b_k\}$ and $\{\vc c_k\}$ are sets of trainable parameters for each cluster $k$.
Similarly to the original VLAD descriptor, the NetVLAD layer aggregates the first order statistics of residuals $(\vc x_i - \vc c_k)$
in different parts of the descriptor space weighted by the soft-assignment $\bar a_k(\vc x_i)$ of descriptor $\vc x_i$ to cluster $k$.
Note however, that the NetVLAD layer has three independent
sets of parameters $\{\vc w_k\}$, $\{b_k\}$ and $\{\vc c_k\}$, compared to just
$\{\vc c_k\}$ of the original VLAD. This enables greater flexibility than the original VLAD,
as explained in figure \ref{fig:supVLAD}.
Decoupling $\{\vc w_k,b_k\}$ from $\{\vc c_k\}$ has been proposed in
\cite{arandjelovic13} as a means to adapt the VLAD to a new dataset.
All parameters of NetVLAD are learnt for the specific task in an end-to-end manner.

\figSupVLAD

As illustrated in figure \ref{fig:VLAD} the NetVLAD layer can be visualized as a meta-layer that is further decomposed into
basic CNN layers connected up in a directed acyclic graph.
First, note that the first term in eq.~\eqref{eq:vladlayer} is a
soft-max function $\sigma_k(\vc z)=\frac{\exp(z_k)}{\sum_{k'}{\exp(z_{k'})}}$.
Therefore, the soft-assignment of the input array of descriptors $\vc x_i$ into $K$ clusters can be seen as a two step process:
(i)
a convolution with a set of $K$ filters $\{\vc w_k\}$ that have spatial support  $1 \times 1$ and biases $\{b_k\}$, producing the output
$s_k(\vc x_i)=\vc w_k^T \vc x_i + b_k $;
(ii)
the convolution output is then passed through the soft-max function $\sigma_k$ to obtain the final soft-assignment $\bar a_k(\vc x_i)$ that weights
the different terms in the aggregation layer that implements eq.~\eqref{eq:vladlayer}. The output after normalization is a $(K\times D)\times 1$ descriptor.

\paragraph{Relations to other methods.}
Other works have proposed to pool CNN activations using VLAD or
Fisher Vectors (FV) \cite{Gong14,Cimpoi15}, but do not learn the VLAD/FV
parameters nor the input descriptors.
The most related method to ours is the one of Sydorov \etal \cite{Sydorov14},
which proposes to learn
FV parameters jointly with an SVM for the end classification
objective. However, in their work it is not possible to learn the input descriptors
as they are hand-engineered (SIFT), while our VLAD layer is easily
pluggable into any CNN architecture as it is amenable to backpropagation.
``Fisher Networks''~\cite{Simonyan13b}  stack Fisher Vector
layers on top of each other, but the system is not trained
end-to-end,  only hand-crafted features are used, and the layers
are trained greedily in a bottom-up fashion. 
Finally, our architecture is also related to bilinear networks~\cite{Lin15a}, recently developed for a different task of fine-grained category-level recognition.

\paragraph{Max pooling ($f_{max}$).}
We also experiment with Max-pooling
of the D-dimensional features across the $H \times W$ spatial
locations, thus producing a D-dimensional output vector,
which is then L2-normalized.
Both of these operations can be implemented using standard layers in
public CNN packages.
This setup mirrors the method of
\cite{Azizpour14,Razavian15}, but a crucial difference is that we will
learn the representation (section \ref{sec:learning}) while
\cite{Razavian14,Azizpour14,Razavian15} only use pretrained networks.
Results will show (section \ref{sec:res}) that simply using CNNs
off-the-shelf \cite{Razavian14} results in poor performance, and that
training for the end-task is crucial.
Additionally, VLAD will prove itself to be superior to the
Max-pooling baseline.

\section{Learning from Time Machine data}

\label{sec:learning}

\figTM

In the previous section we have designed a new CNN architecture as an image representation for place recognition.
Here we describe how to learn its parameters in an end-to-end manner for the place recognition task.
The two main challenges are: (i) how to gather enough annotated training data and (ii) what is the appropriate loss
for the place recognition task. To address theses issues, we will first show that it is possible to obtain large amounts of weakly labelled imagery depicting the same places over time from the Google Street View Time Machine. Second, we will design a new weakly supervised triplet ranking loss that can deal with the incomplete and noisy position annotations of the  Street View Time Machine imagery.  The details are below.

\paragraph{Weak supervision from the Time Machine.}
We propose to exploit a new source of data -- Google Street View Time Machine --
which provides multiple street-level panoramic images taken at different times at close-by spatial locations on the map. 
As will be seen in section \ref{sec:res},
this novel data source is precious for learning an image representation for place recognition.
As shown in figure~\ref{fig:timemachine}, the same locations are depicted
at different times and seasons, providing the learning algorithm with crucial
information it can use to discover which features are useful or distracting,
and what changes should the image representation be invariant to, in order to achieve
good place recognition performance.

The downside of the Time Machine imagery is that it provides only incomplete and noisy supervision.
Each Time Machine panorama comes with a GPS tag giving only its approximate location on the map, which can be used to
identify close-by panoramas but does not provide correspondences between parts of the depicted scenes.
In detail, as the test queries are perspective images from camera phones, each panorama is represented by a set of perspective images sampled evenly 
in different orientations and two elevation angles~\cite{Knopp10,Chen11b,Gronat13,Torii13}. Each perspective image is labelled with the GPS position 
of the source panorama. As a result, two geographically close perspective images do not necessarily depict the same objects
since they could be facing different directions or occlusions could take place (\eg the two images are around a corner from each other), \etc.
 Therefore, for a given training query $q$, the GPS information
can only be used as a source of (i) \emph{potential} positives $\{p^q_i\}$, \ie images that are 
geographically close to the query, and (ii) \emph{definite} negatives $\{n^q_j\}$, \ie images that are geographically far from the query.\footnote{Note that even faraway images can
depict the same object. For example, the Eiffel Tower can be visible from two faraway locations in Paris. But, for the purpose of localization we consider in this paper such image pairs as negative examples because they are not taken from the same place.}

\paragraph{Weakly supervised triplet ranking loss.}
We wish to learn a representation $f_\theta$ that will optimize place recognition performance.
That is, for a given test query image $q$, the goal is to rank a database image $I_{i*}$ from a close-by location 
higher than all other far away images $I_i$ in the database. In other words, we wish  
the Euclidean distance $d_\theta(q,I)$ %
between the query $q$ and a close-by image $I_{i*}$ to be smaller than the distance to far away images in the database $I_{i}$, \ie
$d_\theta(q,I_{i*}) < d_\theta(q,I_{i})$, for all images $I_i$ further than a certain distance from the query on the map.
Next we show how this requirement can be translated into a ranking loss between training triplets $\{q, I_{i*}, I_{i}\}$. 
 
From the Google Street View Time Machine data, we obtain a training dataset of tuples
$(q, \{p^q_i\}, \{n^q_j\})$, where for each training query image $q$ we have
a set of potential positives $\{p^q_i\}$ and the set of definite negatives $\{n^q_j\}$.
 The set of potential positives contains {\em at least one} positive image that should match the query, but we do not know which one.
To address this ambiguity, we propose to identify the best matching potential positive image $p^q_{i*}$
\begin{equation}
p^q_{i*} = \argmin_{p^q_i} d_\theta(q,p^q_i)
\label{eq:pstar} 
\end{equation}
for each training tuple $(q, \{p^q_i\}, \{n^q_j\})$.
The goal then becomes to learn an image representation $f_\theta$ so that distance $d_\theta(q,p^q_{i*})$ %
between the training query $q$ and the best matching potential positive $p^q_{i*}$
is smaller than the distance $d_\theta(q,n^q_{j})$ between the query $q$ and {\em all} negative images $q_{j}$: 
\begin{equation}
d_\theta(q,p^q_{i*}) < d_\theta(q,n^q_j), ~~~ \forall j.
\label{eq:distrank} 
\end{equation}
Based on this intuition %
we define a {\em weakly supervised ranking loss} $L_\theta$ for a training tuple  $(q, \{p^q_i\}, \{n^q_j\})$ as
\begin{equation}
L_\theta = \sum_j l\left( \min_i d^2_\theta(q,p^q_i) + m - d^2_\theta(q,n^q_j) \right),
\label{eq:loss}
\end{equation}
where $l$ is the hinge loss $l(x)=\max(x,0)$, and $m$ is a constant parameter giving the margin.
Note that equation~\eqref{eq:loss} is a sum of individual losses for negative images $n^q_j$. For each negative, the loss $l$ is zero if the distance between the query and the negative is greater by a margin than the distance between the query and the best matching positive. Conversely,
if the margin between the distance to the negative image and to the best matching positive is violated,
the loss is proportional to the amount of violation.
Note that the above loss is related to the commonly used triplet loss
\cite{Schultz04,Weinberger06,Wang14,Schroff15}, but adapted to our weakly
supervised scenario using a formulation (given by equation~\eqref{eq:pstar}) similar to multiple instance learning~\cite{Foulds10,Kotzias14,Viola05}.

We train the parameters $\theta$ of the representation $f_\theta$ using
Stochastic Gradient Descent (SGD) on a large set of training tuples from Time Machine data.
Details of the training procedure are given in
\isArXiv{appendix \ref{sup:impl}}{the appendix \cite{Arandjelovic15}}.

\section{Experiments}

In this section we describe the used datasets and evaluation methodology (section~\ref{sec:datasets}), 
and give quantitative (section~\ref{sec:res}) and qualitative (section~\ref{sec:res:visual}) results to validate our approach.
Finally, we also test the method on the standard image retrieval benchmarks
(section~\ref{sec:res:retrieval}).
\subsection{Datasets and evaluation methodology}
\label{sec:datasets}

We report results on two publicly available datasets.

\paragraph{Pittsburgh (Pitts250k) \cite{Torii13}}
contains 250k
database images downloaded from Google Street View  
and 24k test queries generated from Street View but taken at different
times, years apart.
We divide this dataset into three roughly equal parts
for training, validation and testing,
each containing around 83k
database images and 8k queries,
where the division was done geographically to ensure the sets contain
independent images.
To facilitate faster training, for some experiments,
a smaller subset (Pitts30k) is used, containing 10k database images
in each of the train/val(idation)/test sets, which are
also geographically disjoint.

\paragraph{Tokyo 24/7 \cite{Torii15}}
contains 76k database images and
315 query images taken using mobile phone cameras.
This is an extremely challenging dataset where the queries were taken at daytime, sunset and night, while the database
images were only taken at daytime as they originate from Google Street View
as described above. 
To form the train/val sets we collected
additional Google Street View panoramas of Tokyo using the
Time Machine feature, and name this set {\bf TokyoTM};
Tokyo 24/7 (=test) and
TokyoTM train/val are all geographically disjoint.
Further details on the splits are given in \isArXiv{appendix \ref{sup:datasets}}{the appendix \cite{Arandjelovic15}}.

\paragraph{Evaluation metric.}
We follow the standard place recognition evaluation procedure
\cite{arandjelovic14a,Gronat13,Sattler12,Torii13,Torii15}.
The query image is deemed correctly localized if at least one of the top $N$ retrieved
database images is within $d=25$ meters from the ground truth position of the query.
The percentage of correctly recognized queries (Recall) is then plotted for different values of $N$.
For Tokyo 24/7 we follow~\cite{Torii15} and perform spatial non-maximal suppression on ranked database images before evaluation.

\paragraph{Implementation details.}
We use two base architectures which are extended with Max pooling ($f_{max}$) and
our NetVLAD ($f_{VLAD}$) layers:
AlexNet \cite{Krizhevsky12} and
VGG-16 \cite{Simonyan15};
both are cropped at the last convolutional layer (conv5), before ReLU.
For NetVLAD we use $K=64$ resulting in 16k and 32k-D image representations
for the two base architectures, respectively.
The initialization procedure, parameters used for training, procedure for sampling training tuples
and other implementation details are given in
\isArXiv{appendix \ref{sup:impl}}{the appendix \cite{Arandjelovic15}}.
All training and evaluation code, as well as our trained networks,
are online at \cite{netvladurl}.

\subsection{Results and discussion}
\label{sec:res}

\figStoa

\vspace{-0.2cm}
\paragraph{Baselines and state-of-the-art.}
To assess benefits of our approach %
we compare our representations trained for place recognition against ``off-the-shelf" networks
pretrained on other tasks. Namely,
given a base network cropped at conv5,
the baselines either use Max pooling ($f_{max}$),
or aggregate the descriptors into VLAD ($f_{VLAD}$),
but perform no further task-specific training.
The three base networks are:
AlexNet \cite{Krizhevsky12},
VGG-16 \cite{Simonyan15},
both are pretrained for ImageNet classification \cite{Deng09},
and
Places205 \cite{Zhou14},
reusing the same architecture as AlexNet but
pretrained for scene classification \cite{Zhou14}.
Pretrained networks have been recently used as off-the-shelf
dense descriptor extractors for instance retrieval~\cite{Azizpour14,Babenko15,Gong14,Razavian14,Razavian15} and
the untrained $f_{max}$ network corresponds to the method of
\cite{Azizpour14,Razavian15}.

Furthermore we compare our CNN representations trained for place recognition
against the state-of-the-art local feature based compact descriptor, which consists of 
VLAD pooling \cite{Jegou10} with intra-normalization \cite{arandjelovic13}
on top of densely extracted RootSIFTs \cite{Lowe04,Arandjelovic12}.
The descriptor is optionally reduced to 4096 dimensions using
PCA (learnt on the training set) combined with whitening and L2-normalization \cite{Jegou12};
this setup together with view synthesis yields the state-of-the-art results on the challenging Tokyo 24/7 dataset
(\cf \cite{Torii15}).

In the following we discuss figure \ref{fig:stoa},
which compares place recognition performance of our method to the baselines outlined above
on the Pittsburgh and Tokyo 24/7 benchmarks.

\paragraph{Dimensionality reduction.}
We follow the standard state-of-the-art procedure to perform dimensionality
reduction of VLAD, as described earlier,
\ie the reduction into 4096-D is performed using
PCA with whitening followed by L2-normalization \cite{Jegou12,Torii15}.
Figure \ref{fig:stoa} shows that the lower dimensional $f_{VLAD}$ (-$\ast$-)
performs
similarly to the full size vector (-o-).

\paragraph{Benefits of end-to-end training for place recognition.}
Representations trained on the end-task of place recognition
consistently outperform by a large margin off-the-shelf CNNs on both benchmarks.
For example, on the Pitts250k-test our trained AlexNet with (trained) NetVLAD aggregation layer achieves recall@1
of 81.0\% compared to only 55.0\% obtained by off-the-shelf AlexNet with standard VLAD aggregation, \ie a relative improvement in recall of 47\%.
Similar improvements can be observed on all three datasets. 
This confirms two important premises of this work:
(i) our approach can learn rich yet compact image representations for place recognition, and
(ii) the popular idea of using pretrained networks ``off-the-shelf''
\cite{Razavian14,Azizpour14,Gong14,Babenko15,Razavian15}
is sub-optimal as the networks trained for object or scene classification
are not necessary suitable for the end-task of place recognition.
We believe this could be attributed to the fact that 
``off-the-shelf " conv5 activations are not trained to be comparable using Euclidean distance.

\paragraph{Comparison with state-of-the-art.}
Figure \ref{fig:stoa} also shows that our trained $f_{VLAD}$
representation with whitening based on VGG-16
({\color{magenta} magenta -$\ast$-})
convincingly outperforms RootSIFT+VLAD+whitening,
as well as the method of Torii \etal~\cite{Torii15}, 
and therefore
sets the state-of-the-art for compact descriptors on all benchmarks.
Note that these are strong baselines that outperform most off-the-shelf CNN
descriptors on the place recognition task.

\paragraph{VLAD versus Max.}
By comparing $f_{VLAD}$ (-o-) methods with their corresponding $f_{max}$ (-x-) counterparts
it is clear that VLAD pooling is much better than Max pooling for both off-the-shelf and trained representations.
NetVLAD performance decreases gracefully
with dimensionality: 128-D NetVLAD performs similarly to 512-D Max
(42.9\% vs 38.4\% recall@1 on Tokyo 24/7),
resulting in {\em four} times
more compact representation for the same performance.
Furthermore, NetVLAD+whitening outperforms Max pooling convincingly when
reduced to the same dimensionality (60\%).
See \isArXiv{appendix \ref{sup:res}}{the appendix \cite{Arandjelovic15}} for more details.

\paragraph{Which layers should be trained?}
In Table \ref{tab:layers} we study the benefits of training different layers for the end-task of place recognition. 
The largest improvements are thanks to training the NetVLAD layer, but training other layers 
results in further improvements, with some overfitting occurring below conv2.

\tabResLayers

\paragraph{Importance of Time Machine training.}
Here we examine whether the network can be trained without the Time Machine (TM) data.
In detail, we have modified the training query set for Pitts30k-train to be sampled from the same 
set as the training database images, \ie the tuples of query and database images used in training were captured at the same time.
Recall@1 with $f_{max}$ on Pitts30k-val for the off-the-shelf AlexNet is 33.5\%, and
training without TM improves this to 38.7\%.
However, training with TM obtains 68.5\% showing
that Time Machine data is crucial
for good place recognition accuracy as without it the network does not generalize well.
The network learns, for example,
that recognizing cars is important for place recognition,
as the same parked cars appear in all images of a place.

\subsection{Qualitative evaluation}
\vspace{-0.2cm}
\label{sec:res:visual}

\figFocus

To visualize what is being learnt by our place recognition architectures,
we adapt the method of Zeiler and Fergus \cite{Zeiler14} for examining 
 occlusion sensitivity of classification networks.
It can be seen in figure~\ref{fig:focus}
that off-the-shelf AlexNet (pretrained on ImageNet) focuses very much
on categories it has been trained to recognize (\eg cars)
and certain shapes,
such as circular blobs
useful for distinguishing 12 different ball types in the ImageNet categories.
The Place205 network is fairly unresponsive to all occlusions as it does not
aim to recognize specific places but scene-level categories,
so even if an important part of the image is occluded, such as
a characteristic part of a building fa\c{c}ade, it still provides a similar
output feature which corresponds to an uninformative
``a building fa\c{c}ade'' image descriptor.
In contrast to these two,
our network trained for specific place recognition automatically learns
to ignore confusing features, such as cars and people, which are not
discriminative for specific locations, and instead focuses on describing
building fa\c{c}ades and skylines.
More qualitative examples are provided in
\isArXiv{appendix \ref{sup:res}}{the appendix \cite{Arandjelovic15}}.

\subsection{Image retrieval}
\vspace{-0.2cm}
\label{sec:res:retrieval}

We use our best performing network (VGG-16, $f_{VLAD}$ with whitening down to 256-D)
trained completely on Pittsburgh, to extract image representations
for standard object and image retrieval benchmarks.
Our representation sets the state-of-the-art for compact image representations (256-D)
by a large margin on all three datasets, obtaining an mAP of
63.5\%, 73.5\% and 79.9\% on
Oxford 5k \cite{Philbin07}, Paris 6k  \cite{Philbin08}, Holidays \cite{Jegou08}, respectively;
for example, this is a +20\% relative improvement on Oxford 5k.
\isArXiv{Appendix \ref{sup:ret}}{The appendix \cite{Arandjelovic15}}
contains more detailed results.

\vspace{-0.1cm}
\section{Conclusions}
\vspace{-0.15cm}

We have designed a new convolutional neural network architecture that
is trained for place recognition in an end-to-end manner from weakly
supervised Street View Time Machine data. Our trained representation
significantly outperforms off-the-shelf CNN models and significantly
improves over the state-of-the-art on the challenging 24/7 Tokyo
dataset, as well as on the Oxford and Paris image retrieval benchmarks.
The two main components of our architecture
-- (i) the NetVLAD pooling layer and (ii) weakly supervised ranking
loss -- are generic CNN building blocks applicable beyond the place
recognition task. The NetVLAD layer offers a powerful pooling
mechanism with learnable parameters that can be easily plugged into
any other CNN architecture. The weakly supervised ranking loss
opens up the possibility of end-to-end learning for other ranking
tasks where large amounts of weakly labelled data are available, for
example, images described with natural language~\cite{Karpathy15}.

{\footnotesize
\paragraph{Acknowledgements.}
This work was partly supported by
RVO13000 - Conceptual development of research organization,
the ERC grant LEAP (no.\ 336845), ANR project Semapolis (ANR-13-CORD-0003),
JSPS KAKENHI Grant Number 15H05313, the Inria CityLab IPL, 
and the Intelligence Advanced Research Projects Activity (IARPA) via Air Force Research Laboratory, contract FA8650-12-C-7212.
The U.S.\ Government is authorized to reproduce and distribute reprints for Governmental purposes notwithstanding any copyright annotation thereon. Disclaimer: The views and conclusions contained herein are those of the authors and should not be interpreted as necessarily representing the official policies or endorsements, either expressed or implied, of IARPA, AFRL, or the U.S.\ Government.
}

{\small
\bibliographystyle{ieee}
\bibliography{vgg_bib/shortstrings,vgg_bib/vgg_local,vgg_bib/vgg_other,to_add}
}

\isArXiv{
\appendix

\def\figTMSup{
\def\wTMSup{0.16\linewidth}
\begin{figure*}[t!]
\begin{center}
\hspace*{-0.82cm}
    \begin{tabular}{cccccc}
\includegraphics[width=\wTMSup]{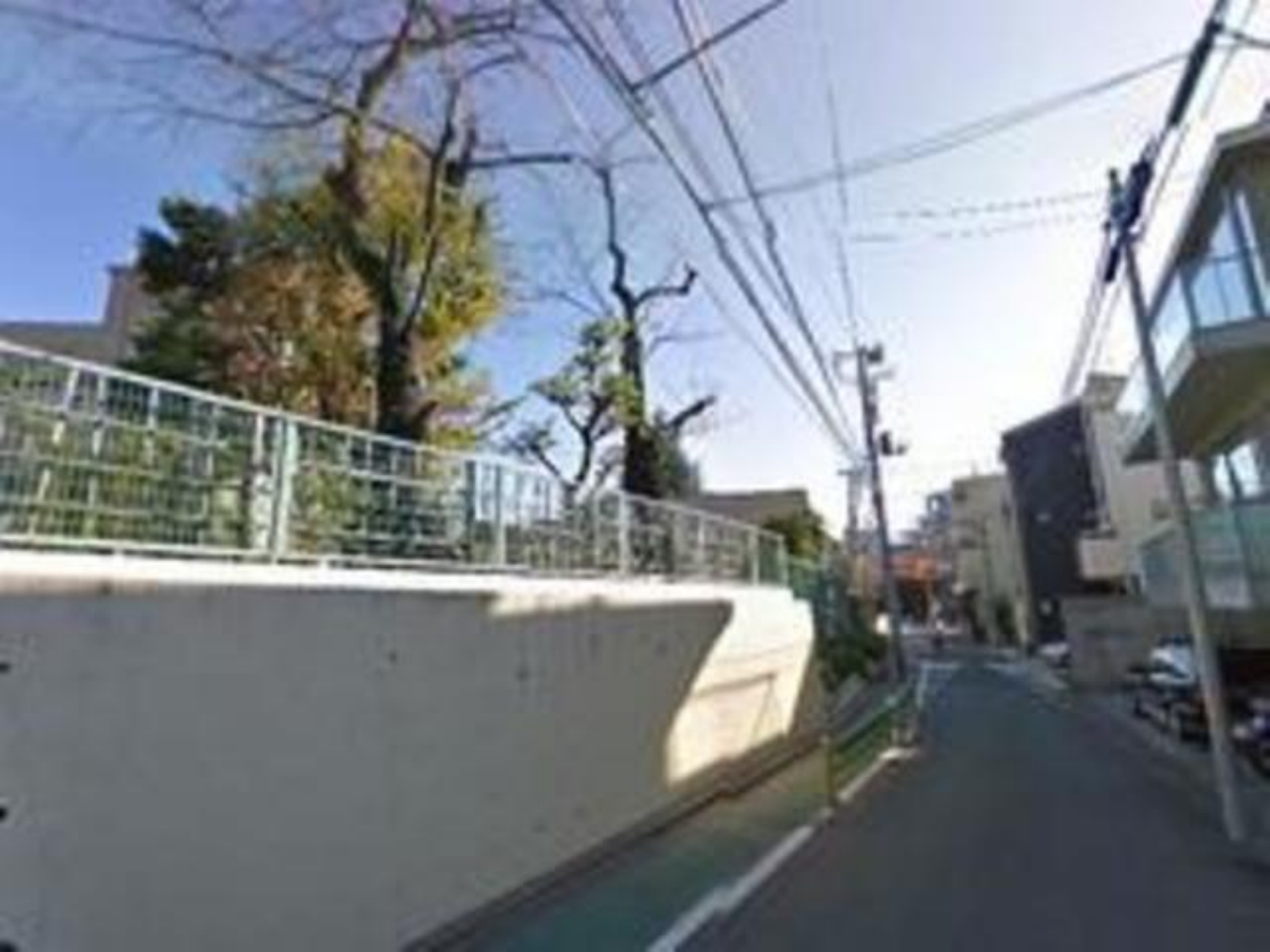}
& \includegraphics[width=\wTMSup]{figures/time_machine_cutouts/bzR8Zji0UYeqxpNq3EmBRA__200912_35654563_139696944_300_012.pdf}
& \includegraphics[width=\wTMSup]{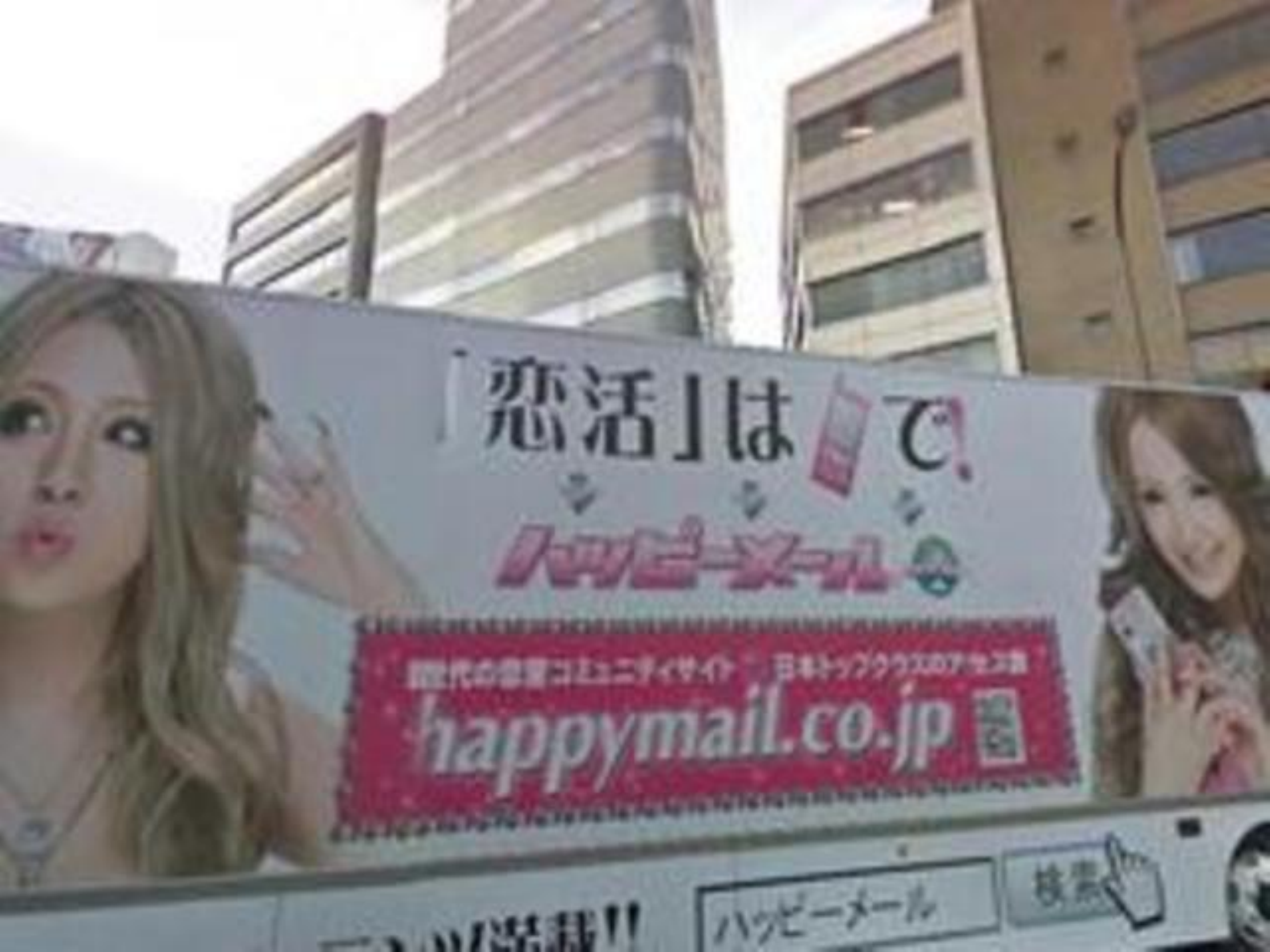}
& \includegraphics[width=\wTMSup]{figures/time_machine_cutouts/Q7QhFUO3xMkgLje98ju_Zw___35650662_139703173_120_012.pdf}
& \includegraphics[width=\wTMSup]{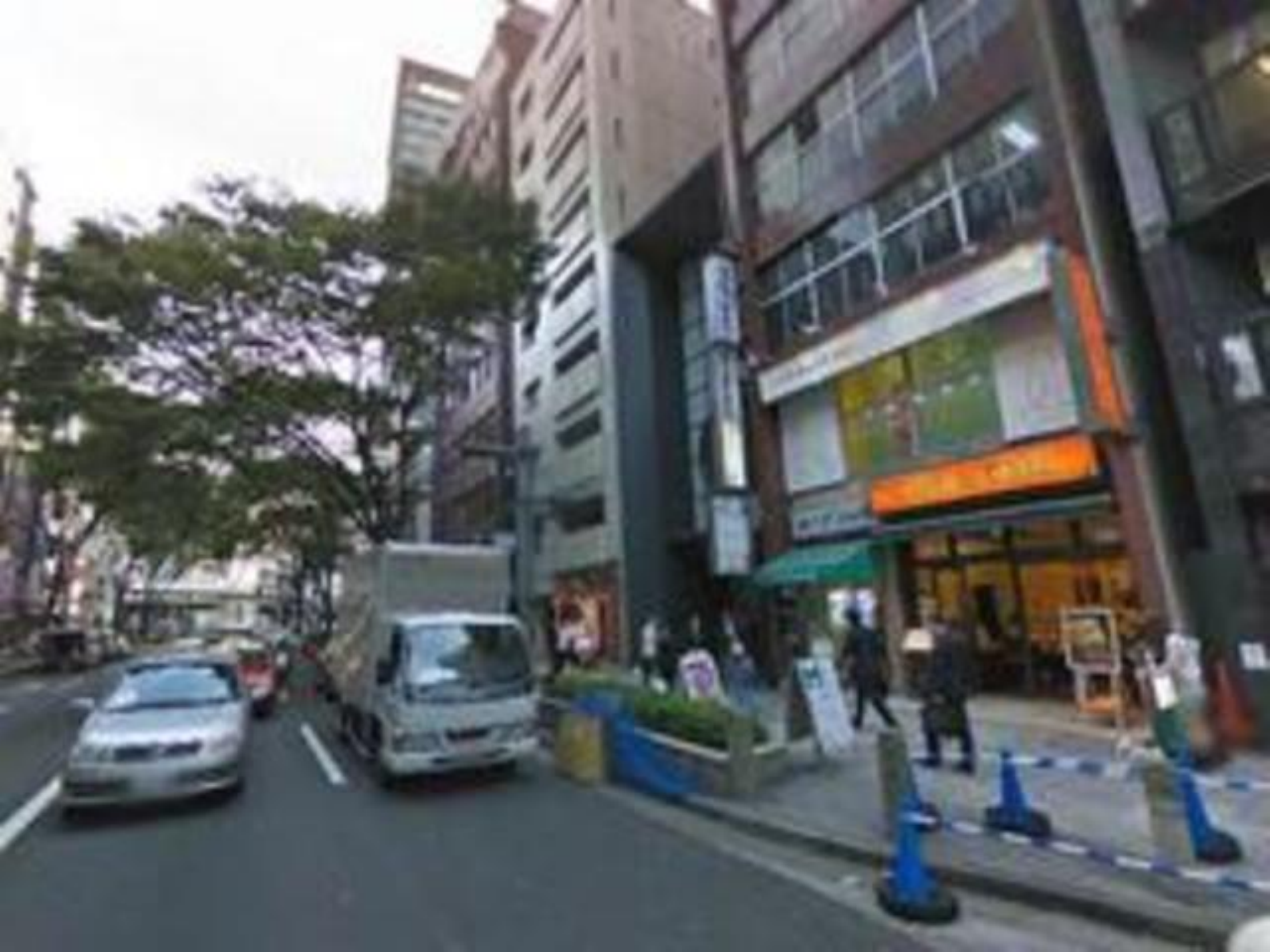}
& \includegraphics[width=\wTMSup]{figures/time_machine_cutouts/lJrZpauFJodydIkU2r6SoQ__200911_35657303_139695911_210_012.pdf}
    \\
\includegraphics[width=\wTMSup]{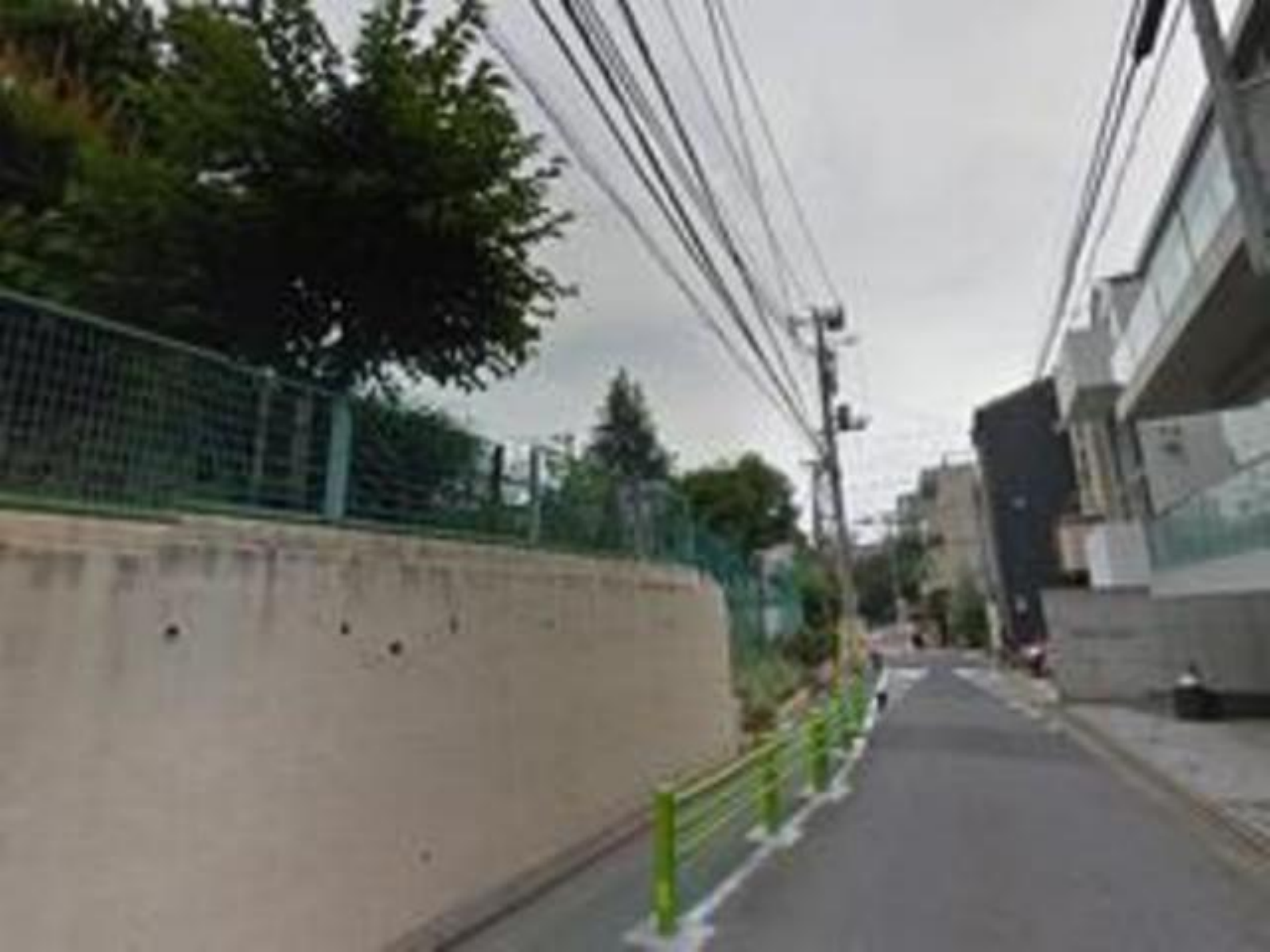}
& \includegraphics[width=\wTMSup]{figures/time_machine_cutouts/bzR8Zji0UYeqxpNq3EmBRA__201306_35654563_139696944_300_012.pdf}
& \includegraphics[width=\wTMSup]{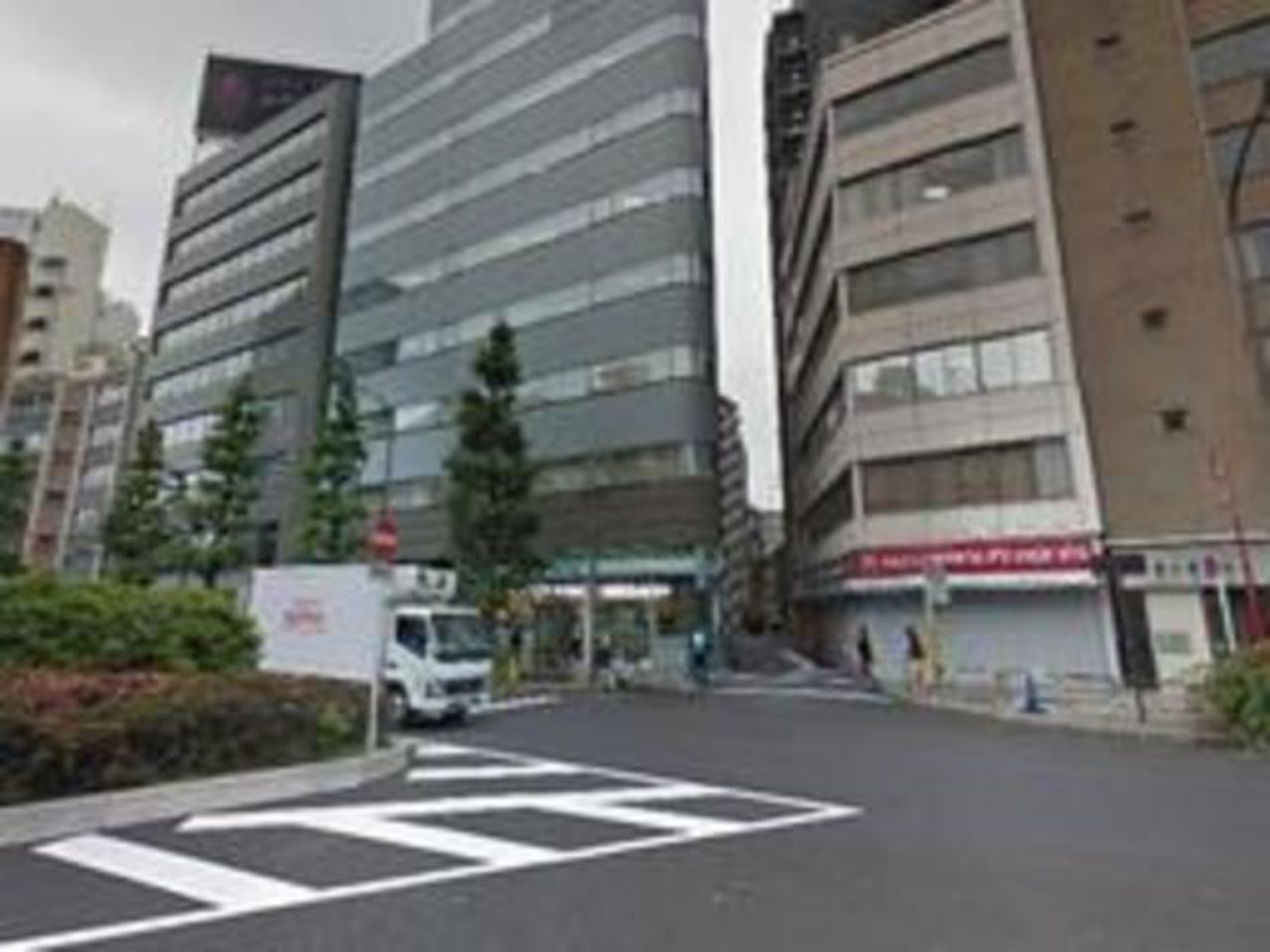}
& \includegraphics[width=\wTMSup]{figures/time_machine_cutouts/Q7QhFUO3xMkgLje98ju_Zw__200912_35650662_139703173_120_012.pdf}
& \includegraphics[width=\wTMSup]{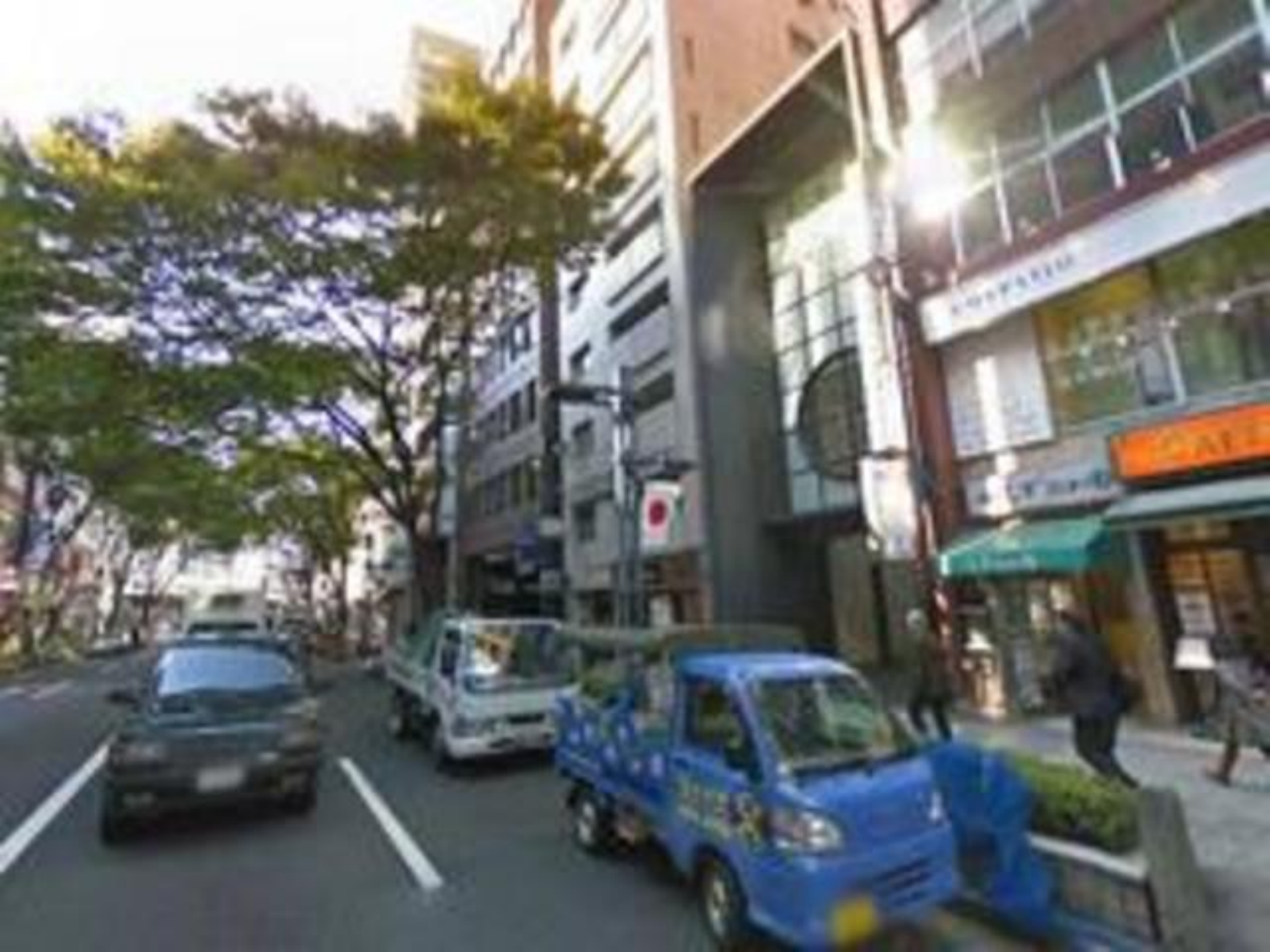}
& \includegraphics[width=\wTMSup]{figures/time_machine_cutouts/lJrZpauFJodydIkU2r6SoQ__201407_35657303_139695911_210_012.pdf}
    \\
\includegraphics[width=\wTMSup]{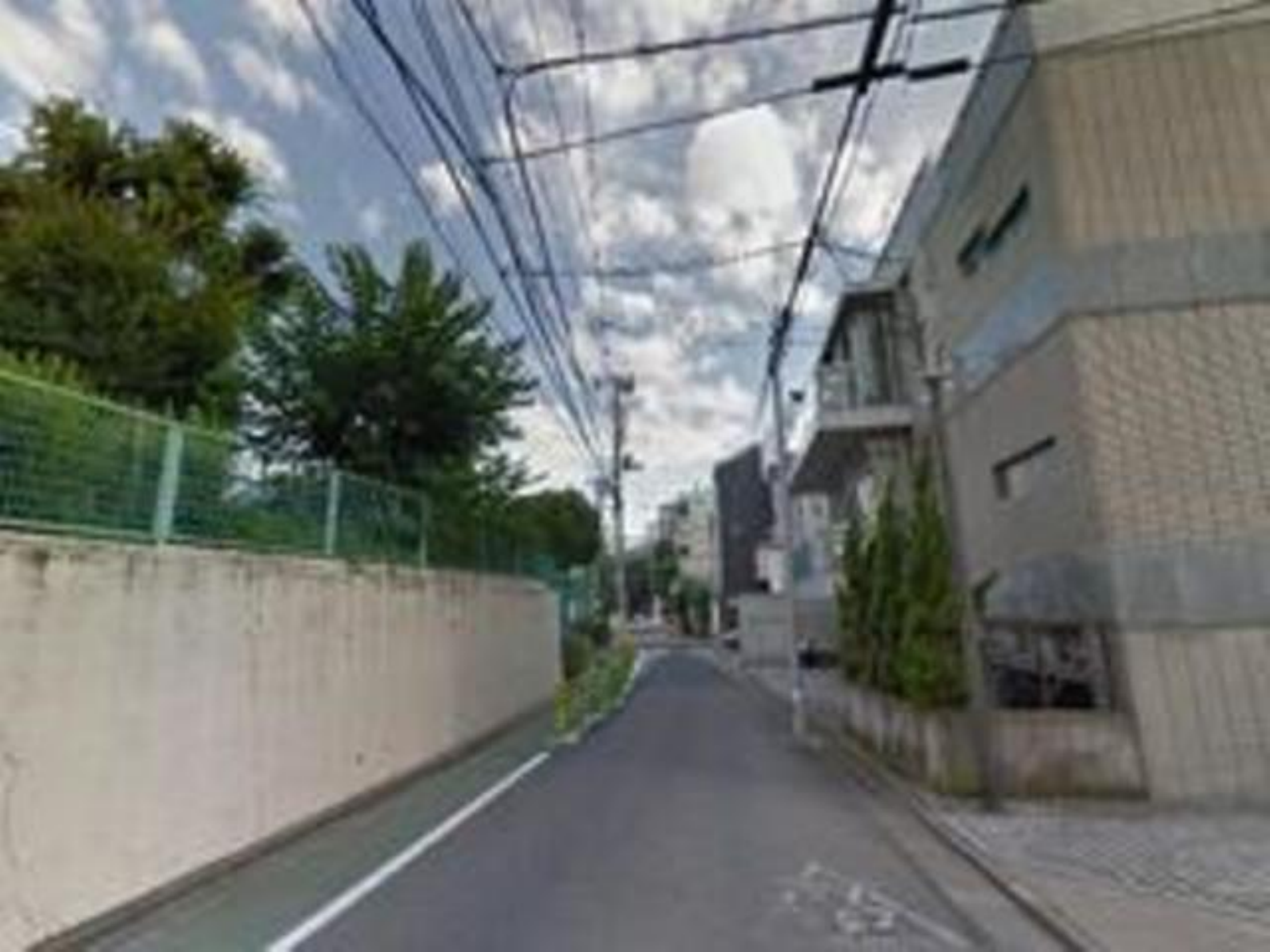}
& \includegraphics[width=\wTMSup]{figures/time_machine_cutouts/bzR8Zji0UYeqxpNq3EmBRA__201407_35654563_139696944_300_012.pdf}
& \includegraphics[width=\wTMSup]{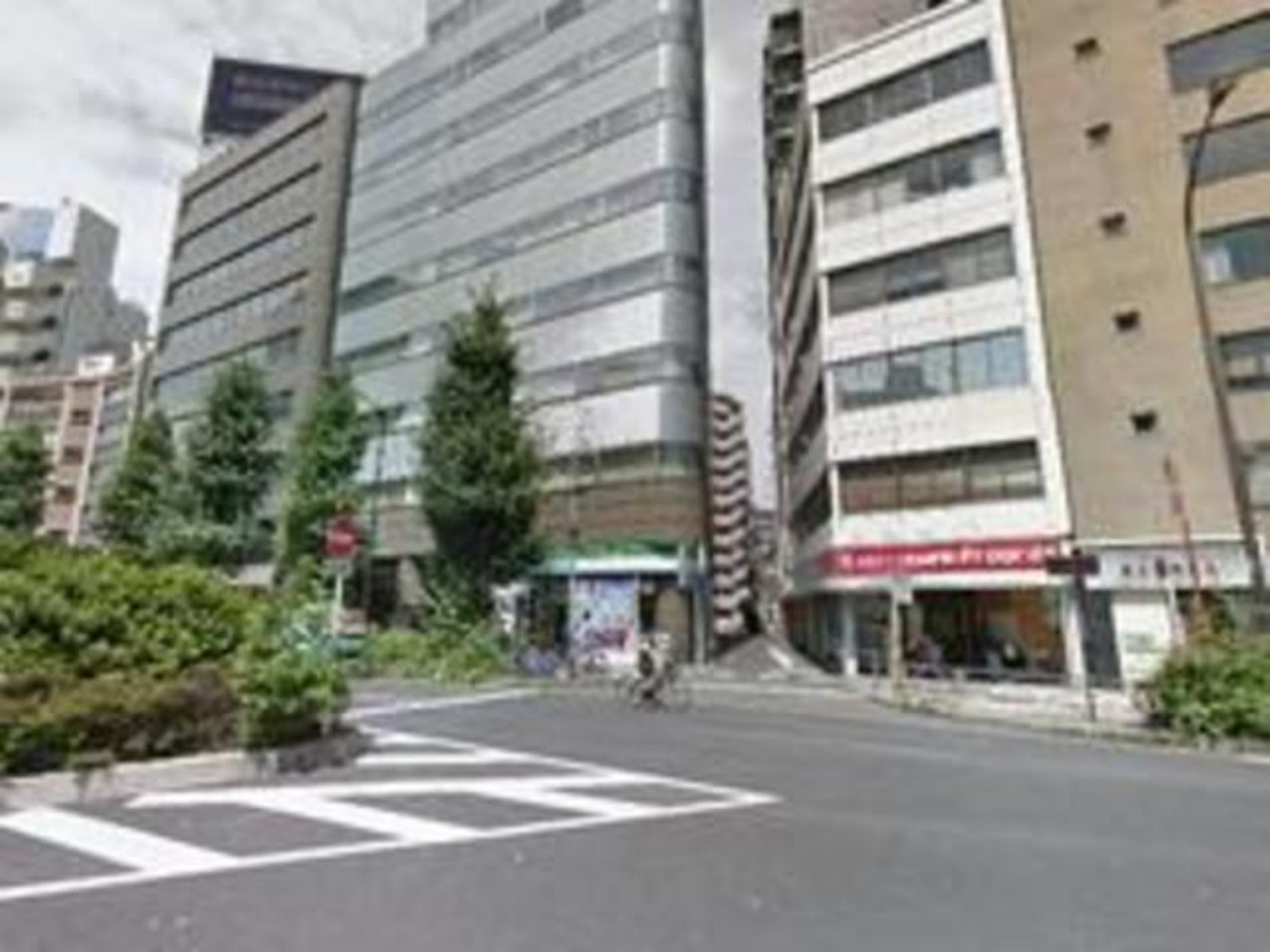}
& \includegraphics[width=\wTMSup]{figures/time_machine_cutouts/Q7QhFUO3xMkgLje98ju_Zw__201306_35650662_139703173_120_012.pdf}
& \includegraphics[width=\wTMSup]{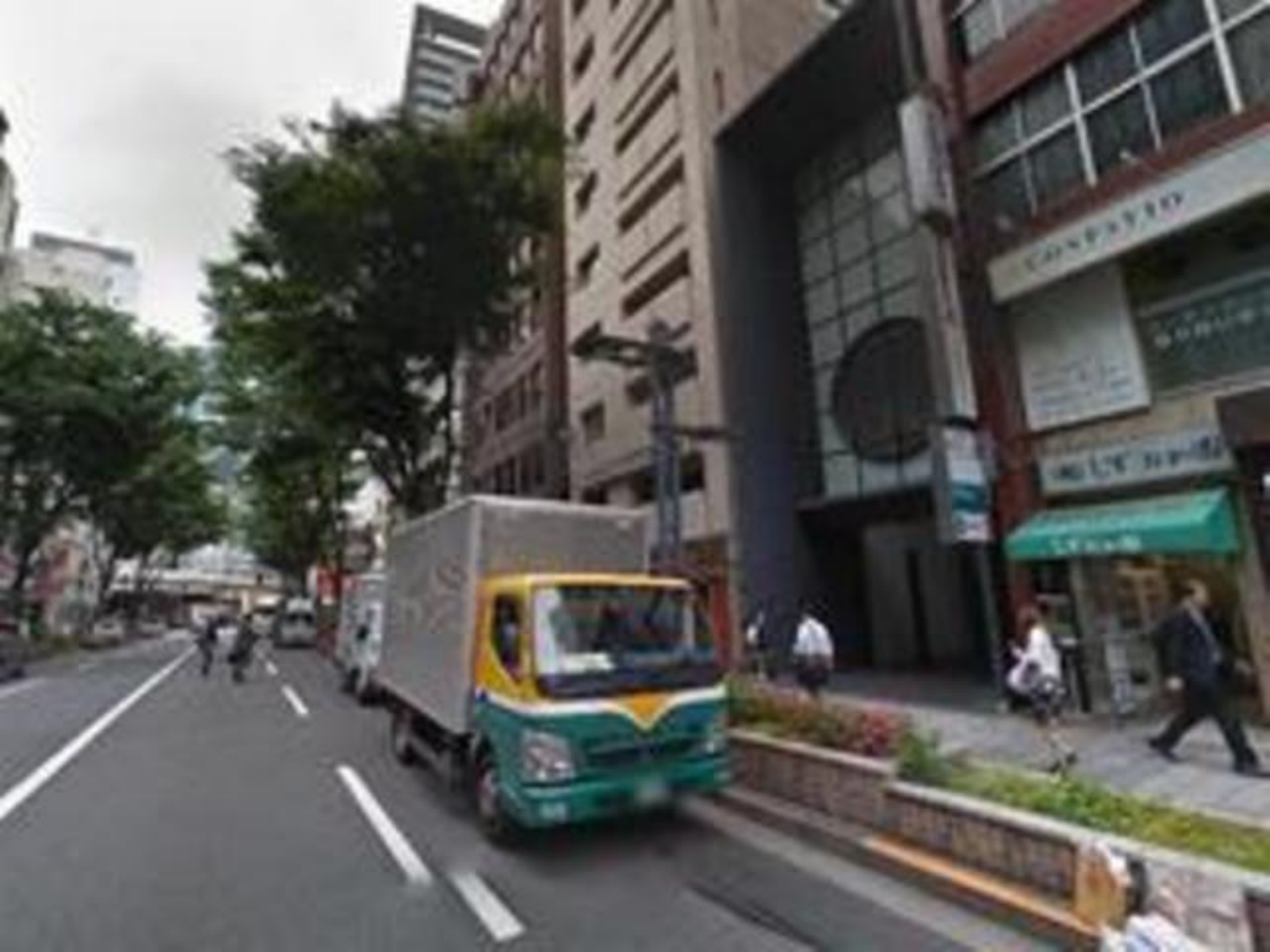}
& \includegraphics[width=\wTMSup]{figures/time_machine_cutouts/lJrZpauFJodydIkU2r6SoQ__201503_35657303_139695911_210_012.pdf}
\\
  {\small(a)}
& {\small(b)}
& {\small(c)}
& {\small(d)}
& {\small(e)}
& {\small(f)}
    \end{tabular}
\end{center}
\vspace{-0.3cm}
    \caption{
{\bf Google Street View Time Machine examples.}
Each column shows perspective images generated from panoramas from nearby locations,
taken at different times.
The goal of this work is to learn from this imagery an image representation that: has a degree of invariance to changes in viewpoint and illumination (a-f); has tolerance to partial occlusions (c-f); suppresses confusing visual information such as clouds (a,c), vehicles (c-f) and people (c-f);
and chooses to either ignore vegetation or learn a season-invariant vegetation representation (a-f).
}
\label{fig:timemachineSup}
\end{figure*}
}

\def\figStoaSup{
\def\wStoa{0.3\linewidth}
\begin{figure*}[p!]
\begin{center}
\hspace*{-0.5cm}
    \subfloat[Legend]{
        \includegraphics[width=0.2\linewidth, viewport=115 42 262 240, clip=true]{figures/tokyoCvpr15_res0.pdf}
    }
    \\
    \subfloat[Pitts30k-test]{
        \includegraphics[width=\wStoa]{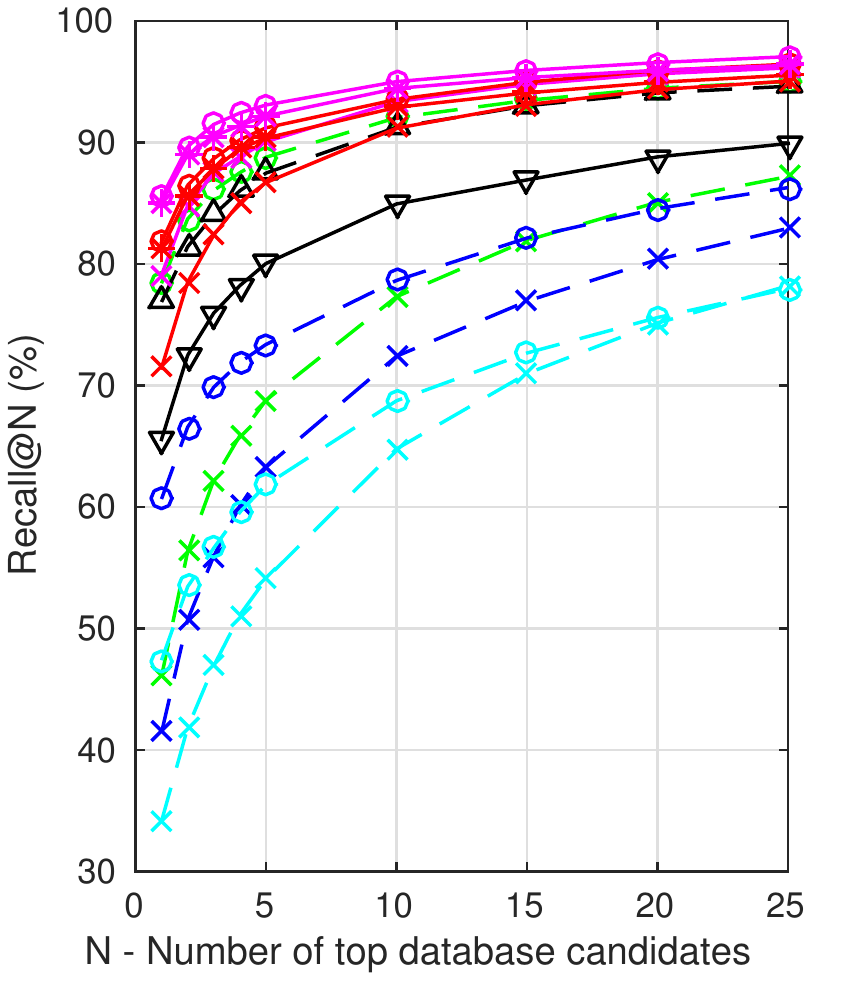}
    }
    \subfloat[Pitts250k-test]{
        \includegraphics[width=\wStoa]{figures/pitts_fs1_test_noleg.pdf}
    }
    \subfloat[TokyoTM-val]{
        \includegraphics[width=\wStoa]{figures/tokyoPetr151021_val_noleg.pdf}
    }
    \\
    \subfloat[Tokyo 24/7 all queries]{
        \includegraphics[width=\wStoa]{figures/tokyoCvpr15_res0_noleg.pdf}
    }
    \subfloat[Tokyo 24/7 daytime]{
        \includegraphics[width=\wStoa]{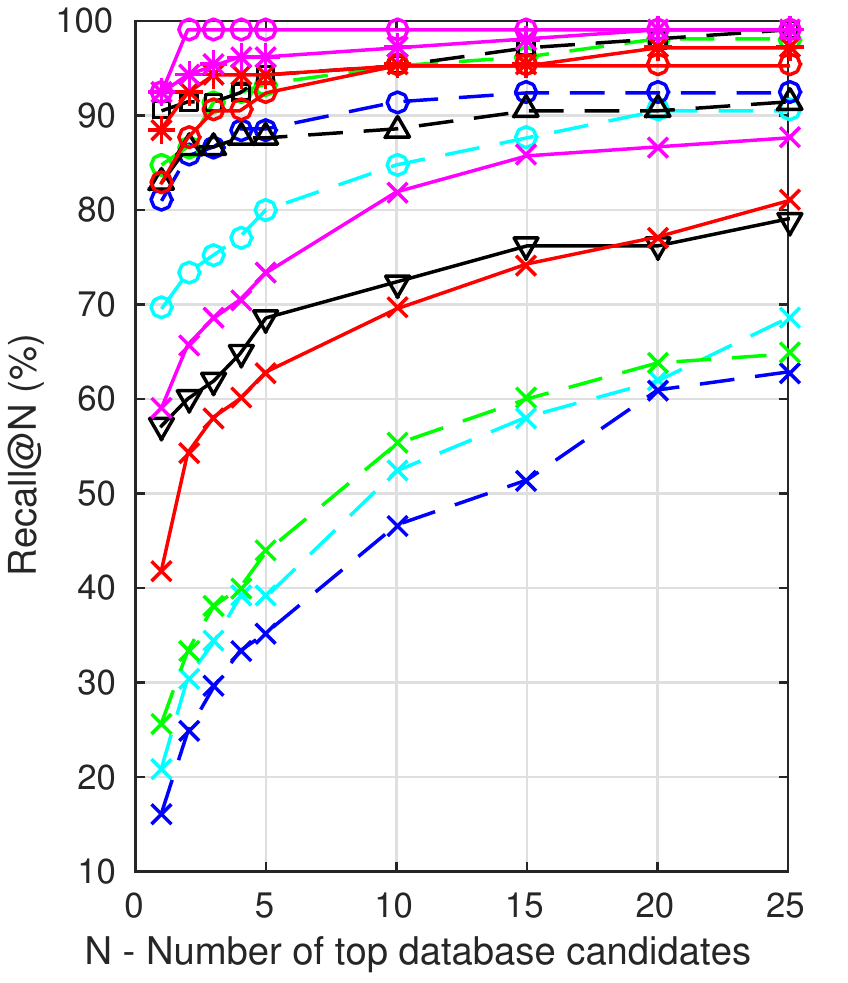}
    }
    \subfloat[Tokyo 24/7 sunset/night]{
        \includegraphics[width=\wStoa]{figures/tokyoCvpr15_res0n_noleg.pdf}
    }
\end{center}
    \caption{
\figStoaCaption
\cite{Torii15} only evaluated on Tokyo 24/7 as the method relies on depth data not available
in other datasets.
\vspace{-0.3cm}
}
\label{fig:stoaSup}
\end{figure*}
}

\def\vertTextFS[#1]{\multirow{1}{*}[1cm]{\rotatebox[origin=b]{90}{\parbox{1cm}{\footnotesize \centering #1}}}}

\def\figFocusSup{
\def\wF{0.12\linewidth}
\begin{figure*}[t!]
\begin{center}
\hspace*{-0.3cm}
\begin{tabular}{c@{~~~~}ccccccc}
    \vertTextFS[Input \\ image] &
    \includegraphics[width=\wF]{figures/focus/002535_pitch1_yaw2.pdf} &
    \includegraphics[width=\wF]{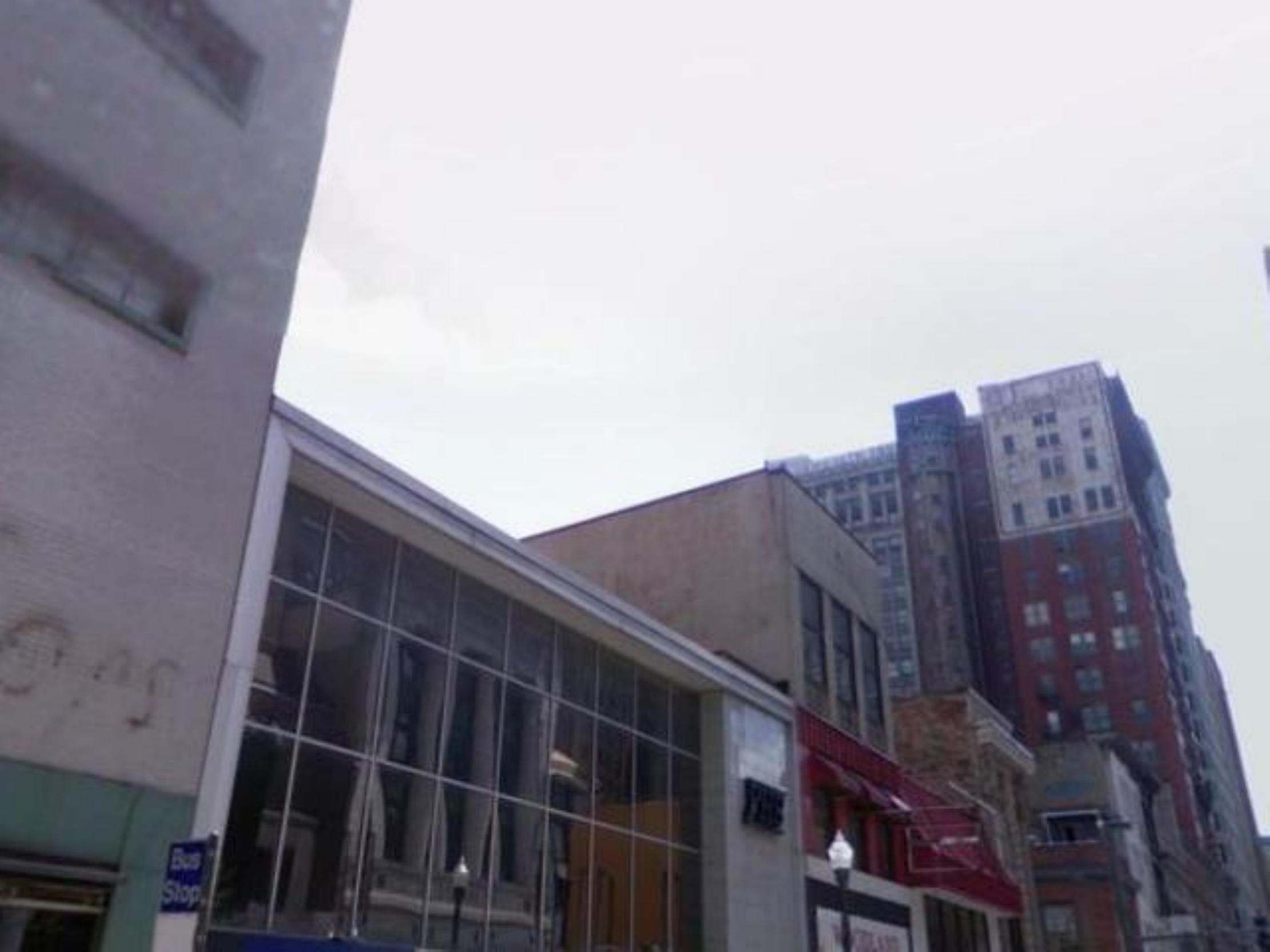} &
    \includegraphics[width=\wF]{figures/focus/000171_pitch1_yaw4.pdf} &
    \includegraphics[width=\wF]{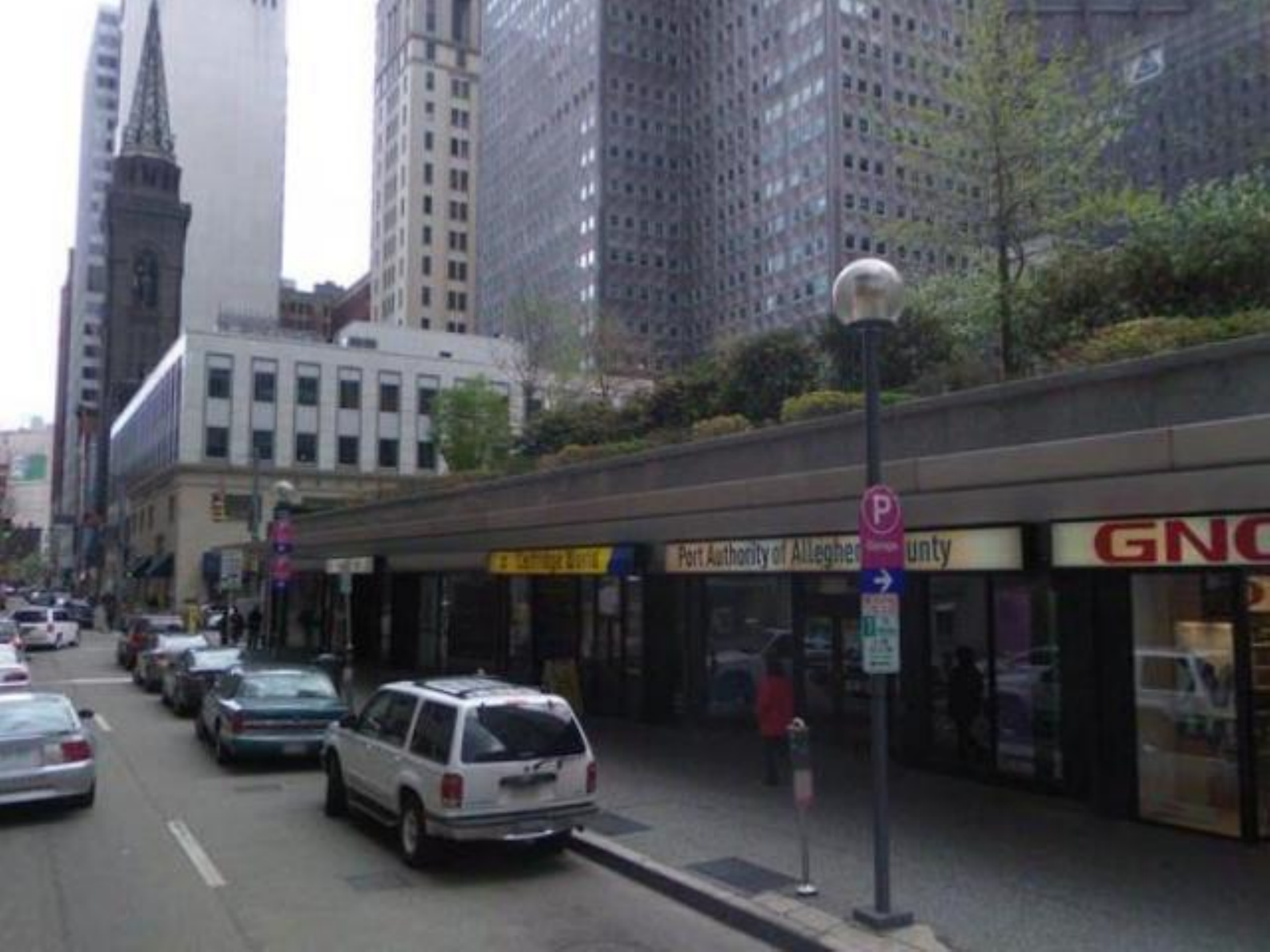} &
    \includegraphics[width=\wF]{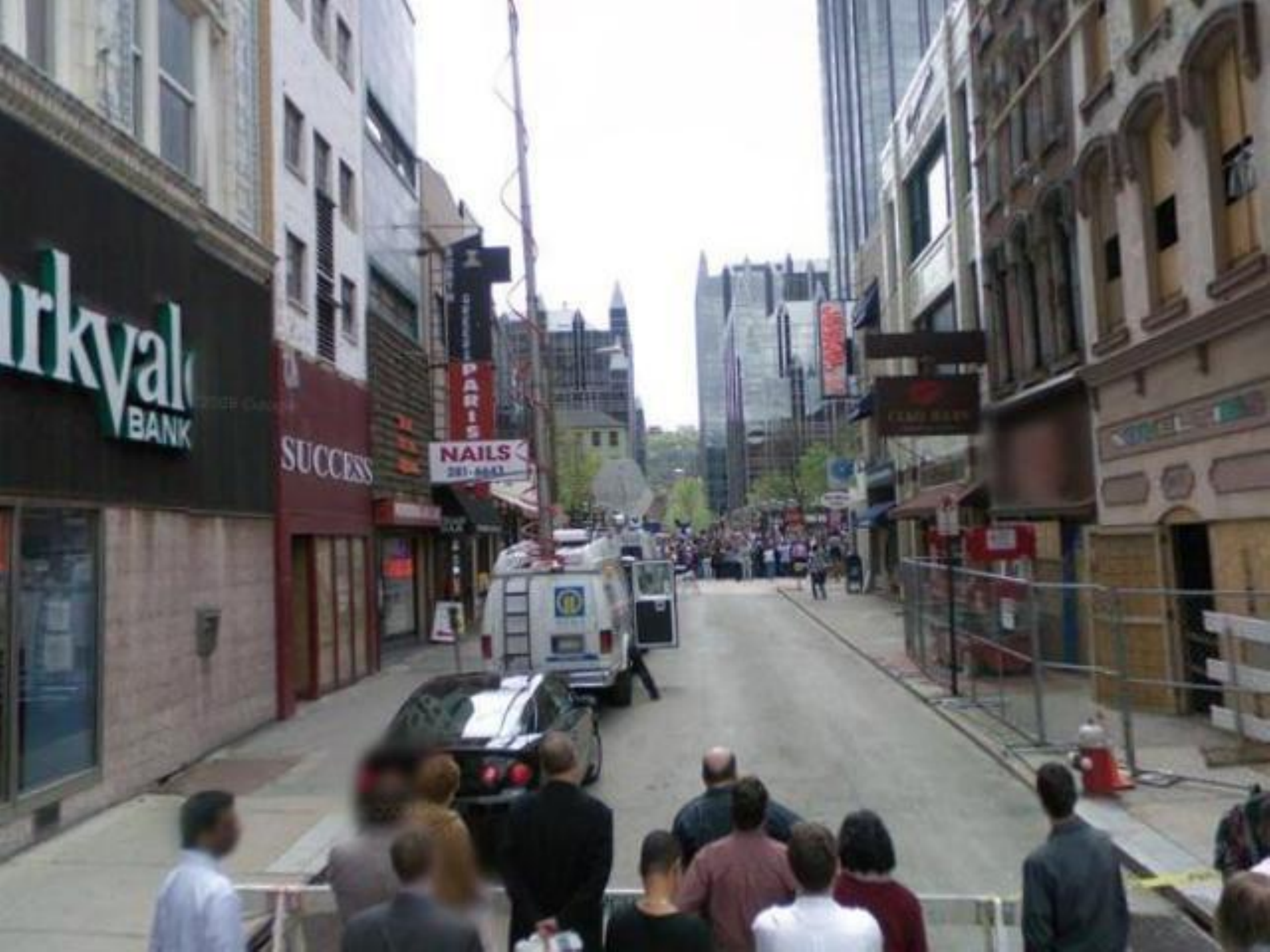} &
    \includegraphics[width=\wF]{figures/focus/000634_pitch1_yaw7.pdf} &
    \includegraphics[width=\wF]{figures/focus/002278_pitch1_yaw12.pdf}
    \\
    \vertTextFS[AlexNet \\ ours] &
    \includegraphics[width=\wF]{figures/focus/002535_pitch1_yaw2_trained_max.pdf} &
    \includegraphics[width=\wF]{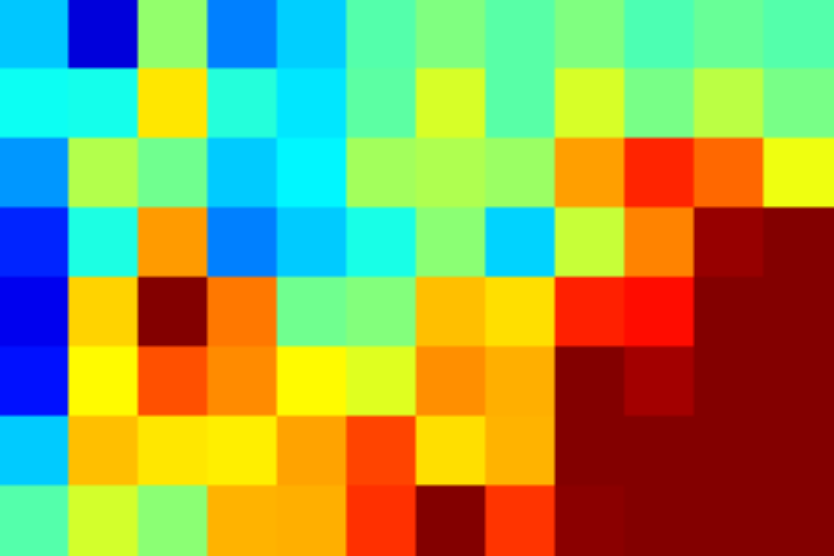} &
    \includegraphics[width=\wF]{figures/focus/000171_pitch1_yaw4_trained_max.pdf} &
    \includegraphics[width=\wF]{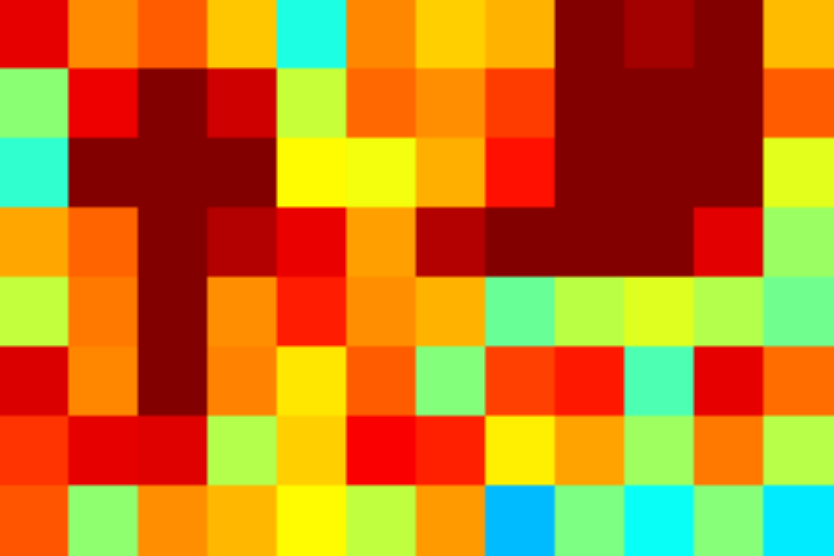} &
    \includegraphics[width=\wF]{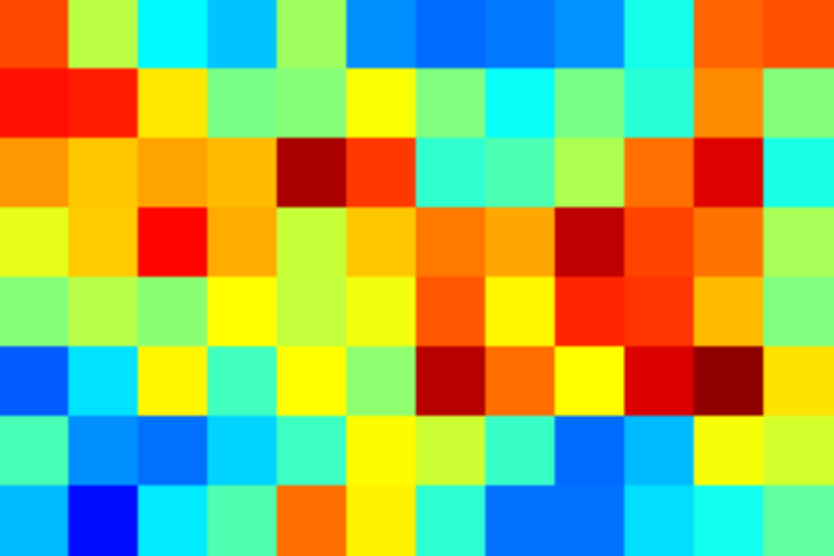} &
    \includegraphics[width=\wF]{figures/focus/000634_pitch1_yaw7_trained_max.pdf} &
    \includegraphics[width=\wF]{figures/focus/002278_pitch1_yaw12_trained_max.pdf}
    \\
    \vertTextFS[AlexNet \\ off-shelf] &
    \includegraphics[width=\wF]{figures/focus/002535_pitch1_yaw2_caffe_max.pdf} &
    \includegraphics[width=\wF]{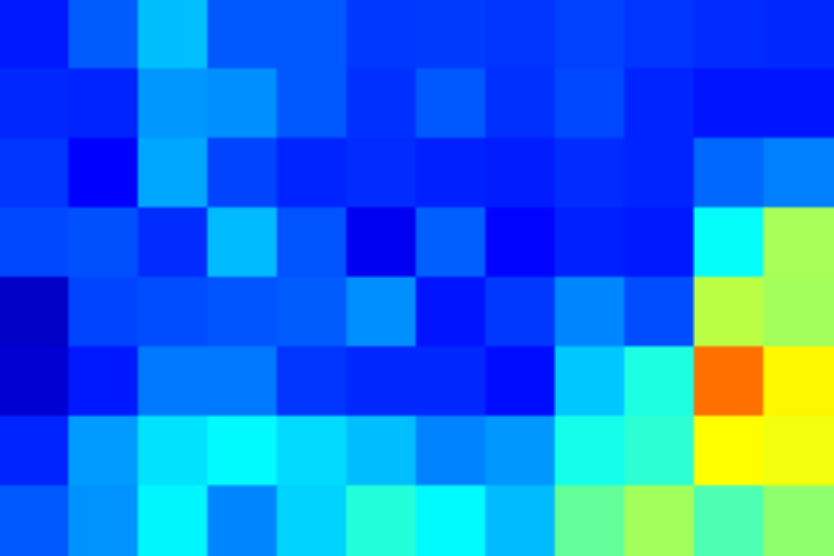} &
    \includegraphics[width=\wF]{figures/focus/000171_pitch1_yaw4_caffe_max.pdf} &
    \includegraphics[width=\wF]{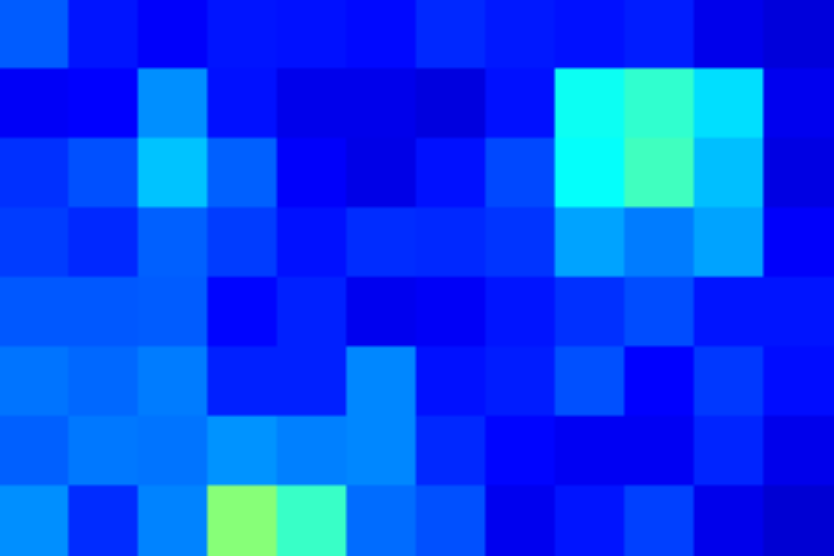} &
    \includegraphics[width=\wF]{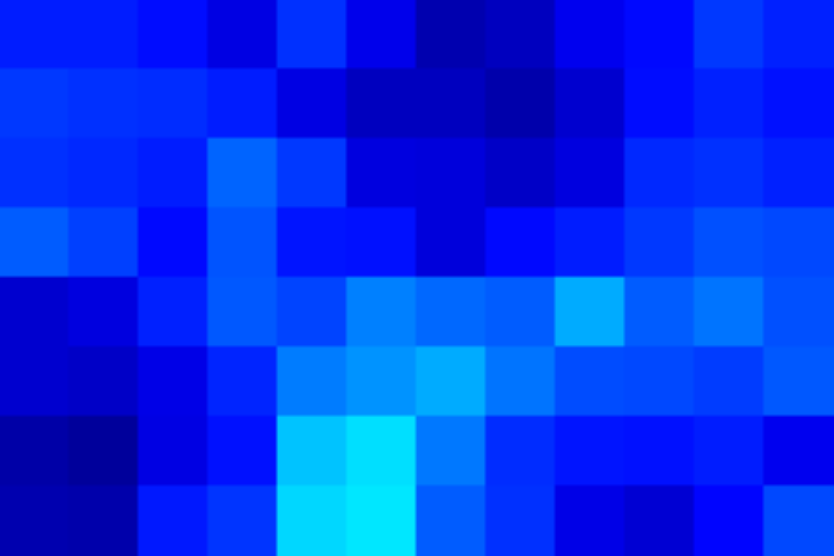} &
    \includegraphics[width=\wF]{figures/focus/000634_pitch1_yaw7_caffe_max.pdf} &
    \includegraphics[width=\wF]{figures/focus/002278_pitch1_yaw12_caffe_max.pdf}
    \\
    \vertTextFS[Places205 \\ off-shelf] &
    \includegraphics[width=\wF]{figures/focus/002535_pitch1_yaw2_places_max.pdf} &
    \includegraphics[width=\wF]{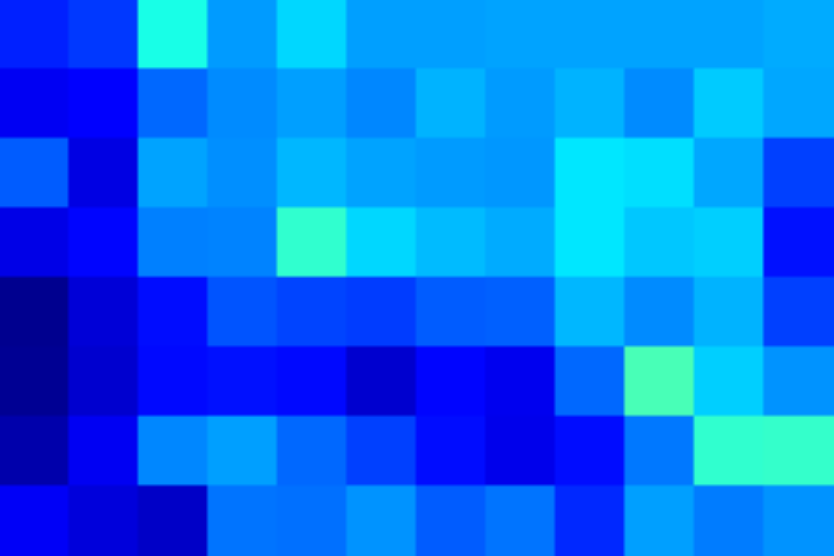} &
    \includegraphics[width=\wF]{figures/focus/000171_pitch1_yaw4_places_max.pdf} &
    \includegraphics[width=\wF]{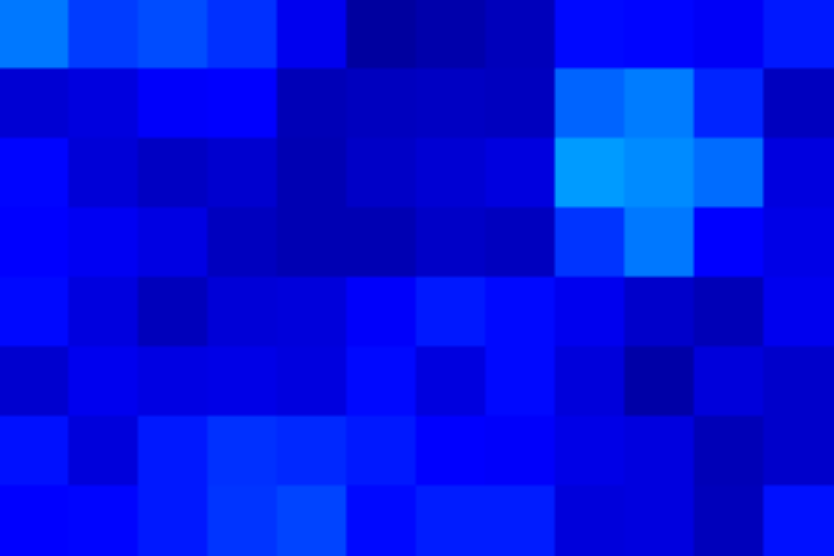} &
    \includegraphics[width=\wF]{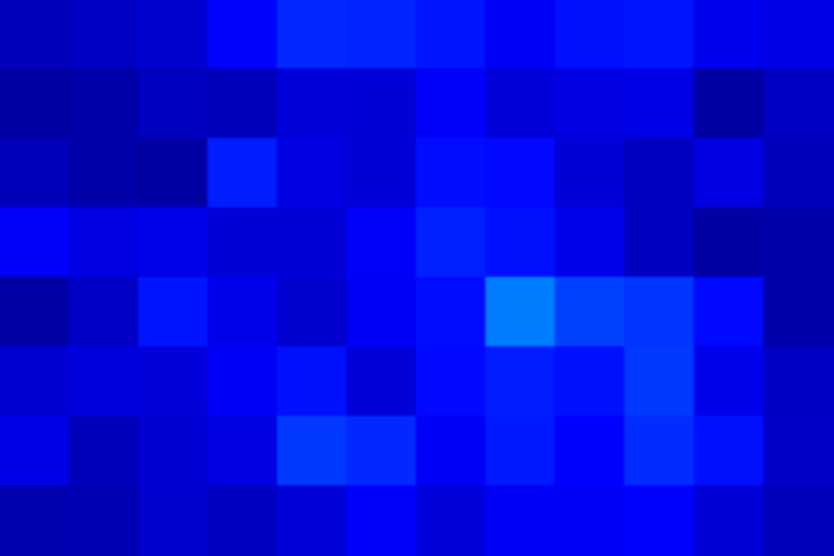} &
    \includegraphics[width=\wF]{figures/focus/000634_pitch1_yaw7_places_max.pdf} &
    \includegraphics[width=\wF]{figures/focus/002278_pitch1_yaw12_places_max.pdf}
\end{tabular}
\end{center}
    \caption{
\figFocusCaption
}
\label{fig:focusSup}
\end{figure*}
}

\def\tabDb{
\begin{table}[t!]
\begin{center}
\begin{tabular}{l|rr}
    Dataset & Database & Query set \\
    \hline\hline
    Pitts250k-train & 91,464 & 7,824 \\
    Pitts250k-val   & 78,648 & 7,608 \\
    Pitts250k-test  & 83,952 & 8,280 \\
    \hline
    Pitts30k-train & 10,000 & 7,416 \\
    Pitts30k-val   & 10,000 & 7,608 \\
    Pitts30k-test  & 10,000 & 6,816 \\
    \hline
    Tokyo Time Machine-train & 49,104 & 7,277 \\
    Tokyo Time Machine-val   & 49,056 & 7,186 \\
    Tokyo 24/7 (-test)       & 75,984 &  315 \\
\end{tabular}
\end{center}
\caption{ {\bf Datasets.}
Sizes of datasets used in experiments.
All train/val(idation)/test datasets are mutually disjoint geographically.
}
\label{tab:db}
\end{table}
}

\def\qualCapt{
    {\bf Examples of retrieval results for challenging queries on Tokyo 24/7.}
Each column corresponds to one test case:
the query is shown in the first row,
the top retrieved image using
our best method (trained VGG-16 NetVLAD + whitening) in the second,
and the top retrieved image using
the best baseline (RootSIFT + VLAD + whitening) in the last row.
The green and red borders correspond to correct and incorrect retrievals, respectively. Note that our learnt descriptor can recognize the same place despite large changes in appearance due to illumination (day/night), viewpoint and partial occlusion by cars, trees and people.
}

\def\posex[#1]{\fcolorbox{green}{green}{\includegraphics[width=\wQ]{figures/qual/#1.pdf}}}
\def\negex[#1]{\fcolorbox{red}{red}{\includegraphics[width=\wQ]{figures/qual/#1.pdf}}}
\def\qex[#1]{\fcolorbox{white}{white}{\includegraphics[width=\wQ]{figures/qual/#1/query_blur.pdf}}}

\def\wQ{0.20\linewidth}
\fboxsep=1mm%
\fboxrule=0pt%

\def\vertText[#1]{\multirow{1}{*}[2cm]{\rotatebox[origin=c]{90}{\parbox{2cm}{\centering #1}}}}

\def\figQualAB{
\begin{figure*}[t!]
\begin{center}
\hspace*{-0.3cm}
\begin{tabular}{lc@{~~~}c@{~~~}c@{~~~}c}
\vertText[Query] &
\qex[vggvladours/000057] & \qex[vggvladours/000015] & \qex[vggvladours/000032] & \qex[vggvladours/000089] \\
\vertText[Ours] &
\posex[vggvladours/000057/000001] & \posex[vggvladours/000015/000001] & \posex[vggvladours/000032/000001] & \posex[vggvladours/000089/000001] \\
\vertText[Best baseline] &
\negex[sift/000057/000001] & \negex[sift/000015/000001] & \negex[sift/000032/000001] & \negex[sift/000089/000001]
\\ \\
\hline \\
\vertText[Query] &
\qex[vggvladours/000029] & \qex[vggvladours/000039] & \qex[vggvladours/000008] & \qex[vggvladours/000043] \\
\vertText[Ours] &
\posex[vggvladours/000029/000001] & \posex[vggvladours/000039/000001] & \posex[vggvladours/000008/000001] & \posex[vggvladours/000043/000001] \\
\vertText[Best baseline] &
\negex[sift/000029/000001] & \posex[sift/000039/000001] & \negex[sift/000008/000001] & \negex[sift/000043/000001]
\end{tabular}
\end{center}
    \caption{
\qualCapt
}
\label{fig:ranked12}
\end{figure*}
}

\def\figQualC{
\begin{figure*}[p!]
\begin{center}
\hspace*{-0.3cm}
\begin{tabular}{lc@{~~~}c@{~~~}c@{~~~}c}
\vertText[Query] &
\qex[vggvladours/000048] & \qex[vggvladours/000063] & \qex[vggvladours/000093] & \qex[vggvladours/000129] \\
\vertText[Ours] &
\posex[vggvladours/000048/000001] & \posex[vggvladours/000063/000001] & \posex[vggvladours/000093/000001] & \negex[vggvladours/000129/000001] \\
\vertText[Best baseline] &
\posex[sift/000048/000001] & \negex[sift/000063/000001] & \negex[sift/000093/000001] & \negex[sift/000129/000001] \\
\end{tabular}
\end{center}
    \caption{
\qualCapt
The last column corresponds to a difficult query, which is hard for our method
because of its overall very dark appearance.
}
\label{fig:ranked3}
\end{figure*}
}

\def\tabResRetDim{
\begin{table}[!t]
\begin{center}
\small
\hspace*{-0.4cm}%
\begin{tabular}{l|r|c@{~~~}c|c@{~~~}c|c@{~~~}c}
    Method & \multicolumn{1}{c|}{Dim.}
    & \multicolumn{2}{c|}{Oxford 5k}
    & \multicolumn{2}{c|}{Paris 6k}
    & \multicolumn{2}{c}{Holidays}
    \\
    & & full & crop & full & crop & orig & rot
    \\
    \hline

    NetVLAD off-shelf & 16
        & 28.7 & 28.1 & 36.8 & 38.2 & 56.6 & 60.3
    \\
    & 32
        & 36.5 & 36.0 & 48.9 & 51.9 & 68.0 & 71.7
    \\
    & 64
        & 40.1 & 38.9 & 55.7 & 58.1 & 76.5 & 80.4
    \\
    & 128
        & 49.0 & 49.8 & 60.1 & 63.2 & 79.1 & 83.3
    \\
    & 256 & 53.4 & 55.5 & 64.3 & 67.7 & 82.1 & 86.0
    \\
    & 512
        & 56.7 & 59.0 & 67.5 & 70.2 & 82.9 & 86.7
    \\
    & 1024
        & 60.2 & 62.6 & 70.9 & 73.3 & 83.9 & 87.3
    \\
    & 2048
        & 62.8 & 65.4 & 73.7 & 75.6 & 84.9 & 88.2
    \\
    & 4096
        & 64.4 & 66.6 & 75.1 & 77.4 & 84.9 & 88.3
    \\
    \hline
    NetVLAD trained & 16
        & 32.5 & 29.9 & 45.1 & 44.9 & 54.8 & 58.6
    \\
    & 32
        & 43.4 & 42.6 & 53.5 & 54.4 & 67.5 & 71.2
    \\
    & 64
        & 53.6 & 51.1 & 61.8 & 63.0 & 75.4 & 79.3
    \\
    & 128
        & 60.4 & 61.4 & 68.7 & 69.5 & 78.8 & 82.6
    \\
    & 256
        & 62.5 & 63.5 & 72.0 & 73.5 & 79.9 & 84.3
    \\
    & 512
        & 65.6 & 67.6 & 73.4 & 74.9 & 81.7 & 86.1
    \\
    & 1024
        & 66.9 & 69.2 & 75.7 & 76.5 & 82.4 & 86.5
    \\
    & 2048
        & 67.7 & 70.8 & 77.0 & 78.3 & 82.8 & 86.9
    \\
    & 4096
        & 69.1 & 71.6 & 78.5 & 79.7 & 83.1 & 87.5
    \\
\end{tabular}
\end{center}
 \vspace{-0.2cm}
\caption{ {\bf Image and object retrieval for varying dimensionality of NetVLAD.}
We compare our best trained network (VGG-16, $f_{VLAD}$),
and the corresponding off-the-shelf network (whitening learnt on Pittsburgh),
on standard image and object retrieval benchmarks,
while varying the dimensionality (Dim.) of the image representation.
}
\label{tab:retrievalDim}
\end{table}
}

\section*{Appendices}
The below appendices describe the implementation details (appendix~\ref{sup:impl}), give details of the Google Street View Time Machine datasets (appendix~\ref{sup:datasets}) and provide additional results (appendix~\ref{sup:res}).

\section{Implementation details}
\label{sup:impl}

We use two base architectures which are extended with Max pooling ($f_{max}$) and
our NetVLAD ($f_{VLAD}$) layers:
AlexNet \cite{Krizhevsky12} and
VGG-16 \cite{Simonyan15};
both are cropped at the last convolutional layer (conv5), before ReLU.
For Max we use raw conv5 descriptors (with no normalization) while for VLAD and NetVLAD
we add an additional descriptor-wise L2-normalization layer after conv5.
We found that not normalizing for Max and normalizing for VLAD/NetVLAD
generalizes across architectures, \ie these are the best configurations
for both AlexNet and VGG-16.

The number of clusters used in all VLAD / NetVLAD experiments is $K=64$.
The NetVLAD layer parameters are initialized to reproduce the conventional VLAD
vectors by clustering conv5 descriptors
extracted from a subsample of the train set for each dataset.
The $\alpha$ parameter used for initialization is chosen to be large, such that
the soft assignment weights $\bar a_k(\vc x_i)$ are very sparse
in order to mimic the conventional VLAD well.
Specifically, $\alpha$ is computed so that the the ratio of
the largest and the second largest soft assignment weight $\bar a_k(\vc x_i)$
is on average equal to 100.

We use the margin $m=0.1$,
learning rate 0.001 or 0.0001 depending on the experiment,
which is halved every 5 epochs,
momentum 0.9, weight decay 0.001, batch size of 4 tuples
(a tuple contains many images, \cf equation \eqref{eq:loss} of the main paper),
and train for at most 30 epochs but convergence usually occurs much faster.
The network which yields the best recall@5 on the validation set is used
for testing.

As the VGG-16 network is much deeper and more GPU-memory hungry than AlexNet,
it was not possible to train it in its entirety. Instead, in the light of experiments in table \ref{tab:layers} of the main paper,
the VGG-16 network is only trained down to conv5 layer.

To create the training tuple for a query,
we use all of its potential positives
(images within 10 meters),
and we perform randomized hard negative mining for the negatives
(images further away than 25 meters).
The mining is done by keeping the 10 hardest negatives from a pool of
1000 randomly sampled negatives and 10 hardest negatives from the previous epoch.
We find that remembering previous hard negatives adds stability to the training process.

Naively implemented, the aforementioned training procedure would be too slow.
Processing each training tuple would require a forward pass on more than
 1010 full-resolution images.
Instead, we compute image representations for the entire
training query and database sets and cache them for a certain amount of time.
The hard negative mining then uses these cached but slightly stale representations
to obtain the 10 hardest examples and the forward and backward passes
are only performed on these 10, compared to the original 1010,
thus providing a huge computational saving.
However, it is important to recompute the cached representations every once in a while.
We have observed slow convergence if the cache is fixed for too long as the network
learns quickly to be better than the fixed cache and then wastes time overfitting it.  
We found that recomputing the cached representations for hard negative mining every 500 to 1000 training queries 
yields a good trade-off between epoch duration, convergence speed
and quality of the solution.
As described earlier, we half the learning rate every 5 epochs --
this causes the cached representations to change less rapidly, so
we half the recomputation frequency every 5 epochs as well.
All training and evaluation code, as well as our trained networks,
are online \cite{netvladurl}, implemented in the MatConvNet framework \cite{Vedaldi15}.
Additional tuning of parameters and jittering could further improve performance 
as we have still observed some amount of overfitting. %

\section{Google Street View Time Machine datasets}
\label{sup:datasets}
\label{sup:TM}

Table \ref{tab:db} shows the sizes of datasets used in this work,
described in section \ref{sec:datasets} of the main paper. The newly collected Tokyo Time Machine (TokyoTM) database was
generated from downloaded Time Machine panoramas, such that
each panorama is represented by a set of 12 perspective images sampled evenly
in different orientations~\cite{Knopp10,Chen11b,Gronat13,Torii13,Torii15}.
Figure \ref{fig:timemachineSup} shows example images from the dataset.
For each query, the positive and negative sets are sampled from the database so that they have a time stamp at least one month away from the time stamp of the query. 
This is done for both training (training/val sets) and evaluation (testing set).
All datasets are publicly available:
Pitts250k from the authors of \cite{Torii13},
Tokyo 24/7 from the authors of \cite{Torii15},
while we will share TokyoTM and Pitts30k on request.

\tabDb

\figTMSup

\section{Additional results and discussions}
\label{sup:res}

\paragraph{VLAD versus Max.}
Figure~\ref{fig:dim} shows that NetVLAD performance decreases gracefully
with dimensionality: On Tokyo 24/7,
128-D NetVLAD performs similarly to 512-D Max,
resulting in {\em four} times
more compact representation for the same performance.
Similarly, on Pitts250k-test NetVLAD achieves a two-fold memory saving
compared to Max.
Furthermore, NetVLAD+whitening outperforms Max pooling convincingly when
reduced to the same dimensionality.

\figDim
\figFocusSup

\figStoaSup

\paragraph{Max versus Sum.}
Recent work \cite{Babenko15} suggests that Sum pooling performs better than Max pooling.
Indeed, in our experiments Sum outperforms Max in the off-the-shelf set-up
(recall@5 on Pitts250k-test -- Sum: 67.9\%, Max: 59.3\%),
but only for VGG-16, not AlexNet.
Our training also works for Sum getting a significant improvement over the off-the-shelf set-up
(+21\% relative),
but after training Max still performs better than Sum
(Max: 88.7\%, Sum: 82.3\%).

\tabResTM
\tabResRet
\tabResRetDim

\paragraph{Further results.}
Figure \ref{fig:stoaSup} reports a complete set of results that did not fit into figure \ref{fig:stoa} of the main paper. 
Namely, it includes results on the Pitts30k-test and the complete breakdown of day versus sunset/night
queries for the Tokyo 24/7 benchmark as done in~\cite{Torii15}.
Table \ref{tab:timemachine} contains additional results showing the importance
of training with Time Machine imagery.
Figure \ref{fig:focusSup} shows additional visualizations of what has been learnt by our method. Please see
 section \ref{sec:res:visual} of the main paper for the details of the visualization.
\figQualAB
\figQualC
Figures \ref{fig:ranked12} and \ref{fig:ranked3} compare the top ranked images of our method versus the best baseline.

\paragraph{Benefits of end-to-end training for place recognition}
As shown in the main paper and in figure \ref{fig:stoaSup}, 
the popular idea of using pretrained networks ``off-the-shelf''
\cite{Razavian14,Azizpour14,Gong14,Babenko15,Razavian15}
is sub-optimal as the networks trained for object or scene classification
are not necessary suitable for the end-task of place recognition.
The failure of the ``off-the-shelf networks" is not surprising --
apart from the obvious benefits of training, it is not clear
why it should be meaningful to directly compare conv5 activations using Euclidean distance
as they are trained to be part of the network architecture. For example,
one can insert an arbitrary affine transformation of the features that can be countered by the following fully connected layer (fc6).
This is not a problem when transferring the pre-trained representation for object classification~\cite{Oquab14,Zeiler14} or detection~\cite{Girshick14} tasks, as such transformation can be countered by the follow-up adaptation~\cite{Oquab14} or classification~\cite{Girshick14,Zeiler14} layers that are trained for the target task. However, this is not the case for retrieval~\cite{Razavian14,Azizpour14,Gong14,Babenko15,Razavian15} when Euclidean distance is applied directly on the output ``off-the-shelf" descriptors.

\label{sup:ret}
\paragraph{Image retrieval experiments.}
We use our best performing network (VGG-16, $f_{VLAD}$ with whitening and dimensionality reduction down to 256-D)
trained completely on Pittsburgh, to extract image representations
for standard object and image retrieval benchmarks
(Oxford 5k \cite{Philbin07}, Paris 6k  \cite{Philbin08}, Holidays \cite{Jegou08}).
Table \ref{tab:retrieval} compares NetVLAD to the
state-of-the-art compact image representations (256-D).
Our representation achieves the best mAP on Oxford and Paris by a large margin,
\eg +20\% relative improvement on Oxford 5k (crop).
It also sets the state-of-the-art on Holidays, but here training is detrimental
as the dataset is less building-oriented
(\eg it also contains paysages, underwater photos, boats, cars, bears, \etc),
while our training only sees images from urban areas. We believe training on data  more diverse than Pittsburgh streets can further improve performance. %
The complete set of NetVLAD results for different output
dimensions is shown in table~\ref{tab:retrievalDim}.

}{
}

\end{document}